\pdfoutput=1

\documentclass[11pt]{article}

\usepackage{acl}

\usepackage{relsize} 

\usepackage{times}
\usepackage{latexsym}

\usepackage{booktabs}
\usepackage{subcaption} 
\usepackage{multirow} 
\usepackage{array} 
\usepackage{titling} 
\usepackage{bbding}
\usepackage{pifont}
\usepackage{hyperref}
\usepackage{cleveref} 

\usepackage[shortlabels]{enumitem}
\usepackage{comment}

\usepackage[T1]{fontenc}

\usepackage[utf8]{inputenc}

\usepackage{graphicx}
\usepackage{float}

\usepackage{rotating} 

\usepackage{microtype}

\usepackage{inconsolata}

\usepackage{color}
\definecolor{darkgreen}{rgb}{0,0.5,0}

\usepackage[normalem]{ulem}

\def\MLdel{\bgroup\markoverwith{\textcolor{red}{\rule[0.4ex]{2pt}{3pt}}}\ULon}

\def\SBdel{\bgroup\markoverwith{\color[rgb]{0.31, 0.416, 0.659}{\rule[0.4ex]{2pt}{3pt}}}\ULon}

\def\PSdel{\bgroup\markoverwith{\textcolor{blue}{\rule[0.4ex]{2pt}{3pt}}}\ULon}
\MakeRobust\PSdel
\def\ODdel{\bgroup\markoverwith{\textcolor{darkgreen}{\rule[0.4ex]{2pt}{3pt}}}\ULon}
\MakeRobust\ODdel

\title{\includegraphics[scale=0.065]{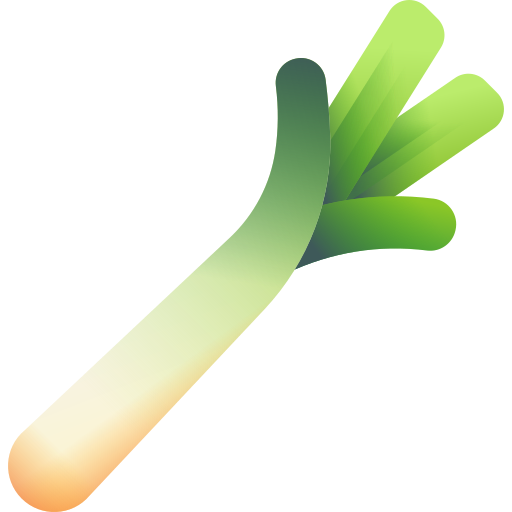}Leak, Cheat, Repeat: Data Contamination \\and Evaluation Malpractices in Closed-Source LLMs}

\author{Simone Balloccu \quad Patrícia Schmidtová \quad Mateusz Lango \quad Ondřej Dušek\\
         Charles University, Faculty of Mathematics and Physics\\
Institute of Formal and Applied Linguistics\\
Prague, Czech Republic\\
\texttt{\{balloccu,schmidtova,lango,odusek\}@ufal.mff.cuni.cz}}

\begin{document}
\maketitle
\begin{abstract}
  Natural Language Processing (NLP) research is increasingly focusing on the use of Large Language Models (LLMs), with some of the most popular ones being either fully or partially closed-source. The lack of access to model details, especially regarding training data, has repeatedly raised concerns about data contamination among researchers. Several attempts have been made to address this issue, but they are limited to anecdotal evidence and trial and error. Additionally, they overlook the problem of \emph{indirect} data leaking, where models are iteratively improved by using data coming from users. In this work, we conduct the first systematic analysis of work using OpenAI's GPT-3.5 and GPT-4, the most prominently used LLMs today, in the context of data contamination. By analysing 255 papers and considering OpenAI's data usage policy, we extensively document the amount of data leaked to these models during the first year after the model's release. We report that these models have been globally exposed to $\sim$4.7M samples from 263 benchmarks. At the same time, we document a number of evaluation malpractices emerging in the reviewed papers, such as unfair or missing baseline comparisons and reproducibility issues. We release our results as a collaborative project on \url{https://leak-llm.github.io/}, where other researchers can contribute to our efforts.
\end{abstract}

\section{Introduction}

The recent emergence of large language models (LLMs), that show remarkable performance on a wide range of tasks, has led not only to a dramatic increase in their use in research but also to a growing number of companies joining the race for the biggest and most powerful models.
In pursuing a competitive advantage, many popular LLMs today are locked behind API access and their details are unknown~\citep{GPT4-technical-report, thoppilan2022lamda, touvron2023llama}. 
This includes model weights~\citep{GPT4-technical-report}, training data~\citep{piktus-etal-2023-roots}, or infrastructural details to assess model carbon footprint \citep{lacoste2019quantifying}. 

In particular, the lack of information on training data raises important questions about the credibility of LLMs performance evaluation. The data from which these models learn, typically collected automatically by scraping documents from the web, may contain training, validation, and -- most critically -- test sets coming from NLP benchmarks. Because of this, researchers and stakeholders may later inadvertently evaluate LLMs on the same data they were trained on.
This phenomenon, known as data contamination, may not be an issue in the general use of commercial LLMs, where adherence to research principles is not mandatory, but it becomes a serious problem when these models are widely used and evaluated in research.

Unfortunately, many proprietary models are locked behind inference-only APIs, making it hard to inspect data contamination. Because of this, existing work on the matter mostly focuses on detecting extreme forms of overfitting and memorization, such as the model's ability to generate benchmarks verbatim. These approaches are not only limited but also neglect that recent proprietary LLMs get iteratively improved from user interactions. If such interactions involve benchmark data (for example when researchers evaluate LLMs against baselines), the model may, in fact, become contaminated even if it was contamination-free during its initial training. We refer to this phenomenon as \emph{indirect data leaking}.

In this paper, we address the issue of indirect data contamination in closed-source\footnote{In this paper we use the terms “proprietary” and “closed-source” interchangeably to refer to these models.} LLMs by conducting a systematic literature review. We review 255 papers and carefully detail data leakage emerging from them. We focus primarily on the models accessible through OpenAI's ChatGPT,\footnote{\href{https://openai.com/blog/chatgpt}{https://openai.com/blog/chatgpt}} (GPT-3.5 and GPT-4\footnote{\href{https://openai.com/gpt-4}{https://openai.com/gpt-4}}) as these are the most frequently used commercial LLMs in NLP research. By considering OpenAI's data usage policy, we assess how much data was reported to be sent to the models in a way that it could be used for further training, hence giving the models an unfair advantage during evaluation.
We also report a series of emergent evaluation malpractices, including lack of comparison with other approaches, differences in the evaluation scale (e.g., evaluating open models on entire benchmarks while comparing to proprietary LLMs evaluated on samples only), lack of code and data access, or data leakage even in situations where it could be avoided.
To our knowledge, this work is the most comprehensive and extensive quantification of the data leakage issue in LLMs to date.

In short, our contributions are as follows:
\begin{enumerate}[label={(\arabic*)}]
    \item We systematically analyse 255 papers evaluating OpenAI's GPT-3.5 and GPT-4 on a variety of tasks in NLP and other domains (\Cref{sec:methods}).
    \item For each paper, we estimate the amount of data leaked in such a way that it could be used for further model training. Overall, we conclude that $\sim$42\% of the reviewed papers leaked data to GPT-3.5 and GPT-4, for a total of $\sim$4.7M benchmark samples across 263 benchmarks (\Cref{ssec:contamination}).
    \item We further analyse the evaluation protocols of the selected papers, and we reveal some critical malpractices limiting both the experiments' reproducibility and fairness (\Cref{ssec:reproducibility,ssec:bad-eval}).
    \item Based on our findings, we propose a list of suggested practices for the evaluation of closed-source LLMs (\Cref{sec:best-practices}).
\end{enumerate}

We believe that our work can contribute to ongoing efforts on quantifying LLM data contamination by pointing out which datasets are worthy of further investigation. We release our survey results as a collaborative repository, in the form of a webpage at \url{https://leak-llm.github.io/}. It features a list of datasets, detailing the extend of data leakage for each of them. We invite other researchers to contribute any additional known leaks to the list.

\section{Prior Work on LLM Data Contamination}
\label{sec:related-work}

Work on LLMs data contamination traces back to OpenAI's GPT-3~\citep{GPT3-original,magar-schwartz-2022-data}, one of the first models with API-only access and limited training data disclosure.
Despite results hinting at the presence of significant data contamination~\citep{GPT3-contamination,magar-schwartz-2022-data}, the model has been used extensively in research and the issue was rarely taken into account when interpreting its performance.
With the release of ChatGPT and following closed-source models to general public,\footnote{Including GPT-4~\citep{GPT4-technical-report}, Google's \href{https://blog.google/technology/ai/lamda/}{LaMDA} \cite{thoppilan2022lamda} and {\href{https://ai.google/discover/palm2/}{PaLM}} \cite{chowdhery_palm_2022}, Cohere's \href{https://cohere.com/models/command}{Command} and Anthropic's \href{https://claude.ai/}{Claude}.} the data contamination topic became an even more pressing issue.

When a model is closed-source, it becomes implicitly complex to assess data contamination from known benchmarks. Therefore, only few practical approaches have been proposed to investigate the issue.

One notable example is the LM Contamination Index,\footnote{\href{https://hitz-zentroa.github.io/lm-contamination/}{https://hitz-zentroa.github.io/lm-contamination/}} featuring a regularly updated estimate of contamination for a list of both open and proprietary models. This approach works by zero-shot prompting the model to generate instances from specific datasets, providing details on the required split and format~\citep{llm_contamination_blog}.
The premise is that no model should be able to replicate specific benchmark formats without having seen them first.

More applied approaches have been proposed recently~\citep{golchin2023time}, where LLMs are prompted to complete a given sentence coming from a known benchmark. The completion is then compared with the original reference through text overlap metrics and a statistical test is used to assess if the model is contaminated. 

Although these preliminary works are promising, they cannot be fully trusted and have some limitations.
Most importantly,  they are based on an assessment of the model's ability to generate an example from the benchmark.
The recall of such methods can be affected by two issues:

\begin{enumerate}[label={(\arabic*)}]
\item Some closed-source models have incorporated special filters into their decoding algorithms that prevent them from generating texts that significantly overlap with their training sets~\cite{copilot,ippolito-etal-2023-preventing}. This creates an additional noise for the detection methods and results in the lack of confidence that even the datasets tested negative for data leakage are not present in LLM training data.

\item Such approaches can only detect the most extreme form of overfitting which results in (almost) complete memorization of data samples by the model. However, even a regular adjustment of the model by training on the leaked data, which does not necessarily lead to its memorization, poses a problem for fair comparisons.
\end{enumerate}

\section{The Issue of Indirect Data Leaking}
\label{sec:indirect-leak}

The related work presented in \Cref{sec:related-work} approaches the issue of data contamination mainly by backtracking models' training data. 
It is commonly assumed that using benchmarks available only to authorised parties, or datasets being constructed after the ChatGPT release, is a guarantee that they have not been leaked.
This ignores the fact that models using reinforcement learning from human feedback (RLHF,~\citealp{NEURIPS2022_b1efde53}), such as those used by ChatGPT, are subject to repeated updates~\citep{aiyappa-etal-2023-trust} with training data also coming from user interactions.
This process leads to a previously overlooked phenomenon, where new data are leaked to the model just through using it. We refer to this problem as \emph{indirect data leaking} and consider it a new development of the issue for two main reasons:

\begin{enumerate}[label={(\arabic*)}]
    \item Unlike plain text scraped from the internet, data from users might be harder to inspect for contamination as it might involve model prompts, textual alterations, or truncation of benchmark samples.
    \item Users supply the data along with instructions on how to perform the task. In LLMs, this can be considered a novel form of gold-standard data for continued training, even in the absence of target labels. Model updates on such data are likely much more effective than plain in-domain text.
\end{enumerate}

The issue (1) is particularly complex to trace, even with a conscious and targeted effort by the LLM vendor. When evaluating a closed-source LLM, users often feed the model with test-set samples (with or without labels) surrounded by additional text, such as instructions in the form of prompts. In some cases, especially when evaluating the LLM robustness, the test-set samples are perturbed and hence no longer an exact match of their original version. Therefore, it is unlikely that LLM vendors could effectively exclude leaked benchmarks from further model fine-tuning, especially at scale. 
For (2), it would be necessary to understand how the LLM vendor uses the data to improve the model. A very likely scenario is continued pre-training, where the data leaked by users is treated as an in-domain corpus (and thus given more influence than pretraining data).
This procedure is known to improve models' performances in the leaked domains~\citep{gururangan-etal-2020-dont}. 
Notably, \citet{shi2023dont} find that fine-tuning a model on in-domain text enriched by textual instructions leads to an increase in the model performance even if gold labels are not shown to the model.
This setup perfectly matches the kind of data shown to chat LLMs when evaluated by researchers.
This means that closed-source LLMs such as GPT-3.5 and GPT-4 can make use of these gold standard examples from widely used NLP benchmarks to gain an unfair advantage over other models. 

We also point out that recent work~\citep{aiyappa-etal-2023-trust} showed that after model updates, ChatGPT performance improved on benchmarks to which it was previously exposed~\citep{zhang2022would}. With these motivations, we conduct a systematic review to quantify how much of such data the models powering ChatGPT could have obtained.

\section{Methodology}
\label{sec:methods}

Following the standard systematic review protocol from the medical domain~\citep{khan2003five}, we analyse the existing work on LLMs evaluation to inspect the issue of indirect data contamination and other evaluation malpractices. We focus on OpenAI's GPT-3.5 and GPT-4 models, as they are the most prominently used in recent NLP research. We organize our work into five macro-steps, corresponding to the following subsections.

\subsection{Framing questions} In reviewing the existing work evaluating the performace of GPT-3.5 and GPT-4, we pose the following research questions:
\begin{enumerate}[label={(\arabic*)}]
    \item Which datasets have been demonstrably leaked to GPT-3.5 and GPT-4 during the last year?
    \item Do all papers evaluating these models include a fair comparison with existing baselines?
\end{enumerate}

\subsection{Identifying relevant work}
\label{sec:identifying}
We employ commonly used online databases\footnote{We query \href{https://scholar.google.com/}{Google Scholar}, \href{https://www.semanticscholar.org/}{Semantic Scholar}, \href{https://dblp.org/}{DBLP}, \href{https://arxiv.org/}{arXiV}, \href{https://aclanthology.org/}{ACL Anthology}.} and major NLP conferences proceedings (including ACL, NAACL, EMNLP, NeurIPS), considering both peer-reviewed work and pre-prints, as the interaction with LLMs happened regardless of publication status.
We filter our queries on work containing the terms ``ChatGPT'', ``GPT-4'', ``GPT-3.5'' ``OpenAI'' ``evaluation'', ``large language models'', ``AI'' either in title, abstract, body, or all of them.

We also do not limit our search to computer science works only, as recent LLMs have been investigated by researchers from many other domains, e.g. healthcare~\citep{kung2023performance}, psychology~\citep{cai2023does} and education~\citep{szefer2023analyzing}. Since the ChatGPT models are our primary focus, we limit our search to works between late November 2022 (when the first model was publicly released) and early October 2023. Among all the papers, we first do a preliminary screening, assessing if they effectively run GPT-3.5 or GPT-4 in any form.\footnote{We encountered a small number of papers also comparing to other closed-source LLMs, such as Anthropic's Claude.}

\subsection{Assessing quality and relevance} 
To assess which work effectively leaked data to ChatGPT, we refer to OpenAI's data usage policy,\footnote{\url{https://help.openai.com/en/articles/5722486-how-your-data-is-used-to-improve-model-performance}}, which explicitly mentions the use of users' data for model training:

\begin{quote}
    "[...] when you use our services for individuals such as ChatGPT or DALL-E, we may use your content to train our models [...]" 
\end{quote}
It also clarifies that the user data are not used for model training if sent via API and business services:

\begin{quote}
    "[...] we don’t use content from our business offerings [...] and our API Platform to train our models [...]" 
\end{quote}

Therefore, only the work interacting with the models through the web interface\footnote{\href{https://chat.openai.com/}{https://chat.openai.com/}} is considered to leak data. We note that while it is possible to opt out of providing the data for model improvement purposes,\textsuperscript{\ref{fn:optout}} we found no evidence suggesting any of the surveyed papers~did~so.

A small number of works used both the web interface and API access.\footnote{Their experiments began prior to March 1st, 2023 and the authors started using the API soon after it was released.}
We carefully review such works to calculate which portion of the data was used in the former setup.
We drew our conclusions from the paper draft history on arXiv; in some cases, this information was also transparently disclosed by the authors.
In the case of work with multiple drafts dating before the model release in November 2022, we consider the earliest draft that includes GPT-3.5 or GPT-4 for the calculation. 

\subsection{Summarizing the evidence} We inspect each surveyed paper, looking for information on the used datasets, split, and number of samples. If no mention of sampling or similar information is made, we assume that the whole dataset has been used. Similarly, if no information on the used split is provided, we assume that the authors treated the dataset as a whole. It could be argued that feeding entire datasets to ChatGPT is unrealistic because of the usage restrictions imposed by OpenAI on the web interface, and the amount of work necessary for manually inputting the data inside the chat. However, we note that quickly after ChatGPT release, many unofficial wrappers have been developed\footnote{E.g. \href{https://github.com/acheong08/ChatGPT}{revChatGPT}, \href{https://github.com/rawandahmad698/PyChatGPT}{
PyChatGPT}, and \href{https://github.com/acheong08/ChatGPT-to-API}{ChatGPT-to-API}.} for circumventing said issues, most of which are still in active use. We also point out that many of the papers we surveyed mentioned the use of such tools explicitly.

We also track secondary information relevant to the evaluation -- for each work, we inspect: (1) if it has been peer-reviewed;\footnote{We do note that part of the work we reviewed might still be under review, also see Footnote \ref{fn:review}.} (2) if the used prompts are available; (3) if a repository to reproduce the experiment is provided; (4) if the authors used a whole dataset or a sample; (5) if GPT-3.5 or GPT-4 were compared to other open models/approaches and if the evaluation scale was the same; (6) if the version of the model used is reported.

\subsection{Interpreting the findings} We report the results of our review both quantitatively and qualitatively. Specifically, we report the number of works surveyed leaking data to GPT-3.5 or GPT-4 in such a way that it can be used by OpenAI to further improve the model (according to their data policy). In this paper we do not distinguish between works leaking data to GPT-3.5, GPT-4, or both. This is because indirect data leaking is caused by browser access, where both models are available through the ChatGPT Plus subscription. We also note that OpenAI confirmed that creating GPT-4 involved the use of ChatGPT to some extent.\footnote{https://openai.com/research/gpt-4} For this reason, we estimate the data leakage to be effectively shared across the two models and for simplicity, we refer to both models as “ChatGPT” from now on.

We also document a series of evaluation practices emerging for the work reviewed that is problematic with respect to objectiveness and reproducibility. Finally, drawing upon our results, we present a series of best practices for researchers evaluating OpenAI's and other closed-source LLMs.  

\section{Results}
\label{sec:results}

Following our methodology, in the first step we identified 255 research papers, 212 of which were found relevant\footnote{The excluded papers either were opinion pieces that minimally tested ChatGPT on certain tasks, or did not include any evaluation. } during the initial screening (see Sec.~\ref{sec:identifying}).
Among the relevant papers, 70 ($\sim32\%$) were peer-reviewed, 
while the remainder (142) consisted of pre-prints.\footnote{We note that, during this paper's review period, 43 of the pre-prints were peer-reviewed and published. However, some of the relevant proceedings have not been released yet, making it impossible to consistently check for paper updates. We cannot rule out that some of these works leaked more data with further experiments, or addressed some evaluation malpractices.\label{fn:review}}
We subsequently analysed the retrieved papers to examine the problem of data contamination and the adopted evaluation practices.

\subsection{Indirect data contamination}
\label{ssec:contamination}

From our analysis, 90 papers ($\sim42\%$) accessed ChatGPT through the web interface, hence providing data that OpenAI could have used to further improve its models.

\begin{figure}[t]
    \centering
    \hspace{-4mm}
    \includegraphics[trim=7 0 7 0, clip, scale=0.26]{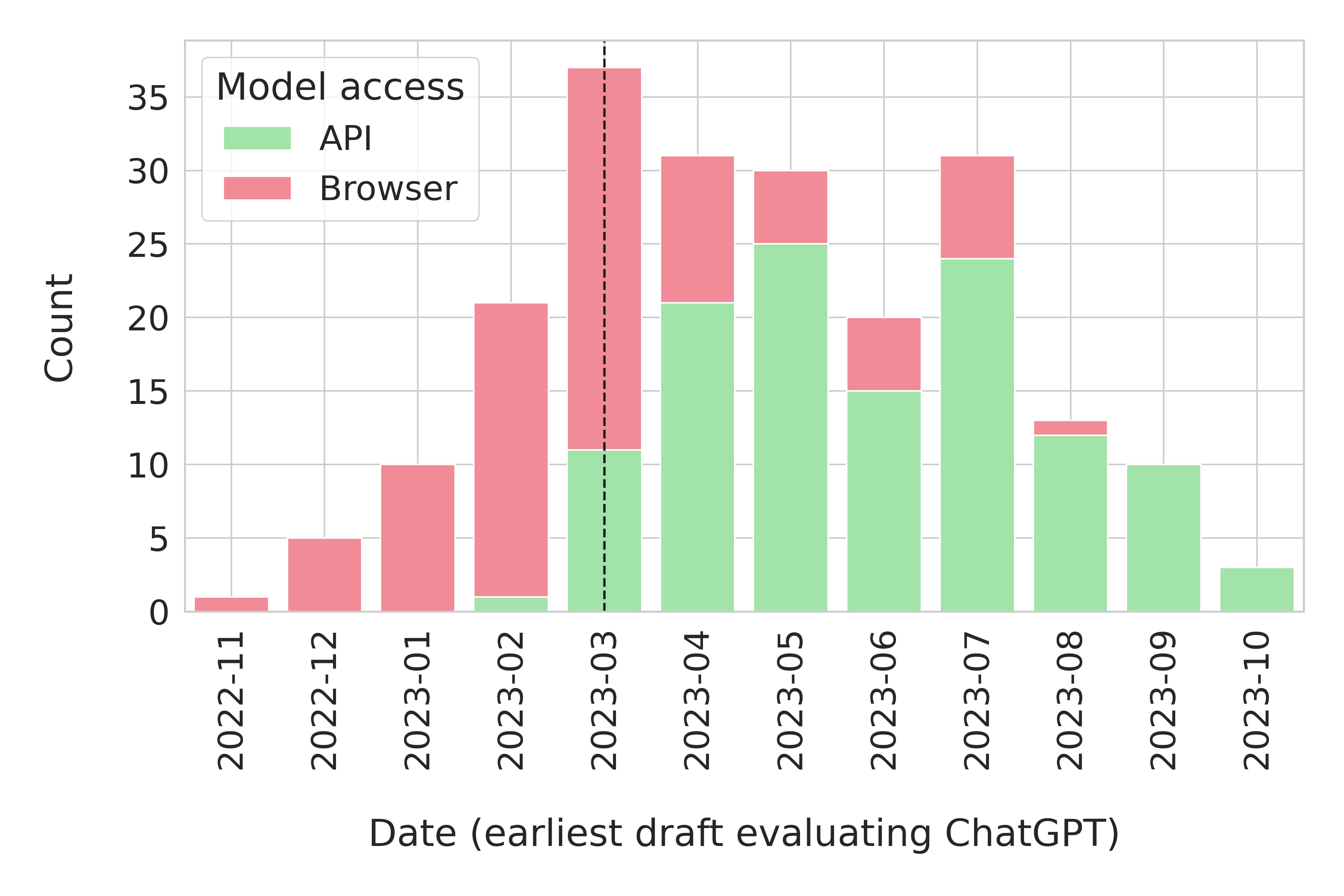}
    \caption{Distribution of the dates when papers evaluating ChatGPT were first uploaded to arXiv or published. The dotted line represents the ChatGPT API release (March 1st, 2023, dotted line in the chart) as a cutoff point. The single paper shown using the API in February is by a research group that reported having early API access.}
    \label{fig:date-distribution}
\end{figure}

We first inspected the time distribution of the reviewed works (\Cref{fig:date-distribution}) to gain insight into when most data leaks happened. Unsurprisingly, the majority of the papers leaking data dates before the official release of ChatGPT API, and it can be seen that web interface access rapidly decreased following March 2023. However, we must note that (1) a considerable amount of work kept using the web interface to access ChatGPT until September 2023 and (2) our analysis cannot inspect the preliminary stages of prompt engineering, which are rarely reported and might still be done through the web interface because of its trial-and-error nature.

The presence of leaked data after the API release may indicate that a part of the research community is either unaware of OpenAI's data policy, or does not consider it a problem when conducting experiments.
Many works, especially small case studies, also reported using the web interface for cost reasons, as it allows free access to the models.

As a second step, we quantified leak severity per dataset and split. For work specifying the amount of data used (either in the paper or through a repository), we consider the given value. For the rest, we calculate it by inspecting the actual dataset.\footnote{We mainly use \href{https://huggingface.co/datasets}{HuggingFace Datasets}, but also refer to \href{https://www.kaggle.com/}{Kaggle} or other sources based on availability.} In seven papers, no number of samples used was specified, so we contacted the authors for clarification. In the two cases where the authors did not respond, we assumed the entire split of a dataset was used. 
We calculated both the number of instances and the percentage of the considered split (or the whole dataset when applicable).  

\begin{figure}
    \centering
    \hspace{-5mm}
    \includegraphics[trim = 25 0 25 0, clip, scale=0.4]{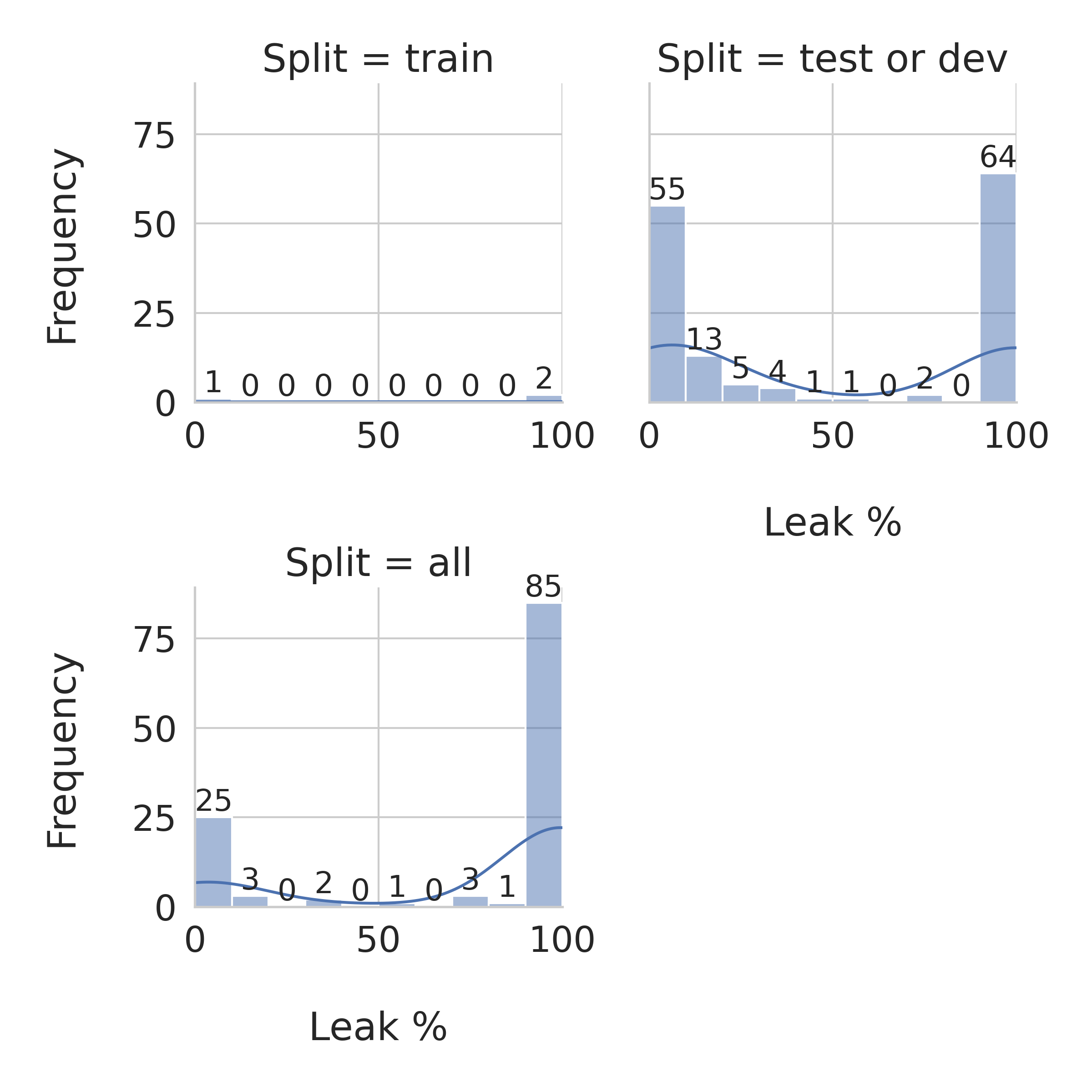}
    \caption{Data leakage distribution. We report the number of times (y) we observed a specific percentage of leaking (x) for the considered split. As some work vaguely describes the used split as ``test or dev set'', we merge these two values in a unique chart.}
    \label{fig:leak_distribution}
\end{figure}

Since a small number of datasets (18) was used in multiple papers in different amounts, we had to consider whether these should be interpreted as individual separate leaks (that should be summed up) or not. We were not able to verify this from the provided data, so we adopted an “optimistic” approach and assumed that the largest leak for a given dataset is always a superset of all smaller ones.\footnote{We also tried a pessimistic approach, where we assumed all the leaks were independent, but due to the small number of works covering the same data, the results are virtually identical.}

Our calculations show that the 90 papers leaked data from 263 unique datasets, for a total of over 4.7M samples (see \Cref{tab:data-leak_1,tab:data-leak_2,tab:data-leak_3} in the Appendix).\footnote{The survey total is 4,714,753 leaked samples.} 

We find most samples ($\sim93.8\%$) coming from datasets treated as whole (with no split), followed by test and development ($\sim5.6\%$),\footnote{As some work vaguely describes the used split as ``test or dev set'', we merge these two values.} and training ($\sim0.6\%$) sets. In line with what we discussed in \Cref{sec:indirect-leak}, we can conclude that ChatGPT was exposed to millions of benchmark samples, enriched with instructions that could be considered de-facto novel gold-standard data in some cases.

We also report that several works included the examples' labels when few-shot prompting ChatGPT or using it as a reference-based evaluation metric. We consider this the worst possible case of data leaking, as it gives the model information about the desired output as well.

To classify leak severity, we examine the frequency distribution of leak sizes (\Cref{fig:leak_distribution}). It appears that most works either leak full splits or very small samples, with only a few works leaking intermediate amounts. With this information, we classify a portion of leaked data as \emph{low} ($< 5\%$), \emph{moderate-low} ($5- 50\%$), \emph{moderate-high} ($50- 95\%$), or \emph{high} ($> 95\%$).

Consequently, we categorize all leaked datasets into these 4 thresholds. Overall, we find a low leak for 66 ($\sim25\%$) datasets, moderate-low for 47 ($\sim18\%$), moderate-high for 10 ($\sim4\%$) and high for 142 ($\sim53\%$). This result is particularly worrying as the majority of datasets were almost completely leaked.

\begin{table}[t]
    \centering
    \small
    \begin{tabular}{lrrrr}
\toprule
       Task name &Lo&M-Lo&M-Hi&Hi
    \\\midrule
AI safety \& ethics&0&0&2&0
\\
Creative NLG&1&0&0&0
\\
Dialogue&2&1&0&5
\\
NLG evaluation&0&0&0&4
\\
Machine Translation&6&4&1&1
\\
Math&0&1&0&8
\\
Natural language generation&2&1&0&14
\\
Natural language inference&6&2&0&15
\\
Language understanding&0&0&0&2
\\
Paraphrasing&2&0&0&0
\\
Politics&0&1&0&3
\\
Programming&0&0&0&1
\\
Psychology&0&0&0&1
\\
Question answering&24&14&5&31
\\
Commonsense reasoning\hspace{-5mm}&3&4&0&9
\\
Semantic similarity&2&1&0&3
\\
Sentiment analysis&8&9&1&8
\\
Summarization&5&6&1&1
\\
Text classification&1&0&0&3
\\
Text extraction&2&1&0&7

\\\bottomrule
    \end{tabular}
    \caption{
    The number of datasets with low (Lo), moderate-low (M-Lo), moderate-high (M-Hi) and high leak severity (Hi) is reported for each task, omitting custom datasets. A more detailed table, including specific dataset names, is provided in the Appendix~\ref{sec:appendix-leak}.
    }
    \label{tab:task_leak_threshold}
\end{table}

\begin{figure*}[!ht]
    \centering
    \includegraphics[trim = 10 20 10 0, clip, scale=0.17]
    {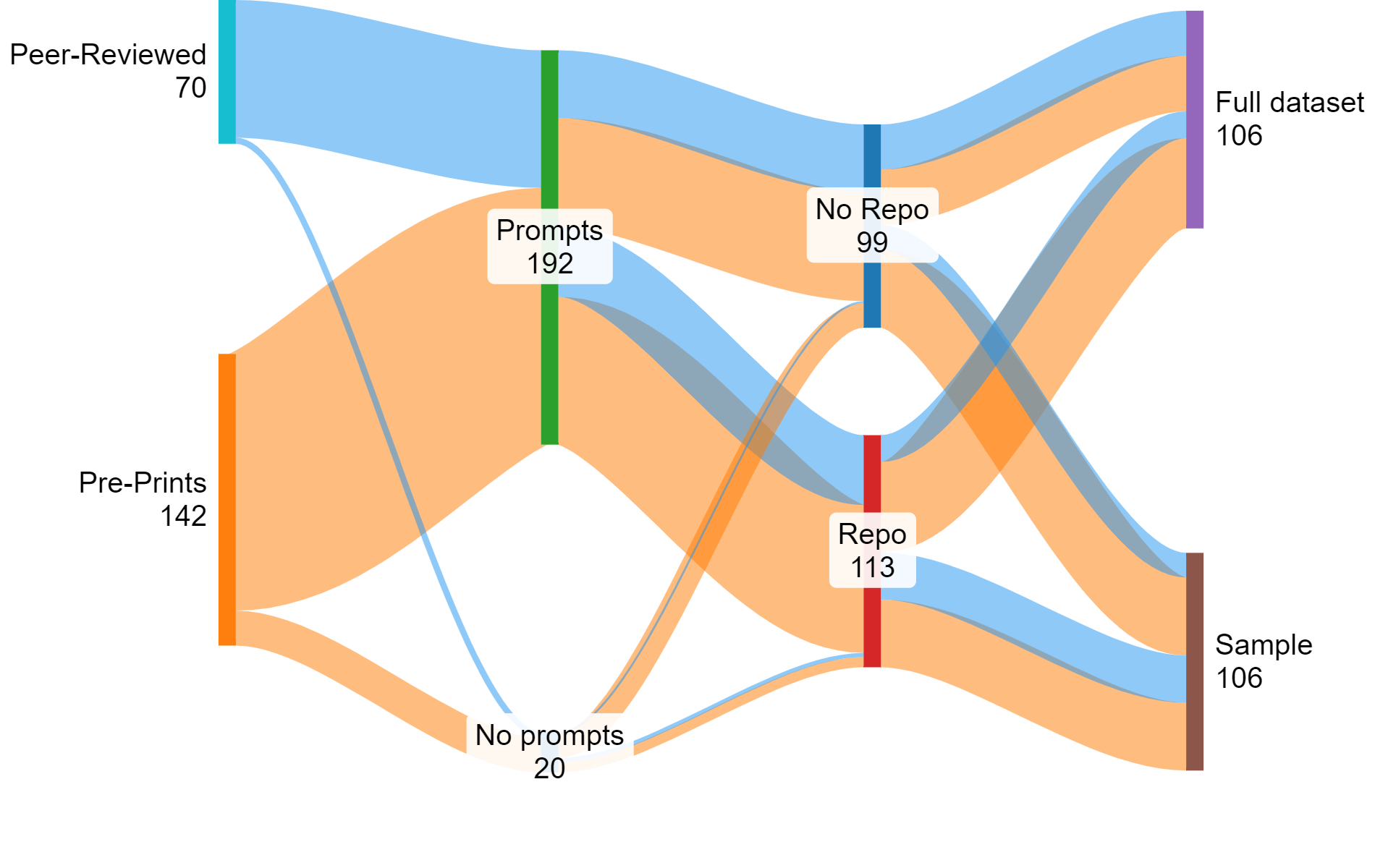}
    \caption{Evaluation reproducibility. Through the above Sankey diagram, we report facilitators and barriers to reproducing the carried-out experiments. This includes providing the used prompts, a repository with usable code and the use of sampling.}
    \label{fig:sankey_reprod}
    
\end{figure*}

Finally, we inspect which NLP tasks are covered by the leaked data (\Cref{tab:task_leak_threshold}). We find that the tasks suffering the most from high leaks are natural language inference, question answering, and natural language generation.
These and other tasks include many highly popular NLP benchmarks, as well as high-quality custom datasets created ad-hoc for individual evaluations
(see \Cref{tab:data-leak_1,tab:data-leak_2,tab:data-leak_3} in the Appendix). To name a few, almost the entire test sets from Semeval2016 Task 6~\citep{mohammad-etal-2016-semeval}, SAMSum~\citep{gliwa-etal-2019-samsum}, and MultiWOZ 2.4~\citep{ye-etal-2022-multiwoz} are leaked.
The custom datasets were frequently phrased as an exam in a field different from NLP, e.g., medicine, physics, psychology, or law. Other custom datasets explored, for example, the LLMs' sense of humour, philosophical and political leaning, or bias. We note that not all the leaked custom datasets have been publicly released. This makes the leak even more severe, as it potentially makes OpenAI the only organisation (besides the authors) with access to such data.

\subsection{Reproducibility}
\label{ssec:reproducibility}

We assess the evaluations' reproducibility by checking whether the prompts used to query ChatGPT were provided, whether a repository containing data or code was available, and whether the datasets used were custom-made.
Finally, we also check for sampling of the original data or other practices that make it impossible to exactly reconstruct the data used.

From our results (\Cref{fig:sankey_reprod}), 192 ($\sim 91\%$) works report the prompts used to convert data into a query and possibly to instruct the model on how to perform a given task. The number of works providing a code repository is significantly smaller, at 113 ($\sim 53\%$). This figure excludes papers that provided a link to a non-existent or empty repository. 
Overall, 72 ($\sim51\%$) of the pre-prints and 34 ($\sim48\%$) peer-reviewed papers provided both prompts and a repository. We report further details on this data in \Cref{sec:appendix-malpractices}.

Another barrier to reproducibility is that most closed-source LLMs are being regularly updated. Therefore, it is crucial to report the used model version, as different versions may lead to significantly different outputs~\citep{chen2023chatgpt}.
In the surveyed works, this was generally done by reporting the running period of the experiments when using the web interface, or by reporting which version of the model has been accessed via the API.
Unfortunately, as regular model updates are a relatively new concept, this practice is not yet common.
Only 29 (40\%) of the peer-reviewed papers and 33 (23\%) of the pre-prints provide this information.

\begin{figure}[t]
    \centering
    \includegraphics[trim = 16 140 20 0, clip, scale=0.12]{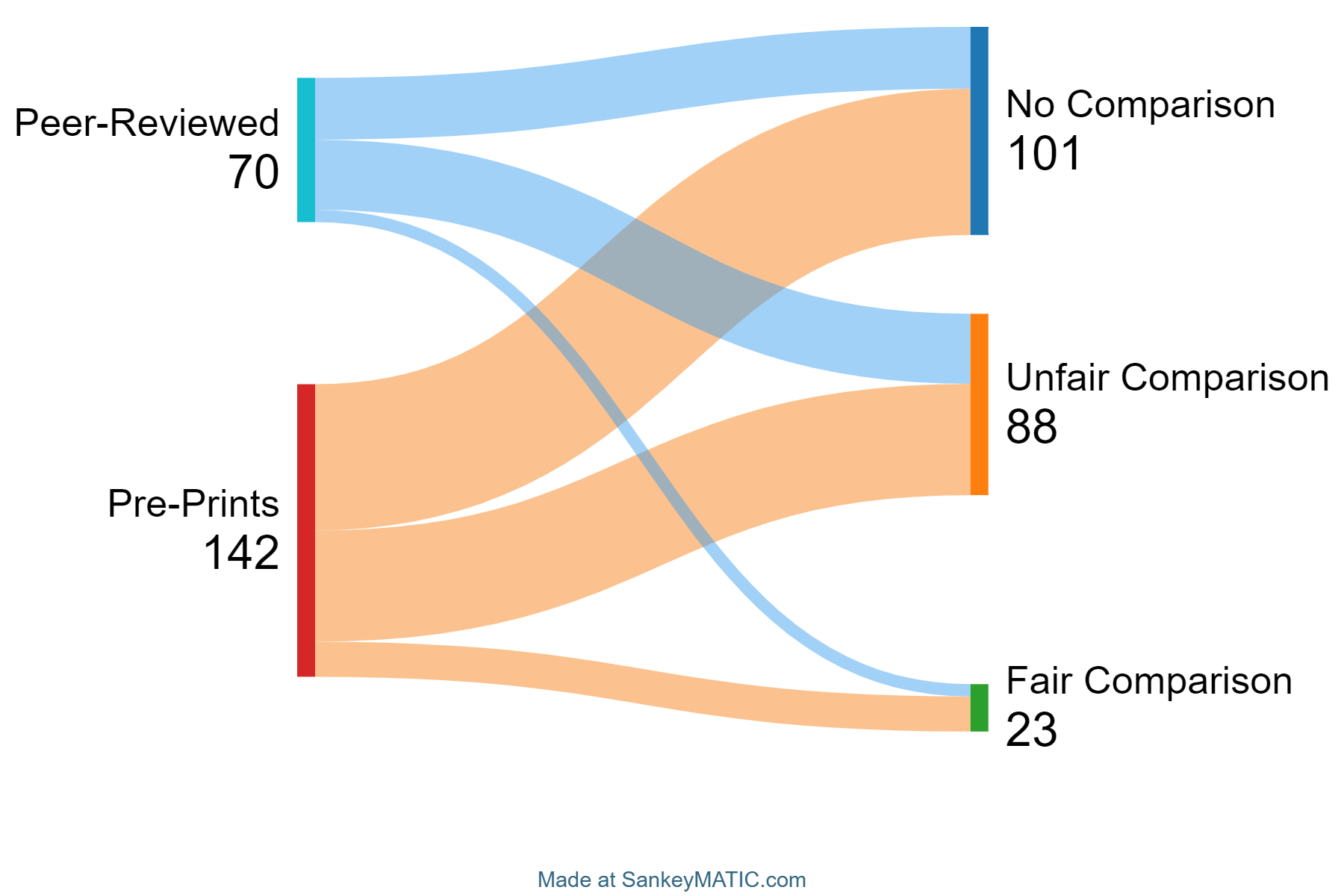}
    \caption{Evaluation fairness. Through the above Sankey diagram, we report whether the proprietary LLMs were compared against other models, and if the comparison was equal. In this context, "Unfair" comparison refers to evaluating different models on different amounts of data.}
    \label{fig:sankey_fair}
\end{figure}

\subsection{Evaluation fairness}
\label{ssec:bad-eval}
We find the evaluation of ChatGPT's performance to be often unfair. First, comparison to any open-source LLM or non-LLM-based method may be missing. Our results (\Cref{fig:sankey_fair}) show that this is similarly prevalent regardless of the publication status, appearing in 71 ($\sim50\%$) of pre-prints and 30 ($\sim43\%$) of published papers.
Second, when a comparison with open models and baselines is made, 54 pre-prints ($\sim 38\%$) and 34 peer-reviewed ($\sim 49\%$) papers compare the results computed on different samples. ChatGPT is typically evaluated on a random sample of the benchmark while other models are compared on its entirety. In many works, ChatGPT's performance is measured on only a handful (10-50) of examples, which substantially lowers the expressive power of the comparison. For instance, considering a simplistic case with binary assessment of model output (correct/incorrect) on 10 examples, the difference should be more than 30\% to be statistically significant,\footnote{Assuming Fisher's exact test, typical $\alpha=5\%$ and moderate model performance around $\hat{p}=0.5$} which is rarely seen. Statistical analysis of results is almost never performed. We report further details on evaluation fairness in \Cref{sec:appendix-malpractices}.

Another concerning practice is how the size of the evaluation data is reported, especially when sampling is used. We find that papers often show the size of the whole evaluation dataset upfront (e.g. in a table or in the dataset description section), but they report the actual sample sizes used for evaluation only later and in a less obvious way (in footnotes, limitations sections, or appendices).
This practice makes the experimental results harder to interpret.

\section{Suggested Practices in Closed-source LLM Evaluation}
\label{sec:best-practices}
Our survey revealed both a significant amount of data leakage in ChatGPT and many worrying trends in its evaluation. In light of this, we list a series of suggested practices that we believe could help mitigate the issues. We believe that researchers looking to objectively evaluate LLMs today should:

\paragraph{Access the model in a way that does not leak data}
The first step when planning proprietary LLMs evaluation should be reading their most up-to-date data policies, and access models accordingly (e.g. API instead of web interface for OpenAI's LLMs). We also acknowledge that in some cases this might not be viable due to budget limits, or an overly steep learning curve for the use of APIs by researchers outside of computer science.\footnote{In such case, as of January 2024, OpenAI allows users to opt out of providing data for model improvement through the \href{https://privacy.openai.com/policies}{OpenAI Privacy Request Portal}.\label{fn:optout}}

\paragraph{Interpret performance with caution} The lack of system specifications and training details can make proprietary LLMs look like incredibly powerful tools with impressive zero-shot performance. This can often be explained by data contamination~\citep{aiyappa-etal-2023-trust}. In our review, we documented that over 4 million samples across more than 200 NLP datasets have been leaked to these models. The performance of closed-source LLMs should always be interpreted while keeping these results 
in mind.

\paragraph{When possible, avoid using closed-source models} We strongly encourage using the available open-source LLMs. While there has been discussion in the research community about proprietary models being consistently better than open-source ones, we note that (1) this is often driven by hype, while there is evidence of the opposite~\citep{chatpgt-jack-trades-master-none}, (2) research done solely on closed LLMs limits scientific progress, bringing benefits mainly to the LLM vendors and (3) LLM vendors can arbitrarily make changes to the models, e.g., making previous versions unavailable, changing their behaviour in a way that may not be visible to the user~\citep{chen2023chatgpt} or changing the data treatment policy. 

\paragraph{Adopt a fair and objective comparison} Evaluating closed-source LLMs is tied to comparing them with pre-existing approaches. Evaluating proprietary models on a limited number of samples while evaluating open ones on dramatically larger sets is scientifically dubious at best. When sampling is required (for example because of budgetary restrictions), it should be applied to all the considered approaches. We also discourage taking state-of-the-art values directly from previous work and suggest to re-run all approaches on the considered data only. 

\paragraph{Make the evaluation reproducible} In light of the known NLP evaluation reproducibility crisis~\citep{belz-etal-2023-non,10.1162/coli_a_00508} we strongly encourage researchers to report as many details about their setup. Besides all the relevant details about the setup for reproducibility, such as random seeds, open model parameters, etc., we note that when the evaluation involves closed models, additional details should be disclosed.
Prompts, as well as the process leading to them, should be detailed since LLMs are very sensitive to even minor changes in prompts~\citep{lu2022fantastically}. The model version and experiment running period should be mentioned as well so that further researchers can use the same model checkpoint if possible. Data, especially if sampled, should be released (ideally in a repository) to avoid potential differences in sampling. 

\paragraph{Report indirect data leaking}
Indirect data leaking is a serious issue, and when it happens it should be reported. Clear information on which benchmarks have been leaked benefits research, helps other researchers orient their experiments, and ultimately leads to a more objective evaluation of proprietary LLMs. We invite all researchers to contribute to our collaborative project at \url{https://leak-llm.github.io/}.

\section{Conclusion and Future Work}
\label{sec:conclusion}

In this work, we present our findings based on the analysis of 255 papers evaluating the performance of GPT-3.5 and GPT-4. We investigate the problem of indirect data contamination and report that 4.7M samples coming from 263 distinct datasets have been exposed to the models in such a way that this data could be used for training by OpenAI. We also report concerning research practices with respect to reproducibility and fairness. Finally, informed by our analysis, we detailed some suggested practices for the evaluation of closed-source LLMs.

\paragraph{Future Work} 
In our future work, we aim to run experiments via the OpenAI API to see the impact of leaked test data on the performance of GPT-3.5 and GPT-4 on the leaked datasets and the tasks in general.

Furthermore, we consider investigating indirect data leakage in other closed-source models, namely from Anthropic or Cohere, which appeared in a small number of papers reviewed in this work. 

\section*{Limitations}
\label{sec:limitations}

We are aware the list of contaminated datasets we compiled in our work is not fully conclusive for one of several reasons:
\begin{enumerate}[label={(\arabic*)}]
    \item We review the information that has been publicly revealed via articles. We postulate more experiments could have revealed test set data to closed-source models but were never published.
    \item In this paper, we focus on the works that use ChatGPT or GPT-4. However, prior to March 1st, 2023, OpenAI's policy stated that they may also use data from the API to improve their models. This would imply that data sent to GPT-3 via the API could have been used for training.
    \item The number of papers investigating the performance of ChatGPT is vast, and despite our best efforts, we could have missed some works.
    \item Information on whether individual works are pre-prints or published is given at the time of writing (early October 2023). This is subject to change, especially given the freshness of many of the works reviewed.
    \item Many datasets released prior to 2021 could have been fully leaked by being a part of the models' pre-training data.
\end{enumerate}

As mentioned in Section \ref{sec:methods}, in some cases the papers were not clear about some aspects of the experiments.
We contacted the authors of such papers for clarification, however, two of them did not respond. Therefore, our best-judgment assumptions may be wrong for these papers.

\section*{Acknowledgements}
This research was supported by the European Research Council (Grant agreement No.~101039303 NG-NLG) and by Charles University project SVV~260~698. Patrícia Schmidtová was also supported by the Women in Quant Finance Network grant awarded by G-Research. 
We would also like to thank Zdeněk Kasner and Dominik Macháček for their valuable feedback on the manuscript.


\bibliography{anthology,custom}

\begin{thebibliography}{30}
\expandafter\ifx\csname natexlab\endcsname\relax\def\natexlab#1{#1}\fi

\bibitem[{Aiyappa et~al.(2023)Aiyappa, An, Kwak, and Ahn}]{aiyappa-etal-2023-trust}
Rachith Aiyappa, Jisun An, Haewoon Kwak, and Yong-yeol Ahn. 2023.
\newblock \href {https://doi.org/10.18653/v1/2023.trustnlp-1.5} {Can we trust the evaluation on {C}hat{GPT}?}
\newblock In \emph{Proceedings of the 3rd Workshop on Trustworthy Natural Language Processing (TrustNLP 2023)}, pages 47--54, Toronto, Canada. Association for Computational Linguistics.

\bibitem[{Belz et~al.(2023)Belz, Thomson, Reiter, and Mille}]{belz-etal-2023-non}
Anya Belz, Craig Thomson, Ehud Reiter, and Simon Mille. 2023.
\newblock \href {https://doi.org/10.18653/v1/2023.findings-acl.226} {Non-repeatable experiments and non-reproducible results: The reproducibility crisis in human evaluation in {NLP}}.
\newblock In \emph{Findings of the Association for Computational Linguistics: ACL 2023}, pages 3676--3687, Toronto, Canada. Association for Computational Linguistics.

\bibitem[{Brown et~al.(2020)Brown, Mann, Ryder, Subbiah, Kaplan, Dhariwal, Neelakantan, Shyam, Sastry, Askell, Agarwal, Herbert-Voss, Krueger, Henighan, Child, Ramesh, Ziegler, Wu, Winter, Hesse, Chen, Sigler, Litwin, Gray, Chess, Clark, Berner, McCandlish, Radford, Sutskever, and Amodei}]{GPT3-original}
Tom Brown, Benjamin Mann, Nick Ryder, Melanie Subbiah, Jared~D Kaplan, Prafulla Dhariwal, Arvind Neelakantan, Pranav Shyam, Girish Sastry, Amanda Askell, Sandhini Agarwal, Ariel Herbert-Voss, Gretchen Krueger, Tom Henighan, Rewon Child, Aditya Ramesh, Daniel Ziegler, Jeffrey Wu, Clemens Winter, Chris Hesse, Mark Chen, Eric Sigler, Mateusz Litwin, Scott Gray, Benjamin Chess, Jack Clark, Christopher Berner, Sam McCandlish, Alec Radford, Ilya Sutskever, and Dario Amodei. 2020.
\newblock \href {https://proceedings.neurips.cc/paper_files/paper/2020/file/1457c0d6bfcb4967418bfb8ac142f64a-Paper.pdf} {Language models are few-shot learners}.
\newblock In \emph{Advances in Neural Information Processing Systems}, volume~33, pages 1877--1901. Curran Associates, Inc.

\bibitem[{Cai et~al.(2023)Cai, Haslett, Duan, Wang, and Pickering}]{cai2023does}
Zhenguang~G Cai, David~A Haslett, Xufeng Duan, Shuqi Wang, and Martin~J Pickering. 2023.
\newblock \href {https://arxiv.org/abs/2303.08014} {Does {ChatGPT} resemble humans in language use?}
\newblock \emph{arXiv preprint arXiv:2303.08014}.

\bibitem[{Chen et~al.(2023)Chen, Zaharia, and Zou}]{chen2023chatgpt}
Lingjiao Chen, Matei Zaharia, and James Zou. 2023.
\newblock \href {https://arxiv.org/abs/2307.09009} {How is {ChatGPT}'s behavior changing over time?}
\newblock \emph{arXiv preprint arXiv:2307.09009}.

\bibitem[{Chowdhery et~al.(2022)Chowdhery, Narang, Devlin, Bosma, Mishra, Roberts, Barham, Chung, Sutton, Gehrmann, Schuh, Shi, Tsvyashchenko, Maynez, Rao, Barnes, Tay, Shazeer, Prabhakaran, Reif, Du, Hutchinson, Pope, Bradbury, Austin, Isard, Gur-Ari, Yin, Duke, Levskaya, Ghemawat, Dev, Michalewski, Garcia, Misra, Robinson, Fedus, Zhou, Ippolito, Luan, Lim, Zoph, Spiridonov, Sepassi, Dohan, Agrawal, Omernick, Dai, Pillai, Pellat, Lewkowycz, Moreira, Child, Polozov, Lee, Zhou, Wang, Saeta, Diaz, Firat, Catasta, Wei, Meier-Hellstern, Eck, Dean, Petrov, and Fiedel}]{chowdhery_palm_2022}
Aakanksha Chowdhery, Sharan Narang, Jacob Devlin, Maarten Bosma, Gaurav Mishra, Adam Roberts, Paul Barham, Hyung~Won Chung, Charles Sutton, Sebastian Gehrmann, Parker Schuh, Kensen Shi, Sasha Tsvyashchenko, Joshua Maynez, Abhishek Rao, Parker Barnes, Yi~Tay, Noam Shazeer, Vinodkumar Prabhakaran, Emily Reif, Nan Du, Ben Hutchinson, Reiner Pope, James Bradbury, Jacob Austin, Michael Isard, Guy Gur-Ari, Pengcheng Yin, Toju Duke, Anselm Levskaya, Sanjay Ghemawat, Sunipa Dev, Henryk Michalewski, Xavier Garcia, Vedant Misra, Kevin Robinson, Liam Fedus, Denny Zhou, Daphne Ippolito, David Luan, Hyeontaek Lim, Barret Zoph, Alexander Spiridonov, Ryan Sepassi, David Dohan, Shivani Agrawal, Mark Omernick, Andrew~M. Dai, Thanumalayan~Sankaranarayana Pillai, Marie Pellat, Aitor Lewkowycz, Erica Moreira, Rewon Child, Oleksandr Polozov, Katherine Lee, Zongwei Zhou, Xuezhi Wang, Brennan Saeta, Mark Diaz, Orhan Firat, Michele Catasta, Jason Wei, Kathy Meier-Hellstern, Douglas Eck, Jeff Dean, Slav Petrov, and Noah Fiedel. 2022.
\newblock \href {http://arxiv.org/abs/2204.02311} {{PaLM}: {Scaling} {Language} {Modeling} with {Pathways}}.
\newblock \emph{arXiv:2204.02311 [cs]}.
\newblock ArXiv: 2204.02311.

\bibitem[{GitHub(2022)}]{copilot}
GitHub. 2022.
\newblock About github copilot.
\newblock \url{https://github.com/features/copilot}.

\bibitem[{Gliwa et~al.(2019)Gliwa, Mochol, Biesek, and Wawer}]{gliwa-etal-2019-samsum}
Bogdan Gliwa, Iwona Mochol, Maciej Biesek, and Aleksander Wawer. 2019.
\newblock \href {https://doi.org/10.18653/v1/D19-5409} {{SAMS}um corpus: A human-annotated dialogue dataset for abstractive summarization}.
\newblock In \emph{Proceedings of the 2nd Workshop on New Frontiers in Summarization}, pages 70--79, Hong Kong, China. Association for Computational Linguistics.

\bibitem[{Golchin and Surdeanu(2023)}]{golchin2023time}
Shahriar Golchin and Mihai Surdeanu. 2023.
\newblock \href {https://arxiv.org/abs/2308.08493} {Time travel in {LLMs}: Tracing data contamination in large language models}.
\newblock \emph{arXiv preprint arXiv:2308.08493}.

\bibitem[{Gururangan et~al.(2020)Gururangan, Marasovi{\'c}, Swayamdipta, Lo, Beltagy, Downey, and Smith}]{gururangan-etal-2020-dont}
Suchin Gururangan, Ana Marasovi{\'c}, Swabha Swayamdipta, Kyle Lo, Iz~Beltagy, Doug Downey, and Noah~A. Smith. 2020.
\newblock \href {https://doi.org/10.18653/v1/2020.acl-main.740} {Don{'}t stop pretraining: Adapt language models to domains and tasks}.
\newblock In \emph{Proceedings of the 58th Annual Meeting of the Association for Computational Linguistics}, pages 8342--8360, Online. Association for Computational Linguistics.

\bibitem[{Ippolito et~al.(2023)Ippolito, Tramer, Nasr, Zhang, Jagielski, Lee, Choquette~Choo, and Carlini}]{ippolito-etal-2023-preventing}
Daphne Ippolito, Florian Tramer, Milad Nasr, Chiyuan Zhang, Matthew Jagielski, Katherine Lee, Christopher Choquette~Choo, and Nicholas Carlini. 2023.
\newblock \href {https://aclanthology.org/2023.inlg-main.3} {Preventing generation of verbatim memorization in language models gives a false sense of privacy}.
\newblock In \emph{Proceedings of the 16th International Natural Language Generation Conference}, pages 28--53, Prague, Czechia. Association for Computational Linguistics.

\bibitem[{Khan et~al.(2003)Khan, Kunz, Kleijnen, and Antes}]{khan2003five}
Khalid~S Khan, Regina Kunz, Jos Kleijnen, and Gerd Antes. 2003.
\newblock \href {https://doi.org/10.1177/014107680309600304} {Five steps to conducting a systematic review}.
\newblock \emph{Journal of the Royal Society of Medicine}, 96(3):118--121.

\bibitem[{Kocoń et~al.(2023)Kocoń, Cichecki, Kaszyca, Kochanek, Szydło, Baran, Bielaniewicz, Gruza, Janz, Kanclerz, Kocoń, Koptyra, Mieleszczenko-Kowszewicz, Miłkowski, Oleksy, Piasecki, Łukasz Radliński, Wojtasik, Woźniak, and Kazienko}]{chatpgt-jack-trades-master-none}
Jan Kocoń, Igor Cichecki, Oliwier Kaszyca, Mateusz Kochanek, Dominika Szydło, Joanna Baran, Julita Bielaniewicz, Marcin Gruza, Arkadiusz Janz, Kamil Kanclerz, Anna Kocoń, Bartłomiej Koptyra, Wiktoria Mieleszczenko-Kowszewicz, Piotr Miłkowski, Marcin Oleksy, Maciej Piasecki, Łukasz Radliński, Konrad Wojtasik, Stanisław Woźniak, and Przemysław Kazienko. 2023.
\newblock \href {https://doi.org/https://doi.org/10.1016/j.inffus.2023.101861} {{ChatGPT}: Jack of all trades, master of none}.
\newblock \emph{Information Fusion}, 99:101861.

\bibitem[{Kung et~al.(2023)Kung, Cheatham, Medenilla, Sillos, De~Leon, Elepa{\~n}o, Madriaga, Aggabao, Diaz-Candido, Maningo et~al.}]{kung2023performance}
TH~Kung, M~Cheatham, A~Medenilla, C~Sillos, L~De~Leon, C~Elepa{\~n}o, M~Madriaga, R~Aggabao, G~Diaz-Candido, J~Maningo, et~al. 2023.
\newblock \href {https://doi.org/10.1371/journal.pdig.0000198} {Performance of {ChatGPT} on {USMLE}: Potential for {AI}-assisted medical education using large language models.}
\newblock \emph{Plos Digit Health}, 2:000198.

\bibitem[{Lacoste et~al.(2019)Lacoste, Luccioni, Schmidt, and Dandres}]{lacoste2019quantifying}
Alexandre Lacoste, Alexandra Luccioni, Victor Schmidt, and Thomas Dandres. 2019.
\newblock \href {http://arxiv.org/abs/1910.09700} {Quantifying the carbon emissions of machine learning}.

\bibitem[{Lu et~al.(2022)Lu, Bartolo, Moore, Riedel, and Stenetorp}]{lu2022fantastically}
Yao Lu, Max Bartolo, Alastair Moore, Sebastian Riedel, and Pontus Stenetorp. 2022.
\newblock \href {https://doi.org/10.18653/v1/2022.acl-long.556} {Fantastically ordered prompts and where to find them: Overcoming few-shot prompt order sensitivity}.
\newblock In \emph{Proceedings of the 60th Annual Meeting of the Association for Computational Linguistics (Volume 1: Long Papers)}, pages 8086--8098, Dublin, Ireland. Association for Computational Linguistics.

\bibitem[{Magar and Schwartz(2022)}]{magar-schwartz-2022-data}
Inbal Magar and Roy Schwartz. 2022.
\newblock \href {https://doi.org/10.18653/v1/2022.acl-short.18} {Data contamination: From memorization to exploitation}.
\newblock In \emph{Proceedings of the 60th Annual Meeting of the Association for Computational Linguistics (Volume 2: Short Papers)}, pages 157--165, Dublin, Ireland. Association for Computational Linguistics.

\bibitem[{Mohammad et~al.(2016)Mohammad, Kiritchenko, Sobhani, Zhu, and Cherry}]{mohammad-etal-2016-semeval}
Saif Mohammad, Svetlana Kiritchenko, Parinaz Sobhani, Xiaodan Zhu, and Colin Cherry. 2016.
\newblock \href {https://doi.org/10.18653/v1/S16-1003} {{S}em{E}val-2016 task 6: Detecting stance in tweets}.
\newblock In \emph{Proceedings of the 10th International Workshop on Semantic Evaluation ({S}em{E}val-2016)}, pages 31--41, San Diego, California. Association for Computational Linguistics.

\bibitem[{OpenAI(2023)}]{GPT4-technical-report}
OpenAI. 2023.
\newblock \href {http://arxiv.org/abs/2303.08774} {{GPT-4} technical report}.

\bibitem[{Ouyang et~al.(2022)Ouyang, Wu, Jiang, Almeida, Wainwright, Mishkin, Zhang, Agarwal, Slama, Ray, Schulman, Hilton, Kelton, Miller, Simens, Askell, Welinder, Christiano, Leike, and Lowe}]{NEURIPS2022_b1efde53}
Long Ouyang, Jeffrey Wu, Xu~Jiang, Diogo Almeida, Carroll Wainwright, Pamela Mishkin, Chong Zhang, Sandhini Agarwal, Katarina Slama, Alex Ray, John Schulman, Jacob Hilton, Fraser Kelton, Luke Miller, Maddie Simens, Amanda Askell, Peter Welinder, Paul~F Christiano, Jan Leike, and Ryan Lowe. 2022.
\newblock \href {https://proceedings.neurips.cc/paper_files/paper/2022/file/b1efde53be364a73914f58805a001731-Paper-Conference.pdf} {Training language models to follow instructions with human feedback}.
\newblock In \emph{Advances in Neural Information Processing Systems}, volume~35, pages 27730--27744. Curran Associates, Inc.

\bibitem[{Piktus et~al.(2023)Piktus, Akiki, Villegas, Lauren{\c{c}}on, Dupont, Luccioni, Jernite, and Rogers}]{piktus-etal-2023-roots}
Aleksandra Piktus, Christopher Akiki, Paulo Villegas, Hugo Lauren{\c{c}}on, G{\'e}rard Dupont, Sasha Luccioni, Yacine Jernite, and Anna Rogers. 2023.
\newblock \href {https://doi.org/10.18653/v1/2023.acl-demo.29} {The {ROOTS} search tool: Data transparency for {LLM}s}.
\newblock In \emph{Proceedings of the 61st Annual Meeting of the Association for Computational Linguistics (Volume 3: System Demonstrations)}, pages 304--314, Toronto, Canada. Association for Computational Linguistics.

\bibitem[{Raffel et~al.(2020)Raffel, Shazeer, Roberts, Lee, Narang, Matena, Zhou, Li, and Liu}]{GPT3-contamination}
Colin Raffel, Noam Shazeer, Adam Roberts, Katherine Lee, Sharan Narang, Michael Matena, Yanqi Zhou, Wei Li, and Peter~J. Liu. 2020.
\newblock \href {https://jmlr.org/papers/v21/20-074.html} {Exploring the limits of transfer learning with a unified text-to-text transformer}.
\newblock \emph{J. Mach. Learn. Res.}, 21(1).

\bibitem[{Sainz et~al.(2023)Sainz, Campos, García-Ferrero, Etxaniz, and Agirre}]{llm_contamination_blog}
Oscar Sainz, Jon~Ander Campos, Iker García-Ferrero, Julen Etxaniz, and Eneko Agirre. 2023.
\newblock \href {https://hitz-zentroa.github.io/lm-contamination/blog/} {Did {ChatGPT} cheat on your test?}
\newblock \url{https://hitz-zentroa.github.io/lm-contamination/blog/}.

\bibitem[{Shi and Lipani(2023)}]{shi2023dont}
Zhengxaing Shi and Aldo Lipani. 2023.
\newblock \href {https://openreview.net/forum?id=s7xWeJQACI} {Don't stop pretraining? make prompt-based fine-tuning powerful learner}.
\newblock In \emph{Thirty-seventh Conference on Neural Information Processing Systems}.

\bibitem[{Szefer and Deshpande(2023)}]{szefer2023analyzing}
Jakub Szefer and Sanjay Deshpande. 2023.
\newblock \href {https://arxiv.org/abs/2304.06122} {Analyzing chatgpt's aptitude in an introductory computer engineering course}.
\newblock \emph{arXiv preprint arXiv:2304.06122}.

\bibitem[{Thomson et~al.(2024)Thomson, Reiter, and Belz}]{10.1162/coli_a_00508}
Craig Thomson, Ehud Reiter, and Anya Belz. 2024.
\newblock \href {https://doi.org/10.1162/coli_a_00508} {{Common Flaws in Running Human Evaluation Experiments in NLP}}.
\newblock \emph{Computational Linguistics}, pages 1--10.

\bibitem[{Thoppilan et~al.(2022)Thoppilan, Freitas, Hall, Shazeer, Kulshreshtha, Cheng, Jin, Bos, Baker, Du, Li, Lee, Zheng, Ghafouri, Menegali, Huang, Krikun, Lepikhin, Qin, Chen, Xu, Chen, Roberts, Bosma, Zhao, Zhou, Chang, Krivokon, Rusch, Pickett, Srinivasan, Man, Meier-Hellstern, Morris, Doshi, Santos, Duke, Soraker, Zevenbergen, Prabhakaran, Diaz, Hutchinson, Olson, Molina, Hoffman-John, Lee, Aroyo, Rajakumar, Butryna, Lamm, Kuzmina, Fenton, Cohen, Bernstein, Kurzweil, Aguera-Arcas, Cui, Croak, Chi, and Le}]{thoppilan2022lamda}
Romal Thoppilan, Daniel~De Freitas, Jamie Hall, Noam Shazeer, Apoorv Kulshreshtha, Heng-Tze Cheng, Alicia Jin, Taylor Bos, Leslie Baker, Yu~Du, YaGuang Li, Hongrae Lee, Huaixiu~Steven Zheng, Amin Ghafouri, Marcelo Menegali, Yanping Huang, Maxim Krikun, Dmitry Lepikhin, James Qin, Dehao Chen, Yuanzhong Xu, Zhifeng Chen, Adam Roberts, Maarten Bosma, Vincent Zhao, Yanqi Zhou, Chung-Ching Chang, Igor Krivokon, Will Rusch, Marc Pickett, Pranesh Srinivasan, Laichee Man, Kathleen Meier-Hellstern, Meredith~Ringel Morris, Tulsee Doshi, Renelito~Delos Santos, Toju Duke, Johnny Soraker, Ben Zevenbergen, Vinodkumar Prabhakaran, Mark Diaz, Ben Hutchinson, Kristen Olson, Alejandra Molina, Erin Hoffman-John, Josh Lee, Lora Aroyo, Ravi Rajakumar, Alena Butryna, Matthew Lamm, Viktoriya Kuzmina, Joe Fenton, Aaron Cohen, Rachel Bernstein, Ray Kurzweil, Blaise Aguera-Arcas, Claire Cui, Marian Croak, Ed~Chi, and Quoc Le. 2022.
\newblock \href {http://arxiv.org/abs/2201.08239} {{LaMDA}: Language models for dialog applications}.
\newblock \emph{arXiv preprint arXiv:2201.08239}.

\bibitem[{Touvron et~al.(2023)Touvron, Lavril, Izacard, Martinet, Lachaux, Lacroix, Rozi{\`e}re, Goyal, Hambro, Azhar et~al.}]{touvron2023llama}
Hugo Touvron, Thibaut Lavril, Gautier Izacard, Xavier Martinet, Marie-Anne Lachaux, Timoth{\'e}e Lacroix, Baptiste Rozi{\`e}re, Naman Goyal, Eric Hambro, Faisal Azhar, et~al. 2023.
\newblock \href {https://arxiv.org/abs/2302.13971} {Llama: Open and efficient foundation language models}.
\newblock \emph{arXiv preprint arXiv:2302.13971}.

\bibitem[{Ye et~al.(2022)Ye, Manotumruksa, and Yilmaz}]{ye-etal-2022-multiwoz}
Fanghua Ye, Jarana Manotumruksa, and Emine Yilmaz. 2022.
\newblock \href {https://aclanthology.org/2022.sigdial-1.34} {{M}ulti{WOZ} 2.4: A multi-domain task-oriented dialogue dataset with essential annotation corrections to improve state tracking evaluation}.
\newblock In \emph{Proceedings of the 23rd Annual Meeting of the Special Interest Group on Discourse and Dialogue}, pages 351--360, Edinburgh, UK. Association for Computational Linguistics.

\bibitem[{Zhang et~al.(2022)Zhang, Ding, and Jing}]{zhang2022would}
Bowen Zhang, Daijun Ding, and Liwen Jing. 2022.
\newblock \href {https://arxiv.org/abs/2212.14548} {How would stance detection techniques evolve after the launch of {ChatGPT}?}
\newblock \emph{arXiv preprint arXiv:2212.14548}.

\end{thebibliography}


\begin{thebibliography}{255}
\expandafter\ifx\csname natexlab\endcsname\relax\def\natexlab#1{#1}\fi

\bibitem[{Ahuja et~al.(2023)Ahuja, Diddee, Hada, Ochieng, Ramesh, Jain, Nambi,
  Ganu, Segal, Axmed, Bali, and Sitaram}]{paper155}
Kabir Ahuja, Harshita Diddee, Rishav Hada, Millicent Ochieng, Krithika Ramesh,
  Prachi Jain, Akshay Nambi, Tanuja Ganu, Sameer Segal, Maxamed Axmed, Kalika
  Bali, and Sunayana Sitaram. 2023.
\newblock \href {http://arxiv.org/abs/2303.12528} {MEGA: Multilingual
  evaluation of generative AI}.

\bibitem[{Aiyappa et~al.(2023)Aiyappa, An, Kwak, and Ahn}]{paper3}
Rachith Aiyappa, Jisun An, Haewoon Kwak, and Yong-Yeol Ahn. 2023.
\newblock \href {http://arxiv.org/abs/2303.12767} {Can we trust the evaluation
  on ChatGPT?}

\bibitem[{Akyurek et~al.(2023)Akyurek, Akyurek, Kalyan, Clark, Wijaya, and
  Tandon}]{paper214}
Afra~Feyza Akyurek, Ekin Akyurek, Ashwin Kalyan, Peter Clark, Derry~Tanti
  Wijaya, and Niket Tandon. 2023.
\newblock \href {https://doi.org/10.18653/v1/2023.acl-long.427} {{RL}4{F}:
  Generating natural language feedback with reinforcement learning for
  repairing model outputs}.
\newblock In \emph{Proceedings of the 61st Annual Meeting of the Association
  for Computational Linguistics (Volume 1: Long Papers)}, pages 7716--7733,
  Toronto, Canada. Association for Computational Linguistics.

\bibitem[{Amin et~al.(2023)Amin, Cambria, and Schuller}]{paper19}
Mostafa~M. Amin, Erik Cambria, and Björn~W. Schuller. 2023.
\newblock \href {http://arxiv.org/abs/2303.03186} {Will affective computing
  emerge from foundation models and general AI? a first evaluation on ChatGPT}.

\bibitem[{An et~al.(2023)An, Gong, Zhong, Zhao, Li, Zhang, Kong, and
  Qiu}]{paper102}
Chenxin An, Shansan Gong, Ming Zhong, Xingjian Zhao, Mukai Li, Jun Zhang,
  Lingpeng Kong, and Xipeng Qiu. 2023.
\newblock \href {http://arxiv.org/abs/2307.11088} {L-eval: Instituting
  standardized evaluation for long context language models}.

\bibitem[{Antaki et~al.(2023)Antaki, Touma, Milad, El-Khoury, and
  Duval}]{paper53}
Fares Antaki, Samir Touma, Daniel Milad, Jonathan El-Khoury, and Renaud Duval.
  2023.
\newblock \href{https://www.sciencedirect.com/science/article/pii/S2666914523000568}{Evaluating the performance of ChatGPT in ophthalmology: An analysis
  of its successes and shortcomings.}
\newblock \emph{Ophthalmology Science}, 3(4):100324.

\bibitem[{Baan et~al.(2023)Baan, Daheim, Ilia, Ulmer, Li, Fernández, Plank,
  Sennrich, Zerva, and Aziz}]{paper113}
Joris Baan, Nico Daheim, Evgenia Ilia, Dennis Ulmer, Haau-Sing Li, Raquel
  Fernández, Barbara Plank, Rico Sennrich, Chrysoula Zerva, and Wilker Aziz.
  2023.
\newblock \href {http://arxiv.org/abs/2307.15703} {Uncertainty in natural
  language generation: From theory to applications}.

\bibitem[{Bai et~al.(2023{\natexlab{a}})Bai, Lv, Zhang, Lyu, Tang, Huang, Du,
  Liu, Zeng, Hou, Dong, Tang, and Li}]{paper104}
Yushi Bai, Xin Lv, Jiajie Zhang, Hongchang Lyu, Jiankai Tang, Zhidian Huang,
  Zhengxiao Du, Xiao Liu, Aohan Zeng, Lei Hou, Yuxiao Dong, Jie Tang, and
  Juanzi Li. 2023{\natexlab{a}}.
\newblock \href {http://arxiv.org/abs/2308.14508} {Longbench: A bilingual,
  multitask benchmark for long context understanding}.

\bibitem[{Bai et~al.(2023{\natexlab{b}})Bai, Ying, Cao, Lv, He, Wang, Yu, Zeng,
  Xiao, Lyu, Zhang, Li, and Hou}]{paper58}
Yushi Bai, Jiahao Ying, Yixin Cao, Xin Lv, Yuze He, Xiaozhi Wang, Jifan Yu,
  Kaisheng Zeng, Yijia Xiao, Haozhe Lyu, Jiayin Zhang, Juanzi Li, and Lei Hou.
  2023{\natexlab{b}}.
\newblock \href {http://arxiv.org/abs/2306.04181} {Benchmarking foundation
  models with language-model-as-an-examiner}.

\bibitem[{Bang et~al.(2023)Bang, Cahyawijaya, Lee, Dai, Su, Wilie, Lovenia, Ji,
  Yu, Chung, Do, Xu, and Fung}]{paper12}
Yejin Bang, Samuel Cahyawijaya, Nayeon Lee, Wenliang Dai, Dan Su, Bryan Wilie,
  Holy Lovenia, Ziwei Ji, Tiezheng Yu, Willy Chung, Quyet~V. Do, Yan Xu, and
  Pascale Fung. 2023.
\newblock \href {http://arxiv.org/abs/2302.04023} {A multitask, multilingual,
  multimodal evaluation of ChatGPT on reasoning, hallucination, and
  interactivity}.

\bibitem[{Belouadi and Eger(2023)}]{paper184}
Jonas Belouadi and Steffen Eger. 2023.
\newblock \href {http://arxiv.org/abs/2212.10474} {ByGPT5: End-to-end
  style-conditioned poetry generation with token-free language models}.

\bibitem[{Bian et~al.(2023)Bian, Han, Sun, Lin, Lu, and He}]{paper109}
Ning Bian, Xianpei Han, Le~Sun, Hongyu Lin, Yaojie Lu, and Ben He. 2023.
\newblock \href {http://arxiv.org/abs/2303.16421} {ChatGPT is a knowledgeable
  but inexperienced solver: An investigation of commonsense problem in large
  language models}.

\bibitem[{Bordt and Luxburg(2023)}]{paper42}
Sebastian Bordt and Ulrike~von Luxburg. 2023.
\newblock \href {http://arxiv.org/abs/2303.09461} {ChatGPT participates in a
  computer science exam}.

\bibitem[{Borji(2023)}]{paper31}
Ali Borji. 2023.
\newblock \href {http://arxiv.org/abs/2302.03494} {A categorical archive of
  ChatGPT failures}.

\bibitem[{Bose et~al.(2023)Bose, Perera, and Dorr}]{paper221}
Ritwik Bose, Ian Perera, and Bonnie Dorr. 2023.
\newblock \href {https://doi.org/10.18653/v1/2023.sicon-1.2} {Detoxifying
  online discourse: A guided response generation approach for reducing toxicity
  in user-generated text}.
\newblock In \emph{Proceedings of the First Workshop on Social Influence in
  Conversations (SICon 2023)}, pages 9--14, Toronto, Canada. Association for
  Computational Linguistics.

\bibitem[{Bubeck et~al.(2023)Bubeck, Chandrasekaran, Eldan, Gehrke, Horvitz,
  Kamar, Lee, Lee, Li, Lundberg, Nori, Palangi, Ribeiro, and Zhang}]{paper41}
Sébastien Bubeck, Varun Chandrasekaran, Ronen Eldan, Johannes Gehrke, Eric
  Horvitz, Ece Kamar, Peter Lee, Yin~Tat Lee, Yuanzhi Li, Scott Lundberg,
  Harsha Nori, Hamid Palangi, Marco~Tulio Ribeiro, and Yi~Zhang. 2023.
\newblock \href {http://arxiv.org/abs/2303.12712} {Sparks of artificial general
  intelligence: Early experiments with GPT-4}.

\bibitem[{Bucur(2023)}]{paper192}
Ana-Maria Bucur. 2023.
\newblock \href {http://arxiv.org/abs/2307.02313} {Utilizing ChatGPT generated
  data to retrieve depression symptoms from social media}.

\bibitem[{Cabello et~al.(2023)Cabello, Li, and Chalkidis}]{paper98}
Laura Cabello, Jiaang Li, and Ilias Chalkidis. 2023.
\newblock \href {http://arxiv.org/abs/2306.03024} {Pokemonchat: Auditing
  ChatGPT for pokémon universe knowledge}.

\bibitem[{Cabrera and Neubig(2023)}]{paper11}
Alex Cabrera and Graham Neubig. 2023.
\newblock \href
  {https://github.com/zeno-ml/zeno-build/tree/main/examples/chatbot/report}
  {Zeno chatbot report}.

\bibitem[{Cai et~al.(2023)Cai, Haslett, Duan, Wang, and
  Pickering}]{cai2023does}
Zhenguang~G Cai, David~A Haslett, Xufeng Duan, Shuqi Wang, and Martin~J
  Pickering. 2023.
\newblock \href{https://arxiv.org/abs/2303.08014}{Does ChatGPT resemble humans in language use?}
\newblock \emph{arXiv preprint arXiv:2303.08014}.

\bibitem[{Cao et~al.(2023)Cao, Zhou, Lee, Cabello, Chen, and
  Hershcovich}]{paper203}
Yong Cao, Li~Zhou, Seolhwa Lee, Laura Cabello, Min Chen, and Daniel
  Hershcovich. 2023.
\newblock \href {https://doi.org/10.18653/v1/2023.c3nlp-1.7} {Assessing
  cross-cultural alignment between {C}hat{GPT} and human societies: An
  empirical study}.
\newblock In \emph{Proceedings of the First Workshop on Cross-Cultural
  Considerations in NLP (C3NLP)}, pages 53--67, Dubrovnik, Croatia. Association
  for Computational Linguistics.

\bibitem[{Carlini et~al.(2023)Carlini, Ippolito, Jagielski, Lee, Tramer, and
  Zhang}]{paper7}
Nicholas Carlini, Daphne Ippolito, Matthew Jagielski, Katherine Lee, Florian
  Tramer, and Chiyuan Zhang. 2023.
\newblock \href {http://arxiv.org/abs/2202.07646} {Quantifying memorization
  across neural language models}.

\bibitem[{Chakraborty et~al.(2023)Chakraborty, Kulkarni, and Li}]{paper166}
Mohna Chakraborty, Adithya Kulkarni, and Qi~Li. 2023.
\newblock \href {https://doi.org/10.18653/v1/2023.acl-long.313} {Zero-shot
  approach to overcome perturbation sensitivity of prompts}.
\newblock In \emph{Proceedings of the 61st Annual Meeting of the Association
  for Computational Linguistics (Volume 1: Long Papers)}, pages 5698--5711,
  Toronto, Canada. Association for Computational Linguistics.

\bibitem[{Chandrasekhar et~al.(2023)Chandrasekhar, Huang, and Huang}]{paper232}
Shreya Chandrasekhar, Chieh-Yang Huang, and Ting-Hao Huang. 2023.
\newblock \href {https://doi.org/10.18653/v1/2023.bionlp-1.8} {Good data, large
  data, or no data? comparing three approaches in developing research aspect
  classifiers for biomedical papers}.
\newblock In \emph{The 22nd Workshop on Biomedical Natural Language Processing
  and BioNLP Shared Tasks}, pages 103--113, Toronto, Canada. Association for
  Computational Linguistics.

\bibitem[{Chang et~al.(2023)Chang, Wang, Wang, Wu, Yang, Zhu, Chen, Yi, Wang,
  Wang, Ye, Zhang, Chang, Yu, Yang, and Xie}]{paper64}
Yupeng Chang, Xu~Wang, Jindong Wang, Yuan Wu, Linyi Yang, Kaijie Zhu, Hao Chen,
  Xiaoyuan Yi, Cunxiang Wang, Yidong Wang, Wei Ye, Yue Zhang, Yi~Chang,
  Philip~S. Yu, Qiang Yang, and Xing Xie. 2023.
\newblock \href {http://arxiv.org/abs/2307.03109} {A survey on evaluation of
  large language models}.

\bibitem[{Chen et~al.(2023{\natexlab{a}})Chen, Pan, Yu, Song, Wang, Yu, and
  Chen}]{paper133}
Jiaao Chen, Xiaoman Pan, Dian Yu, Kaiqiang Song, Xiaoyang Wang, Dong Yu, and
  Jianshu Chen. 2023{\natexlab{a}}.
\newblock \href {http://arxiv.org/abs/2308.00304} {Skills-in-context prompting:
  Unlocking compositionality in large language models}.

\bibitem[{Chen et~al.(2023{\natexlab{b}})Chen, Zaharia, and
  Zou}]{chen2023chatgpt}
Lingjiao Chen, Matei Zaharia, and James Zou. 2023{\natexlab{b}}.
\newblock \href{https://arxiv.org/abs/2307.09009}{How is ChatGPT's behavior changing over time?}
\newblock \emph{arXiv preprint arXiv:2307.09009}.

\bibitem[{Chen et~al.(2023{\natexlab{c}})Chen, Wang, Jiang, Cai, Li, Chen,
  Wang, and Li}]{paper182}
Nuo Chen, Yan Wang, Haiyun Jiang, Deng Cai, Yuhan Li, Ziyang Chen, Longyue
  Wang, and Jia Li. 2023{\natexlab{c}}.
\newblock \href {http://arxiv.org/abs/2211.06869} {Large language models meet
  Harry Potter: A bilingual dataset for aligning dialogue agents with
  characters}.

\bibitem[{Chen et~al.(2023{\natexlab{d}})Chen, Ye, Zu, Xu, Zheng, Peng, Zhou,
  Gui, Zhang, and Huang}]{paper70}
Xuanting Chen, Junjie Ye, Can Zu, Nuo Xu, Rui Zheng, Minlong Peng, Jie Zhou,
  Tao Gui, Qi~Zhang, and Xuanjing Huang. 2023{\natexlab{d}}.
\newblock \href {http://arxiv.org/abs/2303.00293} {How robust is GPT-3.5 to
  predecessors? a comprehensive study on language understanding tasks}.

\bibitem[{Chen and Eger(2022)}]{paper185}
Yanran Chen and Steffen Eger. 2022.
\newblock \href {http://arxiv.org/abs/2212.10522} {Transformers go for the
  lols: Generating (humourous) titles from scientific abstracts end-to-end}.

\bibitem[{Chiang and Lee(2023)}]{paper213}
Cheng-Han Chiang and Hung-yi Lee. 2023.
\newblock \href {https://doi.org/10.18653/v1/2023.acl-long.870} {Can large
  language models be an alternative to human evaluations?}
\newblock In \emph{Proceedings of the 61st Annual Meeting of the Association
  for Computational Linguistics (Volume 1: Long Papers)}, pages 15607--15631,
  Toronto, Canada. Association for Computational Linguistics.

\bibitem[{Choi et~al.(2023)Choi, Hickman, Monahan, and Schwarcz}]{paper48}
Jonathan~H. Choi, Kristin~E. Hickman, Amy Monahan, and Daniel~B. Schwarcz.
  2023.
\newblock \href {https://doi.org/10.2139/ssrn.4335905} {ChatGPT goes to law
  school}.
\newblock \emph{Journal of Legal Education}.

\bibitem[{Chowdhury et~al.(2023)Chowdhury, Lim, Higham, McKinnon, Ventoura, He,
  and De~Pennington}]{paper226}
Mohita Chowdhury, Ernest Lim, Aisling Higham, Rory McKinnon, Nikoletta
  Ventoura, Yajie He, and Nick De~Pennington. 2023.
\newblock \href {https://doi.org/10.18653/v1/2023.clinicalnlp-1.17} {Can large
  language models safely address patient questions following cataract surgery?}
\newblock In \emph{Proceedings of the 5th Clinical Natural Language Processing
  Workshop}, pages 131--137, Toronto, Canada. Association for Computational
  Linguistics.

\bibitem[{Chu and Liu(2023)}]{paper57}
Haoran Chu and Sixiao Liu. 2023.
\newblock \href {https://doi.org/10.31234/osf.io/c3549} {Can AI tell good
  stories? narrative transportation and persuasion with ChatGPT.}
\newblock \emph{PsyArXiv}.

\bibitem[{Clark(2023)}]{paper46}
Ted~M. Clark. 2023.
\newblock \href {https://doi.org/10.1021/acs.jchemed.3c00027} {Investigating
  the use of an artificial intelligence chatbot with general chemistry exam
  questions}.
\newblock \emph{Journal of Chemical Education}, 100(5):1905--1916.

\bibitem[{Dahlkemper et~al.(2023)Dahlkemper, Lahme, and Klein}]{paper47}
Merten~Nikolay Dahlkemper, Simon~Zacharias Lahme, and Pascal Klein. 2023.
\newblock \href {http://arxiv.org/abs/2304.05906} {How do physics students
  evaluate artificial intelligence responses on comprehension questions? a
  study on the perceived scientific accuracy and linguistic quality}.

\bibitem[{Dai et~al.(2023{\natexlab{a}})Dai, Liu, Liao, Huang, Cao, Wu, Zhao,
  Xu, Liu, Liu, Li, Zhu, Cai, Sun, Li, Shen, Liu, and Li}]{paper148}
Haixing Dai, Zhengliang Liu, Wenxiong Liao, Xiaoke Huang, Yihan Cao, Zihao Wu,
  Lin Zhao, Shaochen Xu, Wei Liu, Ninghao Liu, Sheng Li, Dajiang Zhu, Hongmin
  Cai, Lichao Sun, Quanzheng Li, Dinggang Shen, Tianming Liu, and Xiang Li.
  2023{\natexlab{a}}.
\newblock \href {http://arxiv.org/abs/2302.13007} {AugGPT: Leveraging ChatGPT
  for text data augmentation}.

\bibitem[{Dai et~al.(2023{\natexlab{b}})Dai, Shao, Zhao, Yu, Si, Xu, Sun,
  Zhang, and Xu}]{paper95}
Sunhao Dai, Ninglu Shao, Haiyuan Zhao, Weijie Yu, Zihua Si, Chen Xu, Zhongxiang
  Sun, Xiao Zhang, and Jun Xu. 2023{\natexlab{b}}.
\newblock \href {http://arxiv.org/abs/2305.02182} {Uncovering ChatGPT's
  capabilities in recommender systems}.

\bibitem[{Das et~al.(2023)Das, Pandey, and Mukherjee}]{paper81}
Mithun Das, Saurabh~Kumar Pandey, and Animesh Mukherjee. 2023.
\newblock \href {http://arxiv.org/abs/2305.13276} {Evaluating ChatGPT's
  performance for multilingual and emoji-based hate speech detection}.

\bibitem[{Davis(2023)}]{paper32}
Ernest Davis. 2023.
\newblock \href {http://arxiv.org/abs/2302.04752} {Benchmarks for automated
  commonsense reasoning: A survey}.

\bibitem[{Deshpande and Szefer(2023)}]{paper152}
Sanjay Deshpande and Jakub Szefer. 2023.
\newblock \href {http://arxiv.org/abs/2304.06122} {Analyzing ChatGPT's aptitude
  in an introductory computer engineering course}.

\bibitem[{Dhingra et~al.(2023)Dhingra, Singh, SB, Malviya, and Gill}]{paper106}
Sifatkaur Dhingra, Manmeet Singh, Vaisakh SB, Neetiraj Malviya, and
  Sukhpal~Singh Gill. 2023.
\newblock \href {http://arxiv.org/abs/2303.11436} {Mind meets machine:
  Unravelling GPT-4's cognitive psychology}.

\bibitem[{Ding et~al.(2023)Ding, Qin, Liu, Chia, Li, Joty, and Bing}]{paper219}
Bosheng Ding, Chengwei Qin, Linlin Liu, Yew~Ken Chia, Boyang Li, Shafiq Joty,
  and Lidong Bing. 2023.
\newblock \href {https://doi.org/10.18653/v1/2023.acl-long.626} {Is {GPT}-3 a
  good data annotator?}
\newblock In \emph{Proceedings of the 61st Annual Meeting of the Association
  for Computational Linguistics (Volume 1: Long Papers)}, pages 11173--11195,
  Toronto, Canada. Association for Computational Linguistics.

\bibitem[{Duenas et~al.(2023)Duenas, Jimenez, and Mateus~Ferro}]{paper198}
George Duenas, Sergio Jimenez, and Geral Mateus~Ferro. 2023.
\newblock \href {https://doi.org/10.18653/v1/2023.bea-1.30} {You{'}ve got a
  friend in ... a language model? a comparison of explanations of
  multiple-choice items of reading comprehension between {C}hat{GPT} and
  humans}.
\newblock In \emph{Proceedings of the 18th Workshop on Innovative Use of NLP
  for Building Educational Applications (BEA 2023)}, pages 372--381, Toronto,
  Canada. Association for Computational Linguistics.

\bibitem[{Espejel et~al.(2023)Espejel, Ettifouri, Alassan, Chouham, and
  Dahhane}]{paper125}
Jessica~López Espejel, El~Hassane Ettifouri, Mahaman Sanoussi~Yahaya Alassan,
  El~Mehdi Chouham, and Walid Dahhane. 2023.
\newblock \href {http://arxiv.org/abs/2305.12477} {GPT-3.5, GPT-4, or bard?
  evaluating LLMs reasoning ability in zero-shot setting and performance
  boosting through prompts}.

\bibitem[{Fan and Jiang(2023)}]{paper80}
Yaxin Fan and Feng Jiang. 2023.
\newblock \href {http://arxiv.org/abs/2305.08391} {Uncovering the potential of
  ChatGPT for discourse analysis in dialogue: An empirical study}.

\bibitem[{Fang et~al.(2023)Fang, Yang, Lan, Wong, Hu, Chao, and
  Zhang}]{paper253}
Tao Fang, Shu Yang, Kaixin Lan, Derek~F. Wong, Jinpeng Hu, Lidia~S. Chao, and
  Yue Zhang. 2023.
\newblock \href {http://arxiv.org/abs/2304.01746} {Is ChatGPT a highly fluent
  grammatical error correction system? a comprehensive evaluation}.

\bibitem[{Finch et~al.(2023)Finch, Paek, and Choi}]{paper206}
Sarah~E. Finch, Ellie~S. Paek, and Jinho~D. Choi. 2023.
\newblock \href {https://aclanthology.org/2023.sigdial-1.20} {Leveraging large
  language models for automated dialogue analysis}.
\newblock In \emph{Proceedings of the 24th Meeting of the Special Interest
  Group on Discourse and Dialogue}, pages 202--215, Prague, Czechia.
  Association for Computational Linguistics.

\bibitem[{Fraser et~al.(2023)Fraser, Kiritchenko, Nejadgholi, and
  Kerkhof}]{paper248}
Kathleen Fraser, Svetlana Kiritchenko, Isar Nejadgholi, and Anna Kerkhof. 2023.
\newblock \href {https://doi.org/10.18653/v1/2023.sicon-1.4} {What makes a good
  counter-stereotype? Evaluating strategies for automated responses to
  stereotypical text}.
\newblock In \emph{Proceedings of the First Workshop on Social Influence in
  Conversations (SICon 2023)}, pages 25--38, Toronto, Canada. Association for
  Computational Linguistics.

\bibitem[{Frieder et~al.(2023)Frieder, Pinchetti, Chevalier, Griffiths,
  Salvatori, Lukasiewicz, Petersen, and Berner}]{paper44}
Simon Frieder, Luca Pinchetti, Alexis Chevalier, Ryan-Rhys Griffiths, Tommaso
  Salvatori, Thomas Lukasiewicz, Philipp~Christian Petersen, and Julius Berner.
  2023.
\newblock \href {http://arxiv.org/abs/2301.13867} {Mathematical capabilities of
  ChatGPT}.

\bibitem[{Gao et~al.(2023{\natexlab{a}})Gao, Ding, Qin, and Liu}]{paper33}
Jinglong Gao, Xiao Ding, Bing Qin, and Ting Liu. 2023{\natexlab{a}}.
\newblock \href {http://arxiv.org/abs/2305.07375} {Is ChatGPT a good causal
  reasoner? a comprehensive evaluation}.

\bibitem[{Gao et~al.(2023{\natexlab{b}})Gao, Zhao, Yu, and Xu}]{paper168}
Jun Gao, Huan Zhao, Changlong Yu, and Ruifeng Xu. 2023{\natexlab{b}}.
\newblock \href {http://arxiv.org/abs/2303.03836} {Exploring the feasibility of
  ChatGPT for event extraction}.

\bibitem[{Gao et~al.(2023{\natexlab{c}})Gao, Ruan, Sun, Yin, Yang, and
  Wan}]{paper158}
Mingqi Gao, Jie Ruan, Renliang Sun, Xunjian Yin, Shiping Yang, and Xiaojun Wan.
  2023{\natexlab{c}}.
\newblock \href {http://arxiv.org/abs/2304.02554} {Human-like summarization
  evaluation with ChatGPT}.

\bibitem[{Gao et~al.(2023{\natexlab{d}})Gao, Yen, Yu, and Chen}]{paper127}
Tianyu Gao, Howard Yen, Jiatong Yu, and Danqi Chen. 2023{\natexlab{d}}.
\newblock \href {http://arxiv.org/abs/2305.14627} {Enabling large language
  models to generate text with citations}.

\bibitem[{Gao et~al.(2023{\natexlab{e}})Gao, Wang, and Hou}]{paper143}
Yuan Gao, Ruili Wang, and Feng Hou. 2023{\natexlab{e}}.
\newblock \href {http://arxiv.org/abs/2304.02182} {How to design translation
  prompts for ChatGPT: An empirical study}.

\bibitem[{Geerling et~al.(2023)Geerling, Mateer, Wooten, and
  Damodaran}]{paper49}
Wayne Geerling, G.~Dirk Mateer, Jadrian Wooten, and Nikhil Damodaran. 2023.
\newblock \href {https://doi.org/10.1177/05694345231169654} {ChatGPT has aced
  the test of understanding in college economics: Now what?}
\newblock \emph{The American Economist}, 68(2):233--245.

\bibitem[{Ghahroodi et~al.(2023)Ghahroodi, Dalili, Mesforoush, and
  Asgari}]{paper210}
Omid Ghahroodi, Seyed~Arshan Dalili, Sahel Mesforoush, and Ehsaneddin Asgari.
  2023.
\newblock \href {https://doi.org/10.18653/v1/2023.semeval-1.298} {{SUT} at
  {S}em{E}val-2023 task 1: Prompt generation for visual word sense
  disambiguation}.
\newblock In \emph{Proceedings of the 17th International Workshop on Semantic
  Evaluation (SemEval-2023)}, pages 2160--2163, Toronto, Canada. Association
  for Computational Linguistics.

\bibitem[{Ghanadian et~al.(2023)Ghanadian, Nejadgholi, and Osman}]{paper83}
Hamideh Ghanadian, Isar Nejadgholi, and Hussein~Al Osman. 2023.
\newblock \href {http://arxiv.org/abs/2306.09390} {ChatGPT for suicide risk
  assessment on social media: Quantitative evaluation of model performance,
  potentials and limitations}.

\bibitem[{Gilardi et~al.(2023)Gilardi, Alizadeh, and Kubli}]{paper26}
Fabrizio Gilardi, Meysam Alizadeh, and Maël Kubli. 2023.
\newblock \href {http://arxiv.org/abs/2303.15056} {ChatGPT outperforms
  crowd-workers for text-annotation tasks}.

\bibitem[{Gilson et~al.(2023)Gilson, Safranek, Huang, Socrates, Chi, Taylor,
  and Chartash}]{paper51}
Aidan Gilson, Conrad~W Safranek, Thomas Huang, Vimig Socrates, Ling Chi,
  Richard~Andrew Taylor, and David Chartash. 2023.
\newblock \href{https://pubmed.ncbi.nlm.nih.gov/36753318/}{How does ChatGPT perform on the United States Medical Licensing
  Examination? The implications of large language models for medical education
  and knowledge assessment.}
\newblock \emph{JMIR Med Educ}, 9.

\bibitem[{Github(2023)}]{paper9}
Github. 2023.
\newblock \href
  {https://github.com/THU-KEG/EvaluationPapers4ChatGPT#evaluation-papers-for-chatgpt}
  {Evaluation papers for ChatGPT}.

\bibitem[{Golchin and Surdeanu(2023)}]{golchin2023time}
Shahriar Golchin and Mihai Surdeanu. 2023.
\newblock \href{https://arxiv.org/abs/2308.08493}{Time travel in LLMs: Tracing data contamination in large language
  models.}
\newblock \emph{arXiv preprint arXiv:2308.08493}.

\bibitem[{Gowriraj et~al.(2023)Gowriraj, Tiwari, Potnis, Bansal, Mitamura, and
  Nyberg}]{paper199}
Srinivas Gowriraj, Soham~Dinesh Tiwari, Mitali Potnis, Srijan Bansal, Teruko
  Mitamura, and Eric Nyberg. 2023.
\newblock \href {https://doi.org/10.18653/v1/2023.dialdoc-1.11}
  {Language-agnostic transformers and assessing {C}hat{GPT}-based query
  rewriting for multilingual document-grounded {QA}}.
\newblock In \emph{Proceedings of the Third DialDoc Workshop on
  Document-grounded Dialogue and Conversational Question Answering}, pages
  101--108, Toronto, Canada. Association for Computational Linguistics.

\bibitem[{Gu(2023)}]{paper142}
Wenshi Gu. 2023.
\newblock \href {http://arxiv.org/abs/2303.15587} {Linguistically informed
  ChatGPT prompts to enhance japanese-chinese machine translation: A case study
  on attributive clauses}.

\bibitem[{Gubelmann et~al.(2023)Gubelmann, Kalouli, Niklaus, and
  Handschuh}]{paper242}
Reto Gubelmann, Aikaterini-lida Kalouli, Christina Niklaus, and Siegfried
  Handschuh. 2023.
\newblock \href {https://doi.org/10.18653/v1/2023.starsem-1.4} {When truth
  matters - addressing pragmatic categories in natural language inference
  ({NLI}) by large language models ({LLM}s)}.
\newblock In \emph{Proceedings of the 12th Joint Conference on Lexical and
  Computational Semantics (*SEM 2023)}, pages 24--39, Toronto, Canada.
  Association for Computational Linguistics.

\bibitem[{Guo et~al.(2023)Guo, Zhang, Wang, Jiang, Nie, Ding, Yue, and
  Wu}]{paper4}
Biyang Guo, Xin Zhang, Ziyuan Wang, Minqi Jiang, Jinran Nie, Yuxuan Ding,
  Jianwei Yue, and Yupeng Wu. 2023.
\newblock \href {http://arxiv.org/abs/2301.07597} {How close is ChatGPT to
  human experts? comparison corpus, evaluation, and detection}.

\bibitem[{Gururangan et~al.(2020)Gururangan, Marasovi{\'c}, Swayamdipta, Lo,
  Beltagy, Downey, and Smith}]{gururangan-etal-2020-dont}
Suchin Gururangan, Ana Marasovi{\'c}, Swabha Swayamdipta, Kyle Lo, Iz~Beltagy,
  Doug Downey, and Noah~A. Smith. 2020.
\newblock \href {https://doi.org/10.18653/v1/2020.acl-main.740} {Don{'}t stop
  pretraining: Adapt language models to domains and tasks}.
\newblock In \emph{Proceedings of the 58th Annual Meeting of the Association
  for Computational Linguistics}, pages 8342--8360, Online. Association for
  Computational Linguistics.

\bibitem[{Hashem et~al.(2023)Hashem, Wang, Wijaya, Ali, and Li}]{paper224}
Tahsina Hashem, Weiqing Wang, Derry~Tanti Wijaya, Mohammed~Eunus Ali, and
  Yuan-Fang Li. 2023.
\newblock \href {https://aclanthology.org/2023.inlg-main.8} {Generating
  faithful text from a knowledge graph with noisy reference text}.
\newblock In \emph{Proceedings of the 16th International Natural Language
  Generation Conference}, pages 106--122, Prague, Czechia. Association for
  Computational Linguistics.

\bibitem[{Havaldar et~al.(2023)Havaldar, Singhal, Rai, Liu, Guntuku, and
  Ungar}]{paper231}
Shreya Havaldar, Bhumika Singhal, Sunny Rai, Langchen Liu, Sharath~Chandra
  Guntuku, and Lyle Ungar. 2023.
\newblock \href {https://doi.org/10.18653/v1/2023.wassa-1.19} {Multilingual
  language models are not multicultural: A case study in emotion}.
\newblock In \emph{Proceedings of the 13th Workshop on Computational Approaches
  to Subjectivity, Sentiment, {\&} Social Media Analysis}, pages 202--214,
  Toronto, Canada. Association for Computational Linguistics.

\bibitem[{He et~al.(2023{\natexlab{a}})He, Wang, Hu, Liu, Liu, Xu, and
  Shen}]{paper170}
Jiabang He, Lei Wang, Yi~Hu, Ning Liu, Hui Liu, Xing Xu, and Heng~Tao Shen.
  2023{\natexlab{a}}.
\newblock \href {http://arxiv.org/abs/2303.05063} {ICL-D3IE: In-context
  learning with diverse demonstrations updating for document information
  extraction}.

\bibitem[{He et~al.(2023{\natexlab{b}})He, Zeng, Huang, Chen, Xiao, He, Zhou,
  Chen, Wang, Huang, Ye, Li, Chen, Zhang, Gu, Liang, and Xiao}]{paper105}
Qianyu He, Jie Zeng, Wenhao Huang, Lina Chen, Jin Xiao, Qianxi He, Xunzhe Zhou,
  Lida Chen, Xintao Wang, Yuncheng Huang, Haoning Ye, Zihan Li, Shisong Chen,
  Yikai Zhang, Zhouhong Gu, Jiaqing Liang, and Yanghua Xiao.
  2023{\natexlab{b}}.
\newblock \href {http://arxiv.org/abs/2309.09150} {Can large language models
  understand real-world complex instructions?}

\bibitem[{He et~al.(2023{\natexlab{c}})He, Shen, Chen, Backes, and
  Zhang}]{paper180}
Xinlei He, Xinyue Shen, Zeyuan Chen, Michael Backes, and Yang Zhang.
  2023{\natexlab{c}}.
\newblock \href {http://arxiv.org/abs/2303.14822} {MGTBench: Benchmarking
  machine-generated text detection}.

\bibitem[{Heck et~al.(2023)Heck, Lubis, Ruppik, Vukovic, Feng, Geishauser, Lin,
  van Niekerk, and Gasic}]{paper194}
Michael Heck, Nurul Lubis, Benjamin Ruppik, Renato Vukovic, Shutong Feng,
  Christian Geishauser, Hsien-chin Lin, Carel van Niekerk, and Milica Gasic.
  2023.
\newblock \href {https://doi.org/10.18653/v1/2023.acl-short.81} {{C}hat{GPT}
  for zero-shot dialogue state tracking: A solution or an opportunity?}
\newblock In \emph{Proceedings of the 61st Annual Meeting of the Association
  for Computational Linguistics (Volume 2: Short Papers)}, pages 936--950,
  Toronto, Canada. Association for Computational Linguistics.

\bibitem[{Hendy et~al.(2023)Hendy, Abdelrehim, Sharaf, Raunak, Gabr,
  Matsushita, Kim, Afify, and Awadalla}]{paper14}
Amr Hendy, Mohamed Abdelrehim, Amr Sharaf, Vikas Raunak, Mohamed Gabr, Hitokazu
  Matsushita, Young~Jin Kim, Mohamed Afify, and Hany~Hassan Awadalla. 2023.
\newblock \href {http://arxiv.org/abs/2302.09210} {How good are GPT models at
  machine translation? a comprehensive evaluation}.

\bibitem[{Hirosawa et~al.(2023)Hirosawa, Harada, Yokose, Sakamoto, Kawamura,
  and Shimizu}]{paper54}
Takanobu Hirosawa, Yukinori Harada, Masashi Yokose, Tetsu Sakamoto, Ren
  Kawamura, and Taro Shimizu. 2023.
\newblock \href {https://www.mdpi.com/1660-4601/20/4/3378} {Diagnostic accuracy
  of differential-diagnosis lists generated by generative pretrained
  transformer 3 chatbot for clinical vignettes with common chief complaints: A
  pilot study}.
\newblock \emph{International Journal of Environmental Research and Public
  Health}, 20(4).

\bibitem[{Holterman and Deemter(2023)}]{paper37}
Bart Holterman and Kees~van Deemter. 2023.
\newblock \href {http://arxiv.org/abs/2305.14020} {Does ChatGPT have theory of
  mind?}

\bibitem[{Holtzman et~al.(2023)Holtzman, West, and Zettlemoyer}]{paper111}
Ari Holtzman, Peter West, and Luke Zettlemoyer. 2023.
\newblock \href {http://arxiv.org/abs/2308.00189} {Generative models as a
  complex systems science: How can we make sense of large language model
  behavior?}

\bibitem[{Hong et~al.(2023)Hong, Zhang, Zhao, Yu, and Zhang}]{paper222}
Ruixin Hong, Hongming Zhang, Hong Zhao, Dong Yu, and Changshui Zhang. 2023.
\newblock \href {https://doi.org/10.18653/v1/2023.acl-long.218} {Faithful
  question answering with {M}onte-{C}arlo planning}.
\newblock In \emph{Proceedings of the 61st Annual Meeting of the Association
  for Computational Linguistics (Volume 1: Long Papers)}, pages 3944--3965,
  Toronto, Canada. Association for Computational Linguistics.

\bibitem[{Hu et~al.(2023{\natexlab{a}})Hu, Lu, Zhang, Song, Lam, and
  Zhang}]{paper119}
Hanxu Hu, Hongyuan Lu, Huajian Zhang, Yun-Ze Song, Wai Lam, and Yue Zhang.
  2023{\natexlab{a}}.
\newblock \href {http://arxiv.org/abs/2305.10276} {Chain-of-symbol prompting
  elicits planning in large langauge models}.

\bibitem[{Hu et~al.(2023{\natexlab{b}})Hu, Wu, Qi, Min, Chen, Pan, and
  Ali}]{paper174}
Nan Hu, Yike Wu, Guilin Qi, Dehai Min, Jiaoyan Chen, Jeff~Z. Pan, and Zafar
  Ali. 2023{\natexlab{b}}.
\newblock \href {http://arxiv.org/abs/2303.10368} {An empirical study of
  pre-trained language models in simple knowledge graph question answering}.

\bibitem[{Hu et~al.(2023{\natexlab{c}})Hu, Ameer, Zuo, Peng, Zhou, Li, Li, Li,
  Jiang, and Xu}]{paper157}
Yan Hu, Iqra Ameer, Xu~Zuo, Xueqing Peng, Yujia Zhou, Zehan Li, Yiming Li,
  Jianfu Li, Xiaoqian Jiang, and Hua Xu. 2023{\natexlab{c}}.
\newblock \href {http://arxiv.org/abs/2303.16416} {Zero-shot clinical entity
  recognition using ChatGPT}.

\bibitem[{Huang et~al.(2023{\natexlab{a}})Huang, Kwak, and An}]{paper86}
Fan Huang, Haewoon Kwak, and Jisun An. 2023{\natexlab{a}}.
\newblock \href {http://arxiv.org/abs/2302.07736} {Is ChatGPT better than human
  annotators? potential and limitations of ChatGPT in explaining implicit hate
  speech}.

\bibitem[{Huang et~al.(2023{\natexlab{b}})Huang, Tang, Zhang, Zhao, Song, Xia,
  and Wei}]{paper89}
Haoyang Huang, Tianyi Tang, Dongdong Zhang, Wayne~Xin Zhao, Ting Song, Yan Xia,
  and Furu Wei. 2023{\natexlab{b}}.
\newblock \href {http://arxiv.org/abs/2305.07004} {Not all languages are
  created equal in LLMs: Improving multilingual capability by
  cross-lingual-thought prompting}.

\bibitem[{Huang et~al.(2023{\natexlab{c}})Huang, Tian, Fayek, and
  Zhang}]{paper245}
Nannan Huang, Lin Tian, Haytham Fayek, and Xiuzhen Zhang. 2023{\natexlab{c}}.
\newblock \href {https://doi.org/10.18653/v1/2023.wassa-1.14} {Examining bias
  in opinion summarisation through the perspective of opinion diversity}.
\newblock In \emph{Proceedings of the 13th Workshop on Computational Approaches
  to Subjectivity, Sentiment, {\&} Social Media Analysis}, pages 149--161,
  Toronto, Canada. Association for Computational Linguistics.

\bibitem[{Huang et~al.(2023{\natexlab{d}})Huang, Song, Wang, Zhao, Chen,
  Juefei-Xu, and Ma}]{paper117}
Yuheng Huang, Jiayang Song, Zhijie Wang, Shengming Zhao, Huaming Chen, Felix
  Juefei-Xu, and Lei Ma. 2023{\natexlab{d}}.
\newblock \href {http://arxiv.org/abs/2307.10236} {Look before you leap: An
  exploratory study of uncertainty measurement for large language models}.

\bibitem[{Huang et~al.(2023{\natexlab{e}})Huang, Bai, Zhu, Zhang, Zhang, Su,
  Liu, Lv, Zhang, Lei, Fu, Sun, and He}]{paper28}
Yuzhen Huang, Yuzhuo Bai, Zhihao Zhu, Junlei Zhang, Jinghan Zhang, Tangjun Su,
  Junteng Liu, Chuancheng Lv, Yikai Zhang, Jiayi Lei, Yao Fu, Maosong Sun, and
  Junxian He. 2023{\natexlab{e}}.
\newblock \href {http://arxiv.org/abs/2305.08322} {C-eval: A multi-level
  multi-discipline chinese evaluation suite for foundation models}.

\bibitem[{Ippolito et~al.(2023)Ippolito, Tramer, Nasr, Zhang, Jagielski, Lee,
  Choquette~Choo, and Carlini}]{ippolito-etal-2023-preventing}
Daphne Ippolito, Florian Tramer, Milad Nasr, Chiyuan Zhang, Matthew Jagielski,
  Katherine Lee, Christopher Choquette~Choo, and Nicholas Carlini. 2023.
\newblock \href {https://aclanthology.org/2023.inlg-main.3} {Preventing
  generation of verbatim memorization in language models gives a false sense of
  privacy}.
\newblock In \emph{Proceedings of the 16th International Natural Language
  Generation Conference}, pages 28--53, Prague, Czechia. Association for
  Computational Linguistics.

\bibitem[{Jahan et~al.(2023)Jahan, Laskar, Peng, and Huang}]{paper156}
Israt Jahan, Md~Tahmid~Rahman Laskar, Chun Peng, and Jimmy Huang. 2023.
\newblock \href {https://doi.org/10.18653/v1/2023.bionlp-1.30} {Evaluation of
  {C}hat{GPT} on biomedical tasks: A zero-shot comparison with fine-tuned
  generative transformers}.
\newblock In \emph{The 22nd Workshop on Biomedical Natural Language Processing
  and BioNLP Shared Tasks}, pages 326--336, Toronto, Canada. Association for
  Computational Linguistics.

\bibitem[{Jang and Lukasiewicz(2023)}]{paper71}
Myeongjun Jang and Thomas Lukasiewicz. 2023.
\newblock \href {http://arxiv.org/abs/2303.06273} {Consistency analysis of
  ChatGPT}.

\bibitem[{Jentzsch and Kersting(2023)}]{paper202}
Sophie Jentzsch and Kristian Kersting. 2023.
\newblock \href {https://doi.org/10.18653/v1/2023.wassa-1.29} {{C}hat{GPT} is
  fun, but it is not funny! humor is still challenging large language models}.
\newblock In \emph{Proceedings of the 13th Workshop on Computational Approaches
  to Subjectivity, Sentiment, {\&} Social Media Analysis}, pages 325--340,
  Toronto, Canada. Association for Computational Linguistics.

\bibitem[{Jiang et~al.(2023)Jiang, Zhou, Dong, Ye, Zhao, and Wen}]{paper118}
Jinhao Jiang, Kun Zhou, Zican Dong, Keming Ye, Wayne~Xin Zhao, and Ji-Rong Wen.
  2023.
\newblock \href {http://arxiv.org/abs/2305.09645} {StructGPT: A general
  framework for large language model to reason over structured data}.

\bibitem[{Jiao et~al.(2023)Jiao, Wang, Huang, Wang, and Tu}]{paper15}
Wenxiang Jiao, Wenxuan Wang, Jen-tse Huang, Xing Wang, and Zhaopeng Tu. 2023.
\newblock \href {http://arxiv.org/abs/2301.08745} {Is ChatGPT a good
  translator? Yes with GPT-4 as the engine}.

\bibitem[{Jojic et~al.(2023)Jojic, Wang, and Jojic}]{paper108}
Ana Jojic, Zhen Wang, and Nebojsa Jojic. 2023.
\newblock \href {http://arxiv.org/abs/2303.14310} {Gpt is becoming a turing
  machine: Here are some ways to program it}.

\bibitem[{Karpinska and Iyyer(2023)}]{paper160}
Marzena Karpinska and Mohit Iyyer. 2023.
\newblock \href {http://arxiv.org/abs/2304.03245} {Large language models
  effectively leverage document-level context for literary translation, but
  critical errors persist}.

\bibitem[{Kartchner et~al.(2023)Kartchner, Ramalingam, Al-Hussaini, Kronick,
  and Mitchell}]{paper212}
David Kartchner, Selvi Ramalingam, Irfan Al-Hussaini, Olivia Kronick, and
  Cassie Mitchell. 2023.
\newblock \href {https://doi.org/10.18653/v1/2023.bionlp-1.37} {Zero-shot
  information extraction for clinical meta-analysis using large language
  models}.
\newblock In \emph{The 22nd Workshop on Biomedical Natural Language Processing
  and BioNLP Shared Tasks}, pages 396--405, Toronto, Canada. Association for
  Computational Linguistics.

\bibitem[{Kasai et~al.(2023)Kasai, Kasai, Sakaguchi, Yamada, and
  Radev}]{paper149}
Jungo Kasai, Yuhei Kasai, Keisuke Sakaguchi, Yutaro Yamada, and Dragomir Radev.
  2023.
\newblock \href {http://arxiv.org/abs/2303.18027} {Evaluating GPT-4 and ChatGPT
  on japanese medical licensing examinations}.

\bibitem[{Kim et~al.(2023)Kim, Guo, Yu, and Li}]{paper201}
Yuheun Kim, Lu~Guo, Bei Yu, and Yingya Li. 2023.
\newblock \href {https://doi.org/10.18653/v1/2023.wassa-1.33} {Can {C}hat{GPT}
  understand causal language in science claims?}
\newblock In \emph{Proceedings of the 13th Workshop on Computational Approaches
  to Subjectivity, Sentiment, {\&} Social Media Analysis}, pages 379--389,
  Toronto, Canada. Association for Computational Linguistics.

\bibitem[{Kocmi and Federmann(2023)}]{paper25}
Tom Kocmi and Christian Federmann. 2023.
\newblock \href {http://arxiv.org/abs/2302.14520} {Large language models are
  state-of-the-art evaluators of translation quality}.

\bibitem[{Kocoń et~al.(2023)Kocoń, Cichecki, Kaszyca, Kochanek, Szydło,
  Baran, Bielaniewicz, Gruza, Janz, Kanclerz, Kocoń, Koptyra,
  Mieleszczenko-Kowszewicz, Miłkowski, Oleksy, Piasecki, Łukasz Radliński,
  Wojtasik, Woźniak, and Kazienko}]{chatpgt-jack-trades-master-none}
Jan Kocoń, Igor Cichecki, Oliwier Kaszyca, Mateusz Kochanek, Dominika Szydło,
  Joanna Baran, Julita Bielaniewicz, Marcin Gruza, Arkadiusz Janz, Kamil
  Kanclerz, Anna Kocoń, Bartłomiej Koptyra, Wiktoria
  Mieleszczenko-Kowszewicz, Piotr Miłkowski, Marcin Oleksy, Maciej Piasecki,
  Łukasz Radliński, Konrad Wojtasik, Stanisław Woźniak, and Przemysław
  Kazienko. 2023.
\newblock \href {https://doi.org/https://doi.org/10.1016/j.inffus.2023.101861}
  {ChatGPT: Jack of all trades, master of none}.
\newblock \emph{Information Fusion}, 99:101861.

\bibitem[{Koh et~al.(2023)Koh, Plata, and Chai}]{paper90}
Nam~Ho Koh, Joseph Plata, and Joyce Chai. 2023.
\newblock \href {http://arxiv.org/abs/2305.10407} {Bad: Bias detection for
  large language models in the context of candidate screening}.

\bibitem[{Koralus and Wang-Maścianica(2023)}]{paper110}
Philipp Koralus and Vincent Wang-Maścianica. 2023.
\newblock \href {http://arxiv.org/abs/2303.17276} {Humans in humans out: On GPT
  converging toward common sense in both success and failure}.

\bibitem[{Kortemeyer(2023)}]{paper45}
Gerd Kortemeyer. 2023.
\newblock \href {http://arxiv.org/abs/2301.12127} {Could an
  artificial-intelligence agent pass an introductory physics course?}

\bibitem[{Kosinski(2023)}]{paper35}
Michal Kosinski. 2023.
\newblock \href {http://arxiv.org/abs/2302.02083} {Theory of mind might have
  spontaneously emerged in large language models}.

\bibitem[{Kumar et~al.(2023)Kumar, Balachandran, Njoo, Anastasopoulos, and
  Tsvetkov}]{paper220}
Sachin Kumar, Vidhisha Balachandran, Lucille Njoo, Antonios Anastasopoulos, and
  Yulia Tsvetkov. 2023.
\newblock \href {https://doi.org/10.18653/v1/2023.eacl-main.241} {Language
  generation models can cause harm: So what can we do about it? an actionable
  survey}.
\newblock In \emph{Proceedings of the 17th Conference of the European Chapter
  of the Association for Computational Linguistics}, pages 3299--3321,
  Dubrovnik, Croatia. Association for Computational Linguistics.

\bibitem[{Kung et~al.(2023)Kung, Cheatham, Medenilla, Sillos, De~Leon,
  Elepaño, Madriaga, Aggabao, Diaz-Candido, Maningo, and Tseng}]{paper52}
Tiffany~H. Kung, Morgan Cheatham, Arielle Medenilla, Czarina Sillos, Lorie
  De~Leon, Camille Elepaño, Maria Madriaga, Rimel Aggabao, Giezel
  Diaz-Candido, James Maningo, and Victor Tseng. 2023.
\newblock \href {https://doi.org/10.1371/journal.pdig.0000198} {Performance of
  ChatGPT on usmle: Potential for AI-assisted medical education using large
  language models}.
\newblock \emph{PLOS Digital Health}, 2(2):1--12.

\bibitem[{Kuo et~al.(2023)Kuo, Hsueh, and Tsai}]{paper135}
Mu-Tien Kuo, Chih-Chung Hsueh, and Richard Tzong-Han Tsai. 2023.
\newblock \href {http://arxiv.org/abs/2308.15118} {Large language models on the
  chessboard: A study on ChatGPT's formal language comprehension and complex
  reasoning skills}.

\bibitem[{Kıcıman et~al.(2023)Kıcıman, Ness, Sharma, and Tan}]{paper34}
Emre Kıcıman, Robert Ness, Amit Sharma, and Chenhao Tan. 2023.
\newblock \href {http://arxiv.org/abs/2305.00050} {Causal reasoning and large
  language models: Opening a new frontier for causality}.

\bibitem[{Lai et~al.(2023)Lai, Ngo, Veyseh, Man, Dernoncourt, Bui, and
  Nguyen}]{paper29}
Viet~Dac Lai, Nghia~Trung Ngo, Amir Pouran~Ben Veyseh, Hieu Man, Franck
  Dernoncourt, Trung Bui, and Thien~Huu Nguyen. 2023.
\newblock \href {http://arxiv.org/abs/2304.05613} {ChatGPT beyond english:
  Towards a comprehensive evaluation of large language models in multilingual
  learning}.

\bibitem[{Laskar et~al.(2023)Laskar, Bari, Rahman, Bhuiyan, Joty, and
  Huang}]{paper6}
Md~Tahmid~Rahman Laskar, M~Saiful Bari, Mizanur Rahman, Md~Amran~Hossen
  Bhuiyan, Shafiq Joty, and Jimmy~Xiangji Huang. 2023.
\newblock \href {http://arxiv.org/abs/2305.18486} {A systematic study and
  comprehensive evaluation of ChatGPT on benchmark datasets}.

\bibitem[{Leiter et~al.(2023)Leiter, Zhang, Chen, Belouadi, Larionov, Fresen,
  and Eger}]{paper59}
Christoph Leiter, Ran Zhang, Yanran Chen, Jonas Belouadi, Daniil Larionov,
  Vivian Fresen, and Steffen Eger. 2023.
\newblock \href {http://arxiv.org/abs/2302.13795} {ChatGPT: A meta-analysis
  after 2.5 months}.

\bibitem[{Leong et~al.(2023)Leong, Ngui, Susanto, Rengarajan, Sarveswaran, and
  Tjhi}]{paper165}
Wei~Qi Leong, Jian~Gang Ngui, Yosephine Susanto, Hamsawardhini Rengarajan,
  Kengatharaiyer Sarveswaran, and William~Chandra Tjhi. 2023.
\newblock \href {http://arxiv.org/abs/2309.06085} {Bhasa: A holistic southeast
  asian linguistic and cultural evaluation suite for large language models}.

\bibitem[{Li et~al.(2023{\natexlab{a}})Li, Fang, Yang, Wang, Ye, Zhao, and
  Zhang}]{paper145}
Bo~Li, Gexiang Fang, Yang Yang, Quansen Wang, Wei Ye, Wen Zhao, and Shikun
  Zhang. 2023{\natexlab{a}}.
\newblock \href {http://arxiv.org/abs/2304.11633} {Evaluating ChatGPT's
  information extraction capabilities: An assessment of performance,
  explainability, calibration, and faithfulness}.

\bibitem[{Li et~al.(2023{\natexlab{b}})Li, Leng, Yan, Shen, Wang, MI, Fei,
  Feng, Yan, Wang, Zhan, Jia, Wu, and Sun}]{paper123}
Cheng Li, Ziang Leng, Chenxi Yan, Junyi Shen, Hao Wang, Weishi MI, Yaying Fei,
  Xiaoyang Feng, Song Yan, HaoSheng Wang, Linkang Zhan, Yaokai Jia, Pingyu Wu,
  and Haozhen Sun. 2023{\natexlab{b}}.
\newblock \href {http://arxiv.org/abs/2308.09597} {ChatHaruhi: Reviving anime
  character in reality via large language model}.

\bibitem[{Li et~al.(2023{\natexlab{c}})Li, Hui, Qu, Li, Yang, Li, Wang, Qin,
  Cao, Geng, Huo, Zhou, Ma, Li, Chang, Huang, Cheng, and Li}]{paper43}
Jinyang Li, Binyuan Hui, Ge~Qu, Binhua Li, Jiaxi Yang, Bowen Li, Bailin Wang,
  Bowen Qin, Rongyu Cao, Ruiying Geng, Nan Huo, Xuanhe Zhou, Chenhao Ma,
  Guoliang Li, Kevin C.~C. Chang, Fei Huang, Reynold Cheng, and Yongbin Li.
  2023{\natexlab{c}}.
\newblock \href {http://arxiv.org/abs/2305.03111} {Can LLM already serve as a
  database interface? a big bench for large-scale database grounded
  text-to-sqls}.

\bibitem[{Li et~al.(2023{\natexlab{d}})Li, Cheng, Zhao, Nie, and Wen}]{paper96}
Junyi Li, Xiaoxue Cheng, Wayne~Xin Zhao, Jian-Yun Nie, and Ji-Rong Wen.
  2023{\natexlab{d}}.
\newblock \href {http://arxiv.org/abs/2305.11747} {HaluEval: A large-scale
  hallucination evaluation benchmark for large language models}.

\bibitem[{Li et~al.(2023{\natexlab{e}})Li, Fan, Atreja, and
  Hemphill}]{paper159}
Lingyao Li, Lizhou Fan, Shubham Atreja, and Libby Hemphill. 2023{\natexlab{e}}.
\newblock \href {http://arxiv.org/abs/2304.10619} {"hot" ChatGPT: The promise
  of ChatGPT in detecting and discriminating hateful, offensive, and toxic
  comments on social media}.

\bibitem[{Li et~al.(2023{\natexlab{f}})Li, Zhao, Chia, Ding, Joty, Poria, and
  Bing}]{paper122}
Xingxuan Li, Ruochen Zhao, Yew~Ken Chia, Bosheng Ding, Shafiq Joty, Soujanya
  Poria, and Lidong Bing. 2023{\natexlab{f}}.
\newblock \href {http://arxiv.org/abs/2305.13269} {Chain-of-knowledge:
  Grounding large language models via dynamic knowledge adapting over
  heterogeneous sources}.

\bibitem[{Li et~al.(2023{\natexlab{g}})Li, Peng, He, Galley, Gao, and
  Yan}]{paper191}
Zekun Li, Baolin Peng, Pengcheng He, Michel Galley, Jianfeng Gao, and Xifeng
  Yan. 2023{\natexlab{g}}.
\newblock \href {http://arxiv.org/abs/2302.11520} {Guiding large language
  models via directional stimulus prompting}.

\bibitem[{Liang et~al.(2023{\natexlab{a}})Liang, Bommasani, Lee, Tsipras,
  Soylu, Yasunaga, Zhang, Narayanan, Wu, Kumar, Newman, Yuan, Yan, Zhang,
  Cosgrove, Manning, Re, Acosta-Navas, Hudson, Zelikman, Durmus, Ladhak, Rong,
  Ren, Yao, WANG, Santhanam, Orr, Zheng, Yuksekgonul, Suzgun, Kim, Guha,
  Chatterji, Khattab, Henderson, Huang, Chi, Xie, Santurkar, Ganguli,
  Hashimoto, Icard, Zhang, Chaudhary, Wang, Li, Mai, Zhang, and
  Koreeda}]{paper63}
Percy Liang, Rishi Bommasani, Tony Lee, Dimitris Tsipras, Dilara Soylu,
  Michihiro Yasunaga, Yian Zhang, Deepak Narayanan, Yuhuai Wu, Ananya Kumar,
  Benjamin Newman, Binhang Yuan, Bobby Yan, Ce~Zhang, Christian~Alexander
  Cosgrove, Christopher~D Manning, Christopher Re, Diana Acosta-Navas,
  Drew~Arad Hudson, Eric Zelikman, Esin Durmus, Faisal Ladhak, Frieda Rong,
  Hongyu Ren, Huaxiu Yao, Jue WANG, Keshav Santhanam, Laurel Orr, Lucia Zheng,
  Mert Yuksekgonul, Mirac Suzgun, Nathan Kim, Neel Guha, Niladri~S. Chatterji,
  Omar Khattab, Peter Henderson, Qian Huang, Ryan~Andrew Chi, Sang~Michael Xie,
  Shibani Santurkar, Surya Ganguli, Tatsunori Hashimoto, Thomas Icard, Tianyi
  Zhang, Vishrav Chaudhary, William Wang, Xuechen Li, Yifan Mai, Yuhui Zhang,
  and Yuta Koreeda. 2023{\natexlab{a}}.
\newblock \href {https://openreview.net/forum?id=iO4LZibEqW} {Holistic
  evaluation of language models}.
\newblock \emph{Transactions on Machine Learning Research}.
\newblock Featured Certification, Expert Certification.

\bibitem[{Liang et~al.(2023{\natexlab{b}})Liang, Zhang, Li, Liu, Hu, Wu, Zhang,
  Liu, and Wu}]{paper207}
Yancheng Liang, Jiajie Zhang, Hui Li, Xiaochen Liu, Yi~Hu, Yong Wu, Jiaoyao
  Zhang, Yongyan Liu, and Yi~Wu. 2023{\natexlab{b}}.
\newblock \href {https://aclanthology.org/2023.finnlp-1.7} {Breaking the bank
  with {C}hat{GPT}: Few-shot text classification for finance}.
\newblock In \emph{Proceedings of the Fifth Workshop on Financial Technology
  and Natural Language Processing and the Second Multimodal AI For Financial
  Forecasting}, pages 74--80, Macao. -.

\bibitem[{Liu et~al.(2023{\natexlab{a}})Liu, Hu, Wen, and Yu}]{paper74}
Aiwei Liu, Xuming Hu, Lijie Wen, and Philip~S. Yu. 2023{\natexlab{a}}.
\newblock \href {http://arxiv.org/abs/2303.13547} {A comprehensive evaluation
  of ChatGPT's zero-shot text-to-sql capability}.

\bibitem[{Liu et~al.(2023{\natexlab{b}})Liu, Wu, Michael, Suhr, West, Koller,
  Swayamdipta, Smith, and Choi}]{paper40}
Alisa Liu, Zhaofeng Wu, Julian Michael, Alane Suhr, Peter West, Alexander
  Koller, Swabha Swayamdipta, Noah~A. Smith, and Yejin Choi.
  2023{\natexlab{b}}.
\newblock \href {http://arxiv.org/abs/2304.14399} {We're afraid language models
  aren't modeling ambiguity}.

\bibitem[{Liu and Wu(2023)}]{paper136}
Chang Liu and Bo~Wu. 2023.
\newblock \href {http://arxiv.org/abs/2308.11224} {Evaluating large language
  models on graphs: Performance insights and comparative analysis}.

\bibitem[{Liu et~al.(2023{\natexlab{c}})Liu, Ning, Teng, Liu, Zhou, and
  Zhang}]{paper39}
Hanmeng Liu, Ruoxi Ning, Zhiyang Teng, Jian Liu, Qiji Zhou, and Yue Zhang.
  2023{\natexlab{c}}.
\newblock \href {http://arxiv.org/abs/2304.03439} {Evaluating the logical
  reasoning ability of ChatGPT and GPT-4}.

\bibitem[{Liu et~al.(2023{\natexlab{d}})Liu, Teng, Cui, Zhang, Zhou, and
  Zhang}]{paper126}
Hanmeng Liu, Zhiyang Teng, Leyang Cui, Chaoli Zhang, Qiji Zhou, and Yue Zhang.
  2023{\natexlab{d}}.
\newblock \href {http://arxiv.org/abs/2305.12147} {Logicot: Logical
  chain-of-thought instruction-tuning data collection with GPT-4}.

\bibitem[{Liu et~al.(2023{\natexlab{e}})Liu, Xia, Wang, and Zhang}]{paper144}
Jiawei Liu, Chunqiu~Steven Xia, Yuyao Wang, and Lingming Zhang.
  2023{\natexlab{e}}.
\newblock \href {http://arxiv.org/abs/2305.01210} {Is your code generated by
  ChatGPT really correct? rigorous evaluation of large language models for code
  generation}.

\bibitem[{Liu et~al.(2023{\natexlab{f}})Liu, Tan, Xiao, Zhuge, and
  Zhou}]{paper196}
Xin Liu, Yuan Tan, Zhenghang Xiao, Jianwei Zhuge, and Rui Zhou.
  2023{\natexlab{f}}.
\newblock \href {https://doi.org/10.18653/v1/2023.findings-acl.229} {Not the
  end of story: An evaluation of {C}hat{GPT}-driven vulnerability description
  mappings}.
\newblock In \emph{Findings of the Association for Computational Linguistics:
  ACL 2023}, pages 3724--3731, Toronto, Canada. Association for Computational
  Linguistics.

\bibitem[{Liu et~al.(2023{\natexlab{g}})Liu, Iter, Xu, Wang, Xu, and
  Zhu}]{paper24}
Yang Liu, Dan Iter, Yichong Xu, Shuohang Wang, Ruochen Xu, and Chenguang Zhu.
  2023{\natexlab{g}}.
\newblock \href {http://arxiv.org/abs/2303.16634} {G-Eval: NLG evaluation using
  GPT-4 with better human alignment}.

\bibitem[{Liu et~al.(2023{\natexlab{h}})Liu, Han, Ma, Zhang, Yang, Tian, He,
  Li, He, Liu, Wu, Zhao, Zhu, Li, Qiang, Shen, Liu, and Ge}]{paper61}
Yiheng Liu, Tianle Han, Siyuan Ma, Jiayue Zhang, Yuanyuan Yang, Jiaming Tian,
  Hao He, Antong Li, Mengshen He, Zhengliang Liu, Zihao Wu, Lin Zhao, Dajiang
  Zhu, Xiang Li, Ning Qiang, Dingang Shen, Tianming Liu, and Bao Ge.
  2023{\natexlab{h}}.
\newblock \href {http://arxiv.org/abs/2304.01852} {Summary of ChatGPT-related
  research and perspective towards the future of large language models}.

\bibitem[{Liu et~al.(2023{\natexlab{i}})Liu, Zhang, Xia, Wu, Xie, Qin, Zhang,
  and Liu}]{paper217}
Zequn Liu, Wei Zhang, Yingce Xia, Lijun Wu, Shufang Xie, Tao Qin, Ming Zhang,
  and Tie-Yan Liu. 2023{\natexlab{i}}.
\newblock \href {https://doi.org/10.18653/v1/2023.acl-short.138} {{M}ol{XPT}:
  Wrapping molecules with text for generative pre-training}.
\newblock In \emph{Proceedings of the 61st Annual Meeting of the Association
  for Computational Linguistics (Volume 2: Short Papers)}, pages 1606--1616,
  Toronto, Canada. Association for Computational Linguistics.

\bibitem[{Liu et~al.(2023{\natexlab{j}})Liu, Yu, Zhang, Wu, Cao, Dai, Zhao,
  Liu, Shen, Li, Liu, Zhu, and Li}]{paper175}
Zhengliang Liu, Xiaowei Yu, Lu~Zhang, Zihao Wu, Chao Cao, Haixing Dai, Lin
  Zhao, Wei Liu, Dinggang Shen, Quanzheng Li, Tianming Liu, Dajiang Zhu, and
  Xiang Li. 2023{\natexlab{j}}.
\newblock \href {http://arxiv.org/abs/2303.11032} {Deid-GPT: Zero-shot medical
  text de-identification by GPT-4}.

\bibitem[{Liu et~al.(2023{\natexlab{k}})Liu, Litman, Wang, Matsumura, and
  Correnti}]{paper240}
Zhexiong Liu, Diane Litman, Elaine Wang, Lindsay Matsumura, and Richard
  Correnti. 2023{\natexlab{k}}.
\newblock \href {https://doi.org/10.18653/v1/2023.bea-1.24} {Predicting the
  quality of revisions in argumentative writing}.
\newblock In \emph{Proceedings of the 18th Workshop on Innovative Use of NLP
  for Building Educational Applications (BEA 2023)}, pages 275--287, Toronto,
  Canada. Association for Computational Linguistics.

\bibitem[{Loem et~al.(2023)Loem, Kaneko, Takase, and Okazaki}]{paper241}
Mengsay Loem, Masahiro Kaneko, Sho Takase, and Naoaki Okazaki. 2023.
\newblock \href {https://doi.org/10.18653/v1/2023.bea-1.18} {Exploring
  effectiveness of {GPT}-3 in grammatical error correction: A study on
  performance and controllability in prompt-based methods}.
\newblock In \emph{Proceedings of the 18th Workshop on Innovative Use of NLP
  for Building Educational Applications (BEA 2023)}, pages 205--219, Toronto,
  Canada. Association for Computational Linguistics.

\bibitem[{Lu et~al.(2023{\natexlab{a}})Lu, Larcher, and Tran}]{paper99}
Guang Lu, Sylvia~B. Larcher, and Tu~Tran. 2023{\natexlab{a}}.
\newblock \href {http://arxiv.org/abs/2306.01169} {Hybrid long document
  summarization using c2f-far and ChatGPT: A practical study}.

\bibitem[{Lu et~al.(2023{\natexlab{b}})Lu, Ding, Xie, Zhang, Wong, and
  Tao}]{paper215}
Qingyu Lu, Liang Ding, Liping Xie, Kanjian Zhang, Derek~F. Wong, and Dacheng
  Tao. 2023{\natexlab{b}}.
\newblock \href {https://doi.org/10.18653/v1/2023.acl-long.324} {Toward
  human-like evaluation for natural language generation with error analysis}.
\newblock In \emph{Proceedings of the 61st Annual Meeting of the Association
  for Computational Linguistics (Volume 1: Long Papers)}, pages 5892--5907,
  Toronto, Canada. Association for Computational Linguistics.

\bibitem[{Lu et~al.(2023{\natexlab{c}})Lu, Qiu, Ding, Zhang, Kocmi, and
  Tao}]{paper140}
Qingyu Lu, Baopu Qiu, Liang Ding, Kanjian Zhang, Tom Kocmi, and Dacheng Tao.
  2023{\natexlab{c}}.
\newblock \href {http://arxiv.org/abs/2303.13809} {Error analysis prompting
  enables human-like translation evaluation in large language models: A case
  study on ChatGPT}.

\bibitem[{Lu et~al.(2023{\natexlab{d}})Lu, Li, Tong, Zhao, and Qin}]{paper238}
Xin Lu, Zhuojun Li, Yanpeng Tong, Yanyan Zhao, and Bing Qin.
  2023{\natexlab{d}}.
\newblock \href {https://doi.org/10.18653/v1/2023.wassa-1.54} {{HIT}-{SCIR} at
  {WASSA} 2023: Empathy and emotion analysis at the utterance-level and the
  essay-level}.
\newblock In \emph{Proceedings of the 13th Workshop on Computational Approaches
  to Subjectivity, Sentiment, {\&} Social Media Analysis}, pages 574--580,
  Toronto, Canada. Association for Computational Linguistics.

\bibitem[{Luo et~al.(2023)Luo, Xie, and Ananiadou}]{paper94}
Zheheng Luo, Qianqian Xie, and Sophia Ananiadou. 2023.
\newblock \href {http://arxiv.org/abs/2303.15621} {ChatGPT as a factual
  inconsistency evaluator for text summarization}.

\bibitem[{Lyu et~al.(2023{\natexlab{a}})Lyu, Jiang, Zeng, Wang, Zhang, Chen,
  Leung, Tang, Xia, and Luo}]{paper103}
Hanjia Lyu, Song Jiang, Hanqing Zeng, Qifan Wang, Si~Zhang, Ren Chen, Chris
  Leung, Jiajie Tang, Yinglong Xia, and Jiebo Luo. 2023{\natexlab{a}}.
\newblock \href {http://arxiv.org/abs/2307.15780} {LLM-Rec: Personalized
  recommendation via prompting large language models}.

\bibitem[{Lyu et~al.(2023{\natexlab{b}})Lyu, Tan, Zapadka, Ponnatapura, Niu,
  Myers, Wang, and Whitlow}]{paper173}
Qing Lyu, Josh Tan, Michael~E. Zapadka, Janardhana Ponnatapura, Chuang Niu,
  Kyle~J. Myers, Ge~Wang, and Christopher~T. Whitlow. 2023{\natexlab{b}}.
\newblock \href {http://arxiv.org/abs/2303.09038} {Translating radiology
  reports into plain language using ChatGPT and GPT-4 with prompt learning:
  Promising results, limitations, and potential}.

\bibitem[{Maddigan and Susnjak(2023)}]{paper188}
Paula Maddigan and Teo Susnjak. 2023.
\newblock \href {http://arxiv.org/abs/2302.02094} {Chat2vis: Generating data
  visualisations via natural language using ChatGPT, codex and GPT-3 large
  language models}.

\bibitem[{Manakul et~al.(2023)Manakul, Liusie, and Gales}]{paper172}
Potsawee Manakul, Adian Liusie, and Mark J.~F. Gales. 2023.
\newblock \href {http://arxiv.org/abs/2303.08896} {SelfCheckGPT: Zero-resource
  black-box hallucination detection for generative large language models}.

\bibitem[{Mao et~al.(2023)Mao, Chen, Zhang, Guerin, and Cambria}]{paper5}
Rui Mao, Guanyi Chen, Xulang Zhang, Frank Guerin, and Erik Cambria. 2023.
\newblock \href {http://arxiv.org/abs/2308.12488} {Gpteval: A survey on
  assessments of ChatGPT and GPT-4}.

\bibitem[{Mehnen et~al.(2023)Mehnen, Gruarin, Vasileva, and Knapp}]{paper56}
Lars Mehnen, Stefanie Gruarin, Mina Vasileva, and Bernhard Knapp. 2023.
\newblock \href {https://doi.org/10.1101/2023.04.20.23288859} {ChatGPT as a
  medical doctor? a diagnostic accuracy study on common and rare diseases}.
\newblock \emph{medRxiv}.

\bibitem[{Michail et~al.(2023)Michail, Konstantinou, and Clematide}]{paper167}
Andrianos Michail, Stefanos Konstantinou, and Simon Clematide. 2023.
\newblock \href {http://arxiv.org/abs/2303.01194} {Uzh\_clyp at semeval-2023
  task 9: Head-first fine-tuning and ChatGPT data generation for cross-lingual
  learning in tweet intimacy prediction}.

\bibitem[{Moghaddam and Honey(2023)}]{paper36}
Shima~Rahimi Moghaddam and Christopher~J. Honey. 2023.
\newblock \href {http://arxiv.org/abs/2304.11490} {Boosting theory-of-mind
  performance in large language models via prompting}.

\bibitem[{Morabito et~al.(2023)Morabito, Kabbara, and Emami}]{paper243}
Robert Morabito, Jad Kabbara, and Ali Emami. 2023.
\newblock \href {https://doi.org/10.18653/v1/2023.findings-acl.280} {Debiasing
  should be good and bad: Measuring the consistency of debiasing techniques in
  language models}.
\newblock In \emph{Findings of the Association for Computational Linguistics:
  ACL 2023}, pages 4581--4597, Toronto, Canada. Association for Computational
  Linguistics.

\bibitem[{Murahari et~al.(2023)Murahari, Deshpande, Jimenez, Shafran, Wang,
  Cao, and Narasimhan}]{paper218}
Vishvak Murahari, Ameet Deshpande, Carlos Jimenez, Izhak Shafran, Mingqiu Wang,
  Yuan Cao, and Karthik Narasimhan. 2023.
\newblock \href {https://doi.org/10.18653/v1/2023.repl4nlp-1.17} {{MUX}-{PLM}s:
  Pre-training language models with data multiplexing}.
\newblock In \emph{Proceedings of the 8th Workshop on Representation Learning
  for NLP (RepL4NLP 2023)}, pages 196--211, Toronto, Canada. Association for
  Computational Linguistics.

\bibitem[{Nan(2023)}]{paper252}
Duan Nan. 2023.
\newblock \href {https://aclanthology.org/2023.ccl-2.9} {Frontier review of
  multimodal {AI}}.
\newblock In \emph{Proceedings of the 22nd Chinese National Conference on
  Computational Linguistics (Volume 2: Frontier Forum)}, pages 110--118,
  Harbin, China. Chinese Information Processing Society of China.

\bibitem[{Nedilko(2023)}]{paper228}
Andrew Nedilko. 2023.
\newblock \href {https://doi.org/10.18653/v1/2023.wassa-1.61} {Generative
  pretrained transformers for emotion detection in a code-switching setting}.
\newblock In \emph{Proceedings of the 13th Workshop on Computational Approaches
  to Subjectivity, Sentiment, {\&} Social Media Analysis}, pages 616--620,
  Toronto, Canada. Association for Computational Linguistics.

\bibitem[{Nori et~al.(2023)Nori, King, McKinney, Carignan, and
  Horvitz}]{paper107}
Harsha Nori, Nicholas King, Scott~Mayer McKinney, Dean Carignan, and Eric
  Horvitz. 2023.
\newblock \href {http://arxiv.org/abs/2303.13375} {Capabilities of GPT-4 on
  medical challenge problems}.

\bibitem[{Oh et~al.(2023)Oh, Go, Moon, Lee, Jeong, Lee, and Choi}]{paper236}
Shinhyeok Oh, Hyojun Go, Hyeongdon Moon, Yunsung Lee, Myeongho Jeong,
  Hyun~Seung Lee, and Seungtaek Choi. 2023.
\newblock \href {https://doi.org/10.18653/v1/2023.findings-acl.396} {Evaluation
  of question generation needs more references}.
\newblock In \emph{Findings of the Association for Computational Linguistics:
  ACL 2023}, pages 6358--6367, Toronto, Canada. Association for Computational
  Linguistics.

\bibitem[{Omar et~al.(2023)Omar, Mangukiya, Kalnis, and Mansour}]{paper189}
Reham Omar, Omij Mangukiya, Panos Kalnis, and Essam Mansour. 2023.
\newblock \href {http://arxiv.org/abs/2302.06466} {ChatGPT versus traditional
  question answering for knowledge graphs: Current status and future directions
  towards knowledge graph chatbots}.

\bibitem[{Omidvar and An(2023)}]{paper225}
Amin Omidvar and Aijun An. 2023.
\newblock \href {https://doi.org/10.18653/v1/2023.bea-1.62} {Empowering
  conversational agents using semantic in-context learning}.
\newblock In \emph{Proceedings of the 18th Workshop on Innovative Use of NLP
  for Building Educational Applications (BEA 2023)}, pages 766--771, Toronto,
  Canada. Association for Computational Linguistics.

\bibitem[{OpenAI(2023)}]{paper10}
OpenAI. 2023.
\newblock \href {http://arxiv.org/abs/2303.08774} {GPT-4 technical report}.

\bibitem[{Ortega-Martín et~al.(2023)Ortega-Martín, García-Sierra, Ardoiz,
  Álvarez, Armenteros, and Alonso}]{paper153}
Miguel Ortega-Martín, Óscar García-Sierra, Alfonso Ardoiz, Jorge Álvarez,
  Juan~Carlos Armenteros, and Adrián Alonso. 2023.
\newblock \href {http://arxiv.org/abs/2302.06426} {Linguistic ambiguity
  analysis in ChatGPT}.

\bibitem[{Ostyakova et~al.(2023)Ostyakova, Smilga, Petukhova, Molchanova, and
  Kornev}]{paper205}
Lidiia Ostyakova, Veronika Smilga, Kseniia Petukhova, Maria Molchanova, and
  Daniel Kornev. 2023.
\newblock \href {https://aclanthology.org/2023.sigdial-1.23} {{C}hat{GPT} vs.
  crowdsourcing vs. experts: Annotating open-domain conversations with speech
  functions}.
\newblock In \emph{Proceedings of the 24th Meeting of the Special Interest
  Group on Discourse and Dialogue}, pages 242--254, Prague, Czechia.
  Association for Computational Linguistics.

\bibitem[{Otani et~al.(2023)Otani, Araki, Kim, and Hovy}]{paper229}
Naoki Otani, Jun Araki, HyeongSik Kim, and Eduard Hovy. 2023.
\newblock \href {https://doi.org/10.18653/v1/2023.nlp4convai-1.2} {On the
  underspecification of situations in open-domain conversational datasets}.
\newblock In \emph{Proceedings of the 5th Workshop on NLP for Conversational AI
  (NLP4ConvAI 2023)}, pages 12--28, Toronto, Canada. Association for
  Computational Linguistics.

\bibitem[{Pan et~al.(2023)Pan, Chen, Xu, Che, and Qin}]{paper76}
Wenbo Pan, Qiguang Chen, Xiao Xu, Wanxiang Che, and Libo Qin. 2023.
\newblock \href {http://arxiv.org/abs/2304.04256} {A preliminary evaluation of
  ChatGPT for zero-shot dialogue understanding}.

\bibitem[{Peeters and Bizer(2023)}]{paper154}
Ralph Peeters and Christian Bizer. 2023.
\newblock \href {http://arxiv.org/abs/2305.03423} {Using ChatGPT for entity
  matching}.

\bibitem[{Pegoraro et~al.(2023)Pegoraro, Kumari, Fereidooni, and
  Sadeghi}]{paper150}
Alessandro Pegoraro, Kavita Kumari, Hossein Fereidooni, and Ahmad-Reza Sadeghi.
  2023.
\newblock \href {http://arxiv.org/abs/2304.01487} {To ChatGPT, or not to
  ChatGPT: That is the question!}

\bibitem[{Peng et~al.(2023{\natexlab{a}})Peng, Galley, He, Cheng, Xie, Hu,
  Huang, Liden, Yu, Chen, and Gao}]{paper193}
Baolin Peng, Michel Galley, Pengcheng He, Hao Cheng, Yujia Xie, Yu~Hu, Qiuyuan
  Huang, Lars Liden, Zhou Yu, Weizhu Chen, and Jianfeng Gao.
  2023{\natexlab{a}}.
\newblock \href {http://arxiv.org/abs/2302.12813} {Check your facts and try
  again: Improving large language models with external knowledge and automated
  feedback}.

\bibitem[{Peng et~al.(2023{\natexlab{b}})Peng, Li, He, Galley, and
  Gao}]{paper69}
Baolin Peng, Chunyuan Li, Pengcheng He, Michel Galley, and Jianfeng Gao.
  2023{\natexlab{b}}.
\newblock \href {http://arxiv.org/abs/2304.03277} {Instruction tuning with
  GPT-4}.

\bibitem[{Peng et~al.(2023{\natexlab{c}})Peng, Ding, Zhong, Shen, Liu, Zhang,
  Ouyang, and Tao}]{paper141}
Keqin Peng, Liang Ding, Qihuang Zhong, Li~Shen, Xuebo Liu, Min Zhang, Yuanxin
  Ouyang, and Dacheng Tao. 2023{\natexlab{c}}.
\newblock \href {http://arxiv.org/abs/2303.13780} {Towards making the most of
  ChatGPT for machine translation}.

\bibitem[{Peskoff and Stewart(2023)}]{paper209}
Denis Peskoff and Brandon Stewart. 2023.
\newblock \href {https://doi.org/10.18653/v1/2023.acl-short.37} {Credible
  without credit: Domain experts assess generative language models}.
\newblock In \emph{Proceedings of the 61st Annual Meeting of the Association
  for Computational Linguistics (Volume 2: Short Papers)}, pages 427--438,
  Toronto, Canada. Association for Computational Linguistics.

\bibitem[{Pezeshkpour and Hruschka(2023)}]{paper134}
Pouya Pezeshkpour and Estevam Hruschka. 2023.
\newblock \href {http://arxiv.org/abs/2308.11483} {Large language models
  sensitivity to the order of options in multiple-choice questions}.

\bibitem[{Popescu-Belis et~al.(2023)Popescu-Belis, Atrio, Bernath, Boisson,
  Ferrari, Theimer-lienhard, and Vernikos}]{paper244}
Andrei Popescu-Belis, {\`A}lex~R. Atrio, Bastien Bernath, Etienne Boisson, Teo
  Ferrari, Xavier Theimer-lienhard, and Giorgos Vernikos. 2023.
\newblock \href {https://doi.org/10.18653/v1/2023.latechclfl-1.2} {{GP}oe{T}: a
  language model trained for rhyme generation on synthetic data}.
\newblock In \emph{Proceedings of the 7th Joint SIGHUM Workshop on
  Computational Linguistics for Cultural Heritage, Social Sciences, Humanities
  and Literature}, pages 10--20, Dubrovnik, Croatia. Association for
  Computational Linguistics.

\bibitem[{Pu and Demberg(2023)}]{paper195}
Dongqi Pu and Vera Demberg. 2023.
\newblock \href {https://doi.org/10.18653/v1/2023.acl-srw.1} {{C}hat{GPT} vs
  human-authored text: Insights into controllable text summarization and
  sentence style transfer}.
\newblock In \emph{Proceedings of the 61st Annual Meeting of the Association
  for Computational Linguistics (Volume 4: Student Research Workshop)}, pages
  1--18, Toronto, Canada. Association for Computational Linguistics.

\bibitem[{Qin et~al.(2023)Qin, Zhang, Zhang, Chen, Yasunaga, and
  Yang}]{paper16}
Chengwei Qin, Aston Zhang, Zhuosheng Zhang, Jiaao Chen, Michihiro Yasunaga, and
  Diyi Yang. 2023.
\newblock \href {http://arxiv.org/abs/2302.06476} {Is ChatGPT a general-purpose
  natural language processing task solver?}

\bibitem[{Quidwai et~al.(2023)Quidwai, Li, and Dube}]{paper178}
Ali Quidwai, Chunhui Li, and Parijat Dube. 2023.
\newblock \href {https://doi.org/10.18653/v1/2023.bea-1.58} {Beyond black box
  {AI} generated plagiarism detection: From sentence to document level}.
\newblock In \emph{Proceedings of the 18th Workshop on Innovative Use of NLP
  for Building Educational Applications (BEA 2023)}, pages 727--735, Toronto,
  Canada. Association for Computational Linguistics.

\bibitem[{Rajasekharan et~al.(2023)Rajasekharan, Zeng, Padalkar, and
  Gupta}]{paper183}
Abhiramon Rajasekharan, Yankai Zeng, Parth Padalkar, and Gopal Gupta. 2023.
\newblock \href {http://arxiv.org/abs/2302.03780} {Reliable natural language
  understanding with large language models and answer set programming}.

\bibitem[{Rangapur and Wang(2023)}]{paper114}
Aman Rangapur and Haoran Wang. 2023.
\newblock \href {http://arxiv.org/abs/2304.03325} {ChatGPT-crawler: Find out if
  ChatGPT really knows what it's talking about}.

\bibitem[{Rao et~al.(2023{\natexlab{a}})Rao, Pang, Kim, Kamineni, Lie, Prasad,
  Landman, Dreyer, and Succi}]{paper55}
Arya Rao, Michael Pang, John Kim, Meghana Kamineni, Winston Lie, Anoop~K.
  Prasad, Adam Landman, Keith~J Dreyer, and Marc~D. Succi. 2023{\natexlab{a}}.
\newblock \href {https://doi.org/10.1101/2023.02.21.23285886} {Assessing the
  utility of ChatGPT throughout the entire clinical workflow}.
\newblock \emph{medRxiv}.

\bibitem[{Rao et~al.(2023{\natexlab{b}})Rao, Leung, and Miao}]{paper146}
Haocong Rao, Cyril Leung, and Chunyan Miao. 2023{\natexlab{b}}.
\newblock \href {http://arxiv.org/abs/2303.01248} {Can ChatGPT Assess Human Personalities? A General Evaluation Framework}.

\bibitem[{Ravaut et~al.(2023)Ravaut, Joty, and Chen}]{paper246}
Mathieu Ravaut, Shafiq Joty, and Nancy Chen. 2023.
\newblock \href {https://doi.org/10.18653/v1/2023.findings-acl.529}
  {Unsupervised summarization re-ranking}.
\newblock In \emph{Findings of the Association for Computational Linguistics:
  ACL 2023}, pages 8341--8376, Toronto, Canada. Association for Computational
  Linguistics.

\bibitem[{Razzhigaev et~al.(2023)Razzhigaev, Salnikov, Malykh, Braslavski, and
  Panchenko}]{paper237}
Anton Razzhigaev, Mikhail Salnikov, Valentin Malykh, Pavel Braslavski, and
  Alexander Panchenko. 2023.
\newblock \href {https://doi.org/10.18653/v1/2023.acl-demo.51} {A system for
  answering simple questions in multiple languages}.
\newblock In \emph{Proceedings of the 61st Annual Meeting of the Association
  for Computational Linguistics (Volume 3: System Demonstrations)}, pages
  524--537, Toronto, Canada. Association for Computational Linguistics.

\bibitem[{Ren et~al.(2023)Ren, Wang, Qu, Zhao, Liu, Tian, Wu, Wen, and
  Wang}]{paper130}
Ruiyang Ren, Yuhao Wang, Yingqi Qu, Wayne~Xin Zhao, Jing Liu, Hao Tian, Hua Wu,
  Ji-Rong Wen, and Haifeng Wang. 2023.
\newblock \href {http://arxiv.org/abs/2307.11019} {Investigating the factual
  knowledge boundary of large language models with retrieval augmentation}.

\bibitem[{Rezayi et~al.(2023)Rezayi, Liu, Wu, Dhakal, Ge, Dai, Mai, Liu, Zhen,
  Liu, and Li}]{paper137}
Saed Rezayi, Zhengliang Liu, Zihao Wu, Chandra Dhakal, Bao Ge, Haixing Dai,
  Gengchen Mai, Ninghao Liu, Chen Zhen, Tianming Liu, and Sheng Li. 2023.
\newblock \href {http://arxiv.org/abs/2306.11892} {Exploring new frontiers in
  agricultural nlp: Investigating the potential of large language models for
  food applications}.

\bibitem[{Robinson et~al.(2023)Robinson, Ogayo, Mortensen, and
  Neubig}]{paper163}
Nathaniel~R. Robinson, Perez Ogayo, David~R. Mortensen, and Graham Neubig.
  2023.
\newblock \href {http://arxiv.org/abs/2309.07423} {ChatGPT MT: Competitive for
  high- (but not low-) resource languages}.

\bibitem[{Rutinowski et~al.(2023)Rutinowski, Franke, Endendyk, Dormuth, and
  Pauly}]{paper88}
Jérôme Rutinowski, Sven Franke, Jan Endendyk, Ina Dormuth, and Markus Pauly.
  2023.
\newblock \href {http://arxiv.org/abs/2304.07333} {The self-perception and
  political biases of ChatGPT}.

\bibitem[{Sainz et~al.(2023)Sainz, Campos, García-Ferrero, Etxaniz, and
  Agirre}]{paper0}
Oscar Sainz, Jon~Ander Campos, Iker García-Ferrero, Julen Etxaniz, and Eneko
  Agirre. 2023.
\newblock \href {https://hitz-zentroa.github.io/lm-contamination/blog/} {Did
  ChatGPT cheat on your test?}

\bibitem[{Sap et~al.(2023)Sap, LeBras, Fried, and Choi}]{paper181}
Maarten Sap, Ronan LeBras, Daniel Fried, and Yejin Choi. 2023.
\newblock \href {http://arxiv.org/abs/2210.13312} {Neural theory-of-mind? on
  the limits of social intelligence in large lms}.

\bibitem[{Schuster and Markert(2023)}]{paper223}
Jakob Schuster and Katja Markert. 2023.
\newblock \href {https://aclanthology.org/2023.clasp-1.12} {Nut-cracking
  sledgehammers: Prioritizing target language data over bigger language models
  for cross-lingual metaphor detection}.
\newblock In \emph{Proceedings of the 2023 CLASP Conference on Learning with
  Small Data (LSD)}, pages 98--106, Gothenburg, Sweden. Association for
  Computational Linguistics.

\bibitem[{Shakarian et~al.(2023)Shakarian, Koyyalamudi, Ngu, and
  Mareedu}]{paper67}
Paulo Shakarian, Abhinav Koyyalamudi, Noel Ngu, and Lakshmivihari Mareedu.
  2023.
\newblock \href {http://arxiv.org/abs/2302.13814} {An independent evaluation of
  ChatGPT on mathematical word problems (mwp)}.

\bibitem[{Shapira et~al.(2023)Shapira, Zwirn, and Goldberg}]{paper211}
Natalie Shapira, Guy Zwirn, and Yoav Goldberg. 2023.
\newblock \href {https://doi.org/10.18653/v1/2023.findings-acl.663} {How well
  do large language models perform on faux pas tests?}
\newblock In \emph{Findings of the Association for Computational Linguistics:
  ACL 2023}, pages 10438--10451, Toronto, Canada. Association for Computational
  Linguistics.

\bibitem[{Shen et~al.(2023)Shen, Chen, Backes, and Zhang}]{paper151}
Xinyue Shen, Zeyuan Chen, Michael Backes, and Yang Zhang. 2023.
\newblock \href {http://arxiv.org/abs/2304.08979} {In ChatGPT we trust?
  measuring and characterizing the reliability of ChatGPT}.

\bibitem[{Shi et~al.(2023)Shi, Ma, Zhong, Tan, Mai, Li, Liu, and
  Huang}]{paper79}
Yucheng Shi, Hehuan Ma, Wenliang Zhong, Qiaoyu Tan, Gengchen Mai, Xiang Li,
  Tianming Liu, and Junzhou Huang. 2023.
\newblock \href {http://arxiv.org/abs/2305.03513} {Chatgraph: Interpretable
  text classification by converting ChatGPT knowledge to graphs}.

\bibitem[{Shi and Lipani(2023)}]{shi2023dont}
Zhengxaing Shi and Aldo Lipani. 2023.
\newblock \href {https://openreview.net/forum?id=s7xWeJQACI} {Don't stop
  pretraining? make prompt-based fine-tuning powerful learner}.
\newblock In \emph{Thirty-seventh Conference on Neural Information Processing
  Systems}.

\bibitem[{Sobania et~al.(2023)Sobania, Briesch, Hanna, and Petke}]{paper187}
Dominik Sobania, Martin Briesch, Carol Hanna, and Justyna Petke. 2023.
\newblock \href {http://arxiv.org/abs/2301.08653} {An analysis of the automatic
  bug fixing performance of ChatGPT}.

\bibitem[{Song et~al.(2023)Song, Jiang, Shi, Yao, Lu, Feng, Liu, and
  Jing}]{paper177}
Mingyang Song, Haiyun Jiang, Shuming Shi, Songfang Yao, Shilong Lu, Yi~Feng,
  Huafeng Liu, and Liping Jing. 2023.
\newblock \href {http://arxiv.org/abs/2303.13001} {Is ChatGPT a good keyphrase
  generator? a preliminary study}.

\bibitem[{Soni and Wade(2023)}]{paper179}
Mayank Soni and Vincent Wade. 2023.
\newblock \href {http://arxiv.org/abs/2303.17650} {Comparing abstractive
  summaries generated by ChatGPT to real summaries through blinded reviewers
  and text classification algorithms}.

\bibitem[{Stap and Araabi(2023)}]{paper197}
David Stap and Ali Araabi. 2023.
\newblock \href {https://doi.org/10.18653/v1/2023.americasnlp-1.17}
  {{C}hat{GPT} is not a good indigenous translator}.
\newblock In \emph{Proceedings of the Workshop on Natural Language Processing
  for Indigenous Languages of the Americas (AmericasNLP)}, pages 163--167,
  Toronto, Canada. Association for Computational Linguistics.

\bibitem[{Sun et~al.(2023{\natexlab{a}})Sun, Xu, Tang, Wang, Lin, Gong, Ni,
  Shum, and Guo}]{paper132}
Jiashuo Sun, Chengjin Xu, Lumingyuan Tang, Saizhuo Wang, Chen Lin, Yeyun Gong,
  Lionel~M. Ni, Heung-Yeung Shum, and Jian Guo. 2023{\natexlab{a}}.
\newblock \href {http://arxiv.org/abs/2307.07697} {Think-on-graph: Deep and
  responsible reasoning of large language model on knowledge graph}.

\bibitem[{Sun et~al.(2023{\natexlab{b}})Sun, Yan, Ma, Ren, Yin, and
  Ren}]{paper22}
Weiwei Sun, Lingyong Yan, Xinyu Ma, Pengjie Ren, Dawei Yin, and Zhaochun Ren.
  2023{\natexlab{b}}.
\newblock \href {http://arxiv.org/abs/2304.09542} {Is ChatGPT good at search?
  investigating large language models as re-ranking agent}.

\bibitem[{Syriani et~al.(2023)Syriani, David, and Kumar}]{paper101}
Eugene Syriani, Istvan David, and Gauransh Kumar. 2023.
\newblock \href {http://arxiv.org/abs/2307.06464} {Assessing the ability of
  ChatGPT to screen articles for systematic reviews}.

\bibitem[{Tan et~al.(2023{\natexlab{a}})Tan, Ng, and Bing}]{paper227}
Qingyu Tan, Hwee~Tou Ng, and Lidong Bing. 2023{\natexlab{a}}.
\newblock \href {https://doi.org/10.18653/v1/2023.acl-long.828} {Towards
  benchmarking and improving the temporal reasoning capability of large
  language models}.
\newblock In \emph{Proceedings of the 61st Annual Meeting of the Association
  for Computational Linguistics (Volume 1: Long Papers)}, pages 14820--14835,
  Toronto, Canada. Association for Computational Linguistics.

\bibitem[{Tan et~al.(2023{\natexlab{b}})Tan, Min, Li, Li, Hu, Chen, and
  Qi}]{paper68}
Yiming Tan, Dehai Min, Yu~Li, Wenbo Li, Nan Hu, Yongrui Chen, and Guilin Qi.
  2023{\natexlab{b}}.
\newblock \href {http://arxiv.org/abs/2303.07992} {Can ChatGPT replace
  traditional kbqa models? an in-depth analysis of the question answering
  performance of the GPT LLM family}.

\bibitem[{Tan et~al.(2023{\natexlab{c}})Tan, Huang, Jia, Cai, Li, Lu, Zhuang,
  Tu, Xie, Huang, and Jiang}]{paper250}
Zeqi Tan, Shen Huang, Zixia Jia, Jiong Cai, Yinghui Li, Weiming Lu, Yueting
  Zhuang, Kewei Tu, Pengjun Xie, Fei Huang, and Yong Jiang. 2023{\natexlab{c}}.
\newblock \href {https://doi.org/10.18653/v1/2023.semeval-1.277} {{DAMO}-{NLP}
  at {S}em{E}val-2023 task 2: A unified retrieval-augmented system for
  multilingual named entity recognition}.
\newblock In \emph{Proceedings of the 17th International Workshop on Semantic
  Evaluation (SemEval-2023)}, pages 2014--2028, Toronto, Canada. Association
  for Computational Linguistics.

\bibitem[{Tang et~al.(2023{\natexlab{a}})Tang, Goyal, Fabbri, Laban, Xu, Yavuz,
  Kryscinski, Rousseau, and Durrett}]{paper230}
Liyan Tang, Tanya Goyal, Alex Fabbri, Philippe Laban, Jiacheng Xu, Semih Yavuz,
  Wojciech Kryscinski, Justin Rousseau, and Greg Durrett. 2023{\natexlab{a}}.
\newblock \href {https://doi.org/10.18653/v1/2023.acl-long.650} {Understanding
  factual errors in summarization: Errors, summarizers, datasets, error
  detectors}.
\newblock In \emph{Proceedings of the 61st Annual Meeting of the Association
  for Computational Linguistics (Volume 1: Long Papers)}, pages 11626--11644,
  Toronto, Canada. Association for Computational Linguistics.

\bibitem[{Tang et~al.(2023{\natexlab{b}})Tang, Han, Jiang, and Hu}]{paper169}
Ruixiang Tang, Xiaotian Han, Xiaoqian Jiang, and Xia Hu. 2023{\natexlab{b}}.
\newblock \href {http://arxiv.org/abs/2303.04360} {Does synthetic data
  generation of LLMs help clinical text mining?}

\bibitem[{Tu et~al.(2023{\natexlab{a}})Tu, Ma, and Zhang}]{paper60}
Ruibo Tu, Chao Ma, and Cheng Zhang. 2023{\natexlab{a}}.
\newblock \href {http://arxiv.org/abs/2301.13819} {Causal-discovery performance
  of ChatGPT in the context of neuropathic pain diagnosis}.

\bibitem[{Tu et~al.(2023{\natexlab{b}})Tu, Li, Yu, Wang, Hou, and
  Li}]{paper121}
Shangqing Tu, Chunyang Li, Jifan Yu, Xiaozhi Wang, Lei Hou, and Juanzi Li.
  2023{\natexlab{b}}.
\newblock \href {http://arxiv.org/abs/2304.14106} {Chatlog: Recording and
  analyzing ChatGPT across time}.

\bibitem[{Van~Nooten and Daelemans(2023)}]{paper247}
Jens Van~Nooten and Walter Daelemans. 2023.
\newblock \href {https://doi.org/10.18653/v1/2023.wassa-1.23} {Improving
  {D}utch vaccine hesitancy monitoring via multi-label data augmentation with
  {GPT}-3.5}.
\newblock In \emph{Proceedings of the 13th Workshop on Computational Approaches
  to Subjectivity, Sentiment, {\&} Social Media Analysis}, pages 251--270,
  Toronto, Canada. Association for Computational Linguistics.

\bibitem[{Vedula et~al.(2023)Vedula, Kodali, Shrivastava, and
  Kumaraguru}]{paper234}
Bhaskara~Hanuma Vedula, Prashant Kodali, Manish Shrivastava, and Ponnurangam
  Kumaraguru. 2023.
\newblock \href {https://doi.org/10.18653/v1/2023.wassa-1.58}
  {{P}recog{IIITH}@{WASSA}2023: Emotion detection for {U}rdu-{E}nglish
  code-mixed text}.
\newblock In \emph{Proceedings of the 13th Workshop on Computational Approaches
  to Subjectivity, Sentiment, {\&} Social Media Analysis}, pages 601--605,
  Toronto, Canada. Association for Computational Linguistics.

\bibitem[{Vemprala et~al.(2023)Vemprala, Bonatti, Bucker, and
  Kapoor}]{paper139}
Sai Vemprala, Rogerio Bonatti, Arthur Bucker, and Ashish Kapoor. 2023.
\newblock \href {http://arxiv.org/abs/2306.17582} {ChatGPT for robotics: Design
  principles and model abilities}.

\bibitem[{Wang et~al.(2023{\natexlab{a}})Wang, Yue, and Sun}]{paper38}
Boshi Wang, Xiang Yue, and Huan Sun. 2023{\natexlab{a}}.
\newblock \href {http://arxiv.org/abs/2305.13160} {Can ChatGPT defend its
  belief in truth? evaluating LLM reasoning via debate}.

\bibitem[{Wang et~al.(2023{\natexlab{b}})Wang, Chen, Pei, Xie, Kang, Zhang, Xu,
  Xiong, Dutta, Schaeffer, Truong, Arora, Mazeika, Hendrycks, Lin, Cheng,
  Koyejo, Song, and Li}]{paper91}
Boxin Wang, Weixin Chen, Hengzhi Pei, Chulin Xie, Mintong Kang, Chenhui Zhang,
  Chejian Xu, Zidi Xiong, Ritik Dutta, Rylan Schaeffer, Sang~T. Truong, Simran
  Arora, Mantas Mazeika, Dan Hendrycks, Zinan Lin, Yu~Cheng, Sanmi Koyejo, Dawn
  Song, and Bo~Li. 2023{\natexlab{b}}.
\newblock \href {http://arxiv.org/abs/2306.11698} {Decodingtrust: A
  comprehensive assessment of trustworthiness in GPT models}.

\bibitem[{Wang et~al.(2023{\natexlab{c}})Wang, Liang, Meng, Sun, Shi, Li, Xu,
  Qu, and Zhou}]{paper23}
Jiaan Wang, Yunlong Liang, Fandong Meng, Zengkui Sun, Haoxiang Shi, Zhixu Li,
  Jinan Xu, Jianfeng Qu, and Jie Zhou. 2023{\natexlab{c}}.
\newblock \href {http://arxiv.org/abs/2303.04048} {Is ChatGPT a good NLG
  evaluator? a preliminary study}.

\bibitem[{Wang et~al.(2023{\natexlab{d}})Wang, Liang, Meng, Zou, Li, Qu, and
  Zhou}]{paper30}
Jiaan Wang, Yunlong Liang, Fandong Meng, Beiqi Zou, Zhixu Li, Jianfeng Qu, and
  Jie Zhou. 2023{\natexlab{d}}.
\newblock \href {http://arxiv.org/abs/2302.14229} {Zero-shot cross-lingual
  summarization via large language models}.

\bibitem[{Wang et~al.(2023{\natexlab{e}})Wang, Hu, Hou, Chen, Zheng, Wang,
  Yang, Huang, Ye, Geng, Jiao, Zhang, and Xie}]{paper66}
Jindong Wang, Xixu Hu, Wenxin Hou, Hao Chen, Runkai Zheng, Yidong Wang, Linyi
  Yang, Haojun Huang, Wei Ye, Xiubo Geng, Binxin Jiao, Yue Zhang, and Xing Xie.
  2023{\natexlab{e}}.
\newblock \href {http://arxiv.org/abs/2302.12095} {On the robustness of
  ChatGPT: An adversarial and out-of-distribution perspective}.

\bibitem[{Wang et~al.(2023{\natexlab{f}})Wang, Yao, Mitra, Osebe, Yang, and
  Yu}]{paper251}
Junda Wang, Zonghai Yao, Avijit Mitra, Samuel Osebe, Zhichao Yang, and Hong Yu.
  2023{\natexlab{f}}.
\newblock \href {https://doi.org/10.18653/v1/2023.clinicalnlp-1.49}
  {{UMASS}{\_}{B}io{NLP} at {MEDIQA}-chat 2023: Can {LLM}s generate
  high-quality synthetic note-oriented doctor-patient conversations?}
\newblock In \emph{Proceedings of the 5th Clinical Natural Language Processing
  Workshop}, pages 460--471, Toronto, Canada. Association for Computational
  Linguistics.

\bibitem[{Wang and Demszky(2023)}]{paper208}
Rose Wang and Dorottya Demszky. 2023.
\newblock \href {https://doi.org/10.18653/v1/2023.bea-1.53} {Is {C}hat{GPT} a
  good teacher coach? measuring zero-shot performance for scoring and providing
  actionable insights on classroom instruction}.
\newblock In \emph{Proceedings of the 18th Workshop on Innovative Use of NLP
  for Building Educational Applications (BEA 2023)}, pages 626--667, Toronto,
  Canada. Association for Computational Linguistics.

\bibitem[{Wang et~al.(2023{\natexlab{g}})Wang, Zhao, Ouyang, Wang, and
  Shen}]{paper190}
Sheng Wang, Zihao Zhao, Xi~Ouyang, Qian Wang, and Dinggang Shen.
  2023{\natexlab{g}}.
\newblock \href {http://arxiv.org/abs/2302.07257} {ChatCAD: Interactive
  computer-aided diagnosis on medical image using large language models}.

\bibitem[{Wang et~al.(2023{\natexlab{h}})Wang, Hu, Lu, Zhu, Zhang, Subramaniam,
  Loomba, Zhang, Sun, and Wang}]{paper131}
Xiaoxuan Wang, Ziniu Hu, Pan Lu, Yanqiao Zhu, Jieyu Zhang, Satyen Subramaniam,
  Arjun~R. Loomba, Shichang Zhang, Yizhou Sun, and Wei Wang.
  2023{\natexlab{h}}.
\newblock \href {http://arxiv.org/abs/2307.10635} {Scibench: Evaluating
  college-level scientific problem-solving abilities of large language models}.

\bibitem[{Wang et~al.(2023{\natexlab{i}})Wang, Wang, Liu, Chen, Yuan, Peng, and
  Ji}]{paper138}
Xingyao Wang, Zihan Wang, Jiateng Liu, Yangyi Chen, Lifan Yuan, Hao Peng, and
  Heng Ji. 2023{\natexlab{i}}.
\newblock \href {http://arxiv.org/abs/2309.10691} {Mint: Evaluating LLMs in
  multi-turn interaction with tools and language feedback}.

\bibitem[{Wang and Zhao(2023)}]{paper84}
Yuqing Wang and Yun Zhao. 2023.
\newblock \href {http://arxiv.org/abs/2308.05342} {Metacognitive prompting
  improves understanding in large language models}.

\bibitem[{Wang et~al.(2023{\natexlab{j}})Wang, Xie, Ding, Feng, and
  Xia}]{paper75}
Zengzhi Wang, Qiming Xie, Zixiang Ding, Yi~Feng, and Rui Xia.
  2023{\natexlab{j}}.
\newblock \href {http://arxiv.org/abs/2304.04339} {Is ChatGPT a good sentiment
  analyzer? a preliminary study}.

\bibitem[{Wei et~al.(2023)Wei, Cui, Cheng, Wang, Zhang, Huang, Xie, Xu, Chen,
  Zhang, Jiang, and Han}]{paper21}
Xiang Wei, Xingyu Cui, Ning Cheng, Xiaobin Wang, Xin Zhang, Shen Huang, Pengjun
  Xie, Jinan Xu, Yufeng Chen, Meishan Zhang, Yong Jiang, and Wenjuan Han. 2023.
\newblock \href {http://arxiv.org/abs/2302.10205} {Zero-shot information
  extraction via chatting with ChatGPT}.

\bibitem[{White et~al.(2023)White, Hays, Fu, Spencer-Smith, and
  Schmidt}]{paper171}
Jules White, Sam Hays, Quchen Fu, Jesse Spencer-Smith, and Douglas~C. Schmidt.
  2023.
\newblock \href {http://arxiv.org/abs/2303.07839} {ChatGPT prompt patterns for
  improving code quality, refactoring, requirements elicitation, and software
  design}.

\bibitem[{Wu et~al.(2023{\natexlab{a}})Wu, Wang, Wan, Jiao, and Lyu}]{paper18}
Haoran Wu, Wenxuan Wang, Yuxuan Wan, Wenxiang Jiao, and Michael Lyu.
  2023{\natexlab{a}}.
\newblock \href {http://arxiv.org/abs/2303.13648} {ChatGPT or grammarly?
  evaluating ChatGPT on grammatical error correction benchmark}.

\bibitem[{Wu et~al.(2023{\natexlab{b}})Wu, Jiang, Yin, Karlsson, and
  Lin}]{paper216}
Qianhui Wu, Huiqiang Jiang, Haonan Yin, B{\"o}rje Karlsson, and Chin-Yew Lin.
  2023{\natexlab{b}}.
\newblock \href {https://doi.org/10.18653/v1/2023.acl-long.403} {Multi-level
  knowledge distillation for out-of-distribution detection in text}.
\newblock In \emph{Proceedings of the 61st Annual Meeting of the Association
  for Computational Linguistics (Volume 1: Long Papers)}, pages 7317--7332,
  Toronto, Canada. Association for Computational Linguistics.

\bibitem[{Wu et~al.(2023{\natexlab{c}})Wu, Jiang, Jiang, Xie, and
  Tu}]{paper235}
Weiqi Wu, Chengyue Jiang, Yong Jiang, Pengjun Xie, and Kewei Tu.
  2023{\natexlab{c}}.
\newblock \href {https://doi.org/10.18653/v1/2023.acl-long.173} {Do {PLM}s know
  and understand ontological knowledge?}
\newblock In \emph{Proceedings of the 61st Annual Meeting of the Association
  for Computational Linguistics (Volume 1: Long Papers)}, pages 3080--3101,
  Toronto, Canada. Association for Computational Linguistics.

\bibitem[{Xiao et~al.(2023)Xiao, Xu, Zhang, Wang, and Xia}]{paper204}
Changrong Xiao, Sean~Xin Xu, Kunpeng Zhang, Yufang Wang, and Lei Xia. 2023.
\newblock \href {https://doi.org/10.18653/v1/2023.bea-1.52} {Evaluating reading
  comprehension exercises generated by {LLM}s: A showcase of {C}hat{GPT} in
  education applications}.
\newblock In \emph{Proceedings of the 18th Workshop on Innovative Use of NLP
  for Building Educational Applications (BEA 2023)}, pages 610--625, Toronto,
  Canada. Association for Computational Linguistics.

\bibitem[{Xie et~al.(2023{\natexlab{a}})Xie, Zhang, Chen, Lou, and
  Su}]{paper124}
Jian Xie, Kai Zhang, Jiangjie Chen, Renze Lou, and Yu~Su. 2023{\natexlab{a}}.
\newblock \href {http://arxiv.org/abs/2305.13300} {Adaptive chameleon or
  stubborn sloth: Revealing the behavior of large language models in knowledge
  conflicts}.

\bibitem[{Xie et~al.(2023{\natexlab{b}})Xie, Han, Lai, Peng, and
  Huang}]{paper50}
Qianqian Xie, Weiguang Han, Yanzhao Lai, Min Peng, and Jimin Huang.
  2023{\natexlab{b}}.
\newblock \href {http://arxiv.org/abs/2304.05351} {The wall street neophyte: A
  zero-shot analysis of ChatGPT over multimodal stock movement prediction
  challenges}.

\bibitem[{Xu et~al.(2023{\natexlab{a}})Xu, Lin, Han, Zhao, Liu, and
  Cambria}]{paper129}
Fangzhi Xu, Qika Lin, Jiawei Han, Tianzhe Zhao, Jun Liu, and Erik Cambria.
  2023{\natexlab{a}}.
\newblock \href {http://arxiv.org/abs/2306.09841} {Are large language models
  really good logical reasoners? a comprehensive evaluation and beyond}.

\bibitem[{Xu et~al.(2023{\natexlab{b}})Xu, Li, Zhu, Xue, Zhu, Zhao, He, Zhang,
  Kang, and Lan}]{paper27}
Liang Xu, Anqi Li, Lei Zhu, Hang Xue, Changtai Zhu, Kangkang Zhao, Haonan He,
  Xuanwei Zhang, Qiyue Kang, and Zhenzhong Lan. 2023{\natexlab{b}}.
\newblock \href {http://arxiv.org/abs/2307.15020} {SuperCLUE: A comprehensive
  chinese large language model benchmark}.

\bibitem[{Xu et~al.(2023{\natexlab{c}})Xu, Yang, Cui, and Wang}]{paper233}
Zihang Xu, Ziqing Yang, Yiming Cui, and Shijin Wang. 2023{\natexlab{c}}.
\newblock \href {https://doi.org/10.18653/v1/2023.findings-acl.513} {{IDOL}:
  Indicator-oriented logic pre-training for logical reasoning}.
\newblock In \emph{Findings of the Association for Computational Linguistics:
  ACL 2023}, pages 8099--8111, Toronto, Canada. Association for Computational
  Linguistics.

\bibitem[{Yang et~al.(2023{\natexlab{a}})Yang, Jin, Tang, Han, Feng, Jiang,
  Yin, and Hu}]{paper62}
Jingfeng Yang, Hongye Jin, Ruixiang Tang, Xiaotian Han, Qizhang Feng, Haoming
  Jiang, Bing Yin, and Xia Hu. 2023{\natexlab{a}}.
\newblock \href {http://arxiv.org/abs/2304.13712} {Harnessing the power of LLMs
  in practice: A survey on ChatGPT and beyond}.

\bibitem[{Yang and Nicolai(2023)}]{paper162}
Wayne Yang and Garrett Nicolai. 2023.
\newblock \href {http://arxiv.org/abs/2307.05779} {Neural machine translation
  data generation and augmentation using ChatGPT}.

\bibitem[{Yang et~al.(2023{\natexlab{b}})Yang, Li, Zhang, Chen, and
  Cheng}]{paper93}
Xianjun Yang, Yan Li, Xinlu Zhang, Haifeng Chen, and Wei Cheng.
  2023{\natexlab{b}}.
\newblock \href {http://arxiv.org/abs/2302.08081} {Exploring the limits of
  ChatGPT for query or aspect-based text summarization}.

\bibitem[{Yang et~al.(2023{\natexlab{c}})Yang, Li, Wang, Lin, Azarnasab, Ahmed,
  Liu, Liu, Zeng, and Wang}]{paper176}
Zhengyuan Yang, Linjie Li, Jianfeng Wang, Kevin Lin, Ehsan Azarnasab, Faisal
  Ahmed, Zicheng Liu, Ce~Liu, Michael Zeng, and Lijuan Wang.
  2023{\natexlab{c}}.
\newblock \href {http://arxiv.org/abs/2303.11381} {Mm-react: Prompting ChatGPT
  for multimodal reasoning and action}.

\bibitem[{Yao et~al.(2023)Yao, Yu, Zhao, Shafran, Griffiths, Cao, and
  Narasimhan}]{paper120}
Shunyu Yao, Dian Yu, Jeffrey Zhao, Izhak Shafran, Thomas~L. Griffiths, Yuan
  Cao, and Karthik Narasimhan. 2023.
\newblock \href {http://arxiv.org/abs/2305.10601} {Tree of thoughts: Deliberate
  problem solving with large language models}.

\bibitem[{Ye et~al.(2023)Ye, Chen, Xu, Zu, Shao, Liu, Cui, Zhou, Gong, Shen,
  Zhou, Chen, Gui, Zhang, and Huang}]{paper73}
Junjie Ye, Xuanting Chen, Nuo Xu, Can Zu, Zekai Shao, Shichun Liu, Yuhan Cui,
  Zeyang Zhou, Chao Gong, Yang Shen, Jie Zhou, Siming Chen, Tao Gui, Qi~Zhang,
  and Xuanjing Huang. 2023.
\newblock \href {http://arxiv.org/abs/2303.10420} {A comprehensive capability
  analysis of GPT-3 and GPT-3.5 series models}.

\bibitem[{Yuan et~al.(2023{\natexlab{a}})Yuan, Xie, and Ananiadou}]{paper77}
Chenhan Yuan, Qianqian Xie, and Sophia Ananiadou. 2023{\natexlab{a}}.
\newblock \href {http://arxiv.org/abs/2304.05454} {Zero-shot temporal relation
  extraction with ChatGPT}.

\bibitem[{Yuan et~al.(2023{\natexlab{b}})Yuan, Yuan, Tan, Wang, and
  Huang}]{paper112}
Zheng Yuan, Hongyi Yuan, Chuanqi Tan, Wei Wang, and Songfang Huang.
  2023{\natexlab{b}}.
\newblock \href {http://arxiv.org/abs/2304.02015} {How well do large language
  models perform in arithmetic tasks?}

\bibitem[{Zhang et~al.(2022)Zhang, Ding, and Jing}]{zhang2022would}
Bowen Zhang, Daijun Ding, and Liwen Jing. 2022.
\newblock \href{https://arxiv.org/abs/2212.14548}{How would stance detection techniques evolve after the launch of
  ChatGPT?}
\newblock \emph{arXiv preprint arXiv:2212.14548}.

\bibitem[{Zhang et~al.(2023{\natexlab{a}})Zhang, Fu, Ding, Huang, Li, and
  Jing}]{paper87}
Bowen Zhang, Xianghua Fu, Daijun Ding, Hu~Huang, Yangyang Li, and Liwen Jing.
  2023{\natexlab{a}}.
\newblock \href {http://arxiv.org/abs/2304.03087} {Investigating
  chain-of-thought with ChatGPT for stance detection on social media}.

\bibitem[{Zhang et~al.(2023{\natexlab{b}})Zhang, Liu, and Zhang}]{paper17}
Haopeng Zhang, Xiao Liu, and Jiawei Zhang. 2023{\natexlab{b}}.
\newblock \href {http://arxiv.org/abs/2304.04193} {Extractive summarization via
  ChatGPT for faithful summary generation}.

\bibitem[{Zhang et~al.(2023{\natexlab{c}})Zhang, Qian, Liu, Heinecke, Meng,
  Liu, Yu, Wang, Savarese, and Xiong}]{paper254}
Jianguo Zhang, Kun Qian, Zhiwei Liu, Shelby Heinecke, Rui Meng, Ye~Liu, Zhou
  Yu, Huan Wang, Silvio Savarese, and Caiming Xiong. 2023{\natexlab{c}}.
\newblock \href {http://arxiv.org/abs/2307.10172} {Dialogstudio: Towards
  richest and most diverse unified dataset collection for conversational AI}.

\bibitem[{Zhang et~al.(2023{\natexlab{d}})Zhang, Fang, Zhang, Ma, Zhou, Huang,
  Bu, Gui, Chen, Chen, and Feng}]{paper161}
Shaolei Zhang, Qingkai Fang, Zhuocheng Zhang, Zhengrui Ma, Yan Zhou, Langlin
  Huang, Mengyu Bu, Shangtong Gui, Yunji Chen, Xilin Chen, and Yang Feng.
  2023{\natexlab{d}}.
\newblock \href {http://arxiv.org/abs/2306.10968} {BayLing: Bridging
  cross-lingual alignment and instruction following through interactive
  translation for large language models}.

\bibitem[{Zhang et~al.(2023{\natexlab{e}})Zhang, Deng, Liu, Pan, and
  Bing}]{paper82}
Wenxuan Zhang, Yue Deng, Bing Liu, Sinno~Jialin Pan, and Lidong Bing.
  2023{\natexlab{e}}.
\newblock \href {http://arxiv.org/abs/2305.15005} {Sentiment analysis in the
  era of large language models: A reality check}.

\bibitem[{Zhang et~al.(2023{\natexlab{f}})Zhang, Chowdhury, Hong, Gupta, and
  Shang}]{paper186}
Xiyuan Zhang, Ranak~Roy Chowdhury, Dezhi Hong, Rajesh~K. Gupta, and Jingbo
  Shang. 2023{\natexlab{f}}.
\newblock \href {http://arxiv.org/abs/2301.03462} {Modeling label semantics
  improves activity recognition}.

\bibitem[{Zhao et~al.(2023{\natexlab{a}})Zhao, Zhao, Lu, Wang, Tong, and
  Qin}]{paper147}
Weixiang Zhao, Yanyan Zhao, Xin Lu, Shilong Wang, Yanpeng Tong, and Bing Qin.
  2023{\natexlab{a}}.
\newblock \href {http://arxiv.org/abs/2304.09582} {Is ChatGPT equipped with
  emotional dialogue capabilities?}

\bibitem[{Zhao et~al.(2023{\natexlab{b}})Zhao, Zhang, Ma, Su, Liu, Wang, Qiao,
  Guo, Li, and Ma}]{paper200}
Xiaofeng Zhao, Min Zhang, Miaomiao Ma, Chang Su, Yilun Liu, Minghan Wang,
  Xiaosong Qiao, Jiaxin Guo, Yinglu Li, and Wenbing Ma. 2023{\natexlab{b}}.
\newblock \href {https://doi.org/10.18653/v1/2023.semeval-1.221} {{HW}-{TSC} at
  {S}em{E}val-2023 task 7: Exploring the natural language inference
  capabilities of {C}hat{GPT} and pre-trained language model for clinical
  trial}.
\newblock In \emph{Proceedings of the 17th International Workshop on Semantic
  Evaluation (SemEval-2023)}, pages 1603--1608, Toronto, Canada. Association
  for Computational Linguistics.

\bibitem[{Zhao et~al.(2023{\natexlab{c}})Zhao, Ouyang, Yu, Wu, and
  Li}]{paper239}
Xuandong Zhao, Siqi Ouyang, Zhiguo Yu, Ming Wu, and Lei Li. 2023{\natexlab{c}}.
\newblock \href {https://doi.org/10.18653/v1/2023.acl-long.869} {Pre-trained
  language models can be fully zero-shot learners}.
\newblock In \emph{Proceedings of the 61st Annual Meeting of the Association
  for Computational Linguistics (Volume 1: Long Papers)}, pages 15590--15606,
  Toronto, Canada. Association for Computational Linguistics.

\bibitem[{Zheng et~al.(2023{\natexlab{a}})Zheng, Zhou, Meng, Zhou, and
  Huang}]{paper92}
Chujie Zheng, Hao Zhou, Fandong Meng, Jie Zhou, and Minlie Huang.
  2023{\natexlab{a}}.
\newblock \href {http://arxiv.org/abs/2309.03882} {Large language models are
  not robust multiple choice selectors}.

\bibitem[{Zheng et~al.(2023{\natexlab{b}})Zheng, Huang, and Chang}]{paper128}
Shen Zheng, Jie Huang, and Kevin Chen-Chuan Chang. 2023{\natexlab{b}}.
\newblock \href {http://arxiv.org/abs/2304.10513} {Why does ChatGPT fall short
  in providing truthful answers?}

\bibitem[{Zheng et~al.(2023{\natexlab{c}})Zheng, Ross, and Magdy}]{paper249}
Yi~Zheng, Bj{\"o}rn Ross, and Walid Magdy. 2023{\natexlab{c}}.
\newblock \href {https://aclanthology.org/2023.cs4oa-1.5} {What makes good
  counterspeech? a comparison of generation approaches and evaluation metrics}.
\newblock In \emph{Proceedings of the 1st Workshop on CounterSpeech for Online
  Abuse (CS4OA)}, pages 62--71, Prague, Czechia. Association for Computational
  Linguistics.

\bibitem[{Zheng et~al.(2023{\natexlab{d}})Zheng, Qiu, Hu, Wu, Zhu, and
  Xiong}]{paper100}
Zhi Zheng, Zhaopeng Qiu, Xiao Hu, Likang Wu, Hengshu Zhu, and Hui Xiong.
  2023{\natexlab{d}}.
\newblock \href {http://arxiv.org/abs/2307.02157} {Generative job
  recommendations with large language model}.

\bibitem[{Zhong et~al.(2023{\natexlab{a}})Zhong, Ding, Liu, Du, and
  Tao}]{paper65}
Qihuang Zhong, Liang Ding, Juhua Liu, Bo~Du, and Dacheng Tao.
  2023{\natexlab{a}}.
\newblock \href {http://arxiv.org/abs/2302.10198} {Can ChatGPT understand too?
  a comparative study on ChatGPT and fine-tuned bert}.

\bibitem[{Zhong et~al.(2023{\natexlab{b}})Zhong, Wei, Yang, Wu, Liu, Wei, Li,
  Yao, Ma, Li, Zhu, Jiang, Han, Shen, Liu, and Zhang}]{paper116}
Tianyang Zhong, Yaonai Wei, Li~Yang, Zihao Wu, Zhengliang Liu, Xiaozheng Wei,
  Wenjun Li, Junjie Yao, Chong Ma, Xiang Li, Dajiang Zhu, Xi~Jiang, Junwei Han,
  Dinggang Shen, Tianming Liu, and Tuo Zhang. 2023{\natexlab{b}}.
\newblock \href {http://arxiv.org/abs/2304.11107} {Chatabl: Abductive learning
  via natural language interaction with ChatGPT}.

\bibitem[{Zhou et~al.(2023)Zhou, Jiang, Cui, Wang, Xiao, Hou, Cotterell, and
  Sachan}]{paper97}
Wangchunshu Zhou, Yuchen~Eleanor Jiang, Peng Cui, Tiannan Wang, Zhenxin Xiao,
  Yifan Hou, Ryan Cotterell, and Mrinmaya Sachan. 2023.
\newblock \href {http://arxiv.org/abs/2305.13304} {RecurrentGPT: Interactive
  generation of (arbitrarily) long text}.

\bibitem[{Zhu et~al.(2023)Zhu, Zhang, Haq, Hui, and Tyson}]{paper78}
Yiming Zhu, Peixian Zhang, Ehsan-Ul Haq, Pan Hui, and Gareth Tyson. 2023.
\newblock \href {http://arxiv.org/abs/2304.10145} {Can ChatGPT reproduce
  human-generated labels? a study of social computing tasks}.

\bibitem[{Zhuo et~al.(2023)Zhuo, Huang, Chen, and Xing}]{paper85}
Terry~Yue Zhuo, Yujin Huang, Chunyang Chen, and Zhenchang Xing. 2023.
\newblock \href {http://arxiv.org/abs/2301.12867} {Red teaming ChatGPT via
  jailbreaking: Bias, robustness, reliability and toxicity}.

\end{thebibliography}

\appendix

\section{Full list of the reviewed work}
In this section, we list all the work that we reviewed and classified as relevant.

\let\Nsection\section
\renewcommand{\section}[2]{}%

\let\section\Nsection

\section{Detail on evaluation malpractices}
\label{sec:appendix-malpractices}

As the Sankey diagrams showed in \Cref{ssec:reproducibility} and \Cref{ssec:bad-eval} offer limited insights on our findings regarding evaluation reproducibility and fairness, we do provide additional details in this section. We provide concrete numbers for our assessment of reproducibility (Sec.~\ref{ssec:reproducibility}) and evaluation (mal)practices (Sec.~\ref{ssec:bad-eval}) in Tables~\ref{tab:repro_stats} and \ref{tab:fairness_stats}, respectively. 

\begin{table*}[t]
  \centering
  \begin{subtable}{0.5\linewidth} 
    \centering
    \footnotesize
    \begin{tabular}{cccccl}
        \toprule
         Prompts & Repo & Sampl. & Custom & n. (\%) \\
         \midrule
          &  &  &  & 3 (2.11\%) \\
          \midrule
          &  &  & \Checkmark & 1 (0.70\%) \\
          \midrule
          &  & \Checkmark &  & 8 (5.63\%) \\
          \midrule
          & \Checkmark &  &  & 3 (2.11\%) \\
          \midrule
          & \Checkmark & \Checkmark &  & 2 (1.41\%) \\
          \midrule
          \Checkmark &  &  &  & 20 (14.08\%) \\
         \midrule
         \Checkmark &  &  & \Checkmark & 3 (2.11\%) \\
         \midrule
         \Checkmark &  & \Checkmark &  & 27 (19.01\%) \\
         \midrule
         \Checkmark &  & \Checkmark & \Checkmark & 3 (2.11\%) \\
         \midrule
         \Checkmark & \Checkmark &  &  & 37 (26.06\%) \\
         \midrule
         \Checkmark & \Checkmark &  & \Checkmark & 4 (2.82\%) \\
         \midrule
         \Checkmark & \Checkmark & \Checkmark &  & 27 (19.01\%) \\
         \midrule
         \Checkmark & \Checkmark & \Checkmark & \Checkmark & 4 (2.82\%) \\
         \bottomrule
    \end{tabular}
    \subcaption{Pre-prints}
  \end{subtable}%
  \begin{subtable}{0.5\linewidth} 
    \centering
    \footnotesize
    \begin{tabular}{ccccl}
        \toprule
         Prompts & Repo & Sampl. & Custom & n. (\%) \\
        \midrule
          &  &  &  & 1 (1.43\%) \\
          \midrule
          & \Checkmark &  & \Checkmark & 1 (1.43\%) \\
          \midrule
          & \Checkmark & \Checkmark &  & 1 (1.43\%) \\
          \midrule
         \Checkmark &  &  &  & 14 (20.00\%) \\
         \midrule
         \Checkmark &  &  & \Checkmark & 7 (10.00\%) \\
         \midrule
         \Checkmark &  & \Checkmark &  & 9 (12.86\%) \\
         \midrule
         \Checkmark &  & \Checkmark & \Checkmark & 3 (4.29\%) \\
         \midrule
         \Checkmark & \Checkmark &  &  & 8 (11.43\%) \\
         \midrule
         \Checkmark & \Checkmark &  & \Checkmark & 4 (5.71\%) \\
         \midrule
         \Checkmark & \Checkmark & \Checkmark &  & 16 (22.86\%) \\
         \midrule
         \Checkmark & \Checkmark & \Checkmark & \Checkmark & 6 (6.57\%) \\
        \bottomrule \\
        & & & \\
    \end{tabular}
    \subcaption{Peer-reviewed works}
  \end{subtable}
  \caption{Statistics related to the reproducibility of the work reviewed: the availability of used prompts (Prompts) and code/data repository (Repo), the usage of custom datasets (Custom), the application of random sampling or any other practice that does not allow the exact reconstruction of the data used (Sampl.).}
  \label{tab:repro_stats}
\end{table*}

\begin{table*}[!ht]
  \centering
  \begin{subtable}{0.5\linewidth} 
    \vspace{-15mm}
    \centering
    \footnotesize
    \begin{tabular}{cccl}
    \toprule
    Comp. & Scale & n. (\%) \\
    \midrule
      &  & 71 (50.00\%) \\
    \midrule
    \Checkmark &  & 54 (38.03\%) \\
    \midrule
    \Checkmark & \Checkmark & 17 (11.97\%) \\
    \bottomrule
    \end{tabular}
    \subcaption{Pre-prints}
  \end{subtable}%
  \begin{subtable}{0.5\linewidth} 
    \centering
    \footnotesize
    \begin{tabular}{cccl}
    \toprule
    Comp. & Scale & n. (\%) \\
    \midrule
      &  & 30 (42.86\%) \\
    \midrule
     \Checkmark &  & 34 (48.57\%) \\
    \midrule
     \Checkmark & \Checkmark & 6 (8.57\%) \\
    \bottomrule\\
    \end{tabular}
    \subcaption{Peer-reviewed works}
  \end{subtable}
  \caption{Fairness statistics for reviewed work. Statistics related to the practices of performance comparisons between ChatGPT/GPT-4 and other open models: whether such comparisons are performed at all (Comp.) and whether they are of the same scale (Scale).}
  \label{tab:fairness_stats}
\end{table*}

\section{Detailed List of ChatGPT Data Leak}
\label{sec:appendix-leak}

We show which datasets have been leaked to ChatGPT in Tables~\ref{tab:data-leak_1} and \ref{tab:data-leak_2}.

\begin{sidewaystable*}
\centering
\small
\begin{tabular}{p{3cm}p{4.8cm}p{4.8cm}p{4.8cm}p{4.8cm}}
Task name&Lo&M-Lo&M-Hi&Hi\\\toprule

AI safety \& ethics&&&bbq (all), bold (all)&
\\ \midrule \\
Creative text generation& \href{https://www.kaggle.com/datasets/ratthachat/writing-prompts}{WrintingPrompts} (test)&&&
\\  \midrule \\
Dialogue&\href{https://github.com/facebookresearch/opendialkg}{OpenDialKG} (test), \href{https://huggingface.co/datasets/allenai/prosocial-dialog}{ProsocialDialog} (test)&\href{https://huggingface.co/datasets/multi_woz_v22}{MultiWOZ 2.2} (test)&&\href{https://github.com/alexa/dstc11-track5}{DSTC11 track 5} (dev), \href{https://paperswithcode.com/dataset/dstc7-task-2}{DSTC7 Track 2} (all), \href{https://huggingface.co/datasets/ConvLab/multiwoz21}{MultiWOZ 2.1} (test), \href{https://github.com/smartyfh/MultiWOZ2.4}{MultiWOZ 2.4} (test), mutual (test)
\\ \midrule \\
Evaluation of generated texts&&&&\href{https://huggingface.co/datasets/newsroom}{NEWSROOM} (all), \href{https://github.com/thu-coai/OpenMEVA}{OpenMEVA} (all), \href{https://github.com/neulab/REALSumm}{RealSumm} (all), \href{https://github.com/Yale-LILY/SummEval}{SummEval} (all)
\\ \midrule \\
Machine Translation&\href{https://github.com/facebookresearch/flores/tree/main/previous_releases/flores101}{FLORES-101} (test), \href{https://paperswithcode.com/dataset/wmt-2020}{WMT20 (EN-DE; Robustness Task Set 2 - EN-JA; Robustness Task Set 2 - JA-EN; Robustness Task 3; ZH-EN)} (test), \href{https://www.statmt.org/wmt22/translation-task.html}{WMT22} (test)&\href{https://github.com/IndoNLP/nusax}{NusaX} (test), \href{https://github.com/biomedical-translation-corpora/corpora}{WMT19 Biomedical Translation Task} (test), \href{https://paperswithcode.com/dataset/wmt-2014}{WMT 2014 News dataset (EN-FR; EN-DE)} (test)&\href{https://github.com/facebookresearch/flores/tree/main/flores200}{FLORES-200} (dev)& \href{https://github.com/google/wmt-mqm-human-evaluation/tree/main/generalMT2022}{MQM annotations of the WMT 2022 task (EN-DE, EN-RU, ZH-EN)} (test)
\\ \midrule \\
Math&&\href{https://inklab.usc.edu/NumerSense/}{NumerSense} (dev)&&\href{https://www.cs.washington.edu/nlp/arithmetic}{AddSub} (all), \href{https://huggingface.co/datasets/aqua_rat}{AQUA-RAT} (test), \href{https://www.microsoft.com/en-us/download/details.aspx?id=52628}{DRAW-1K} (all), \href{https://github.com/friederrr/GHOSTS}{GHOSTS} (all), \href{https://github.com/openai/grade-school-math}{GSM8K} (test), \href{https://huggingface.co/datasets/ChilleD/MultiArith}{MultiArith} (all), \href{https://gitlab.cs.washington.edu/ALGES/TACL2015/-/blob/master/questions.json?ref_type=heads}{SingleEQ} (all), \href{https://github.com/arkilpatel/SVAMP}{SVAMP} (all)
\\ \hline \\
Medical text generation&\href{https://github.com/bruzwen/ddxplus}{DDXPlus (EN)} (test), \href{https://physionet.org/content/mimic-cxr/2.0.0/}{MIMIC-CXR} (test)&&&\href{https://www.merckmanuals.com/professional/pages-with-widgets/case-studies?mode=list}{Merck Sharpe \& Dohme (MSD) clinical manual} (all)
\\ \hline \\
Natural Language Inference& \href{https://github.com/MJ-Jang/BECEL/tree/main}{BECEL (SNLI; RTE)} (test), \href{https://github.com/mcdm/CommitmentBank}{CommitmentBank} (all), \href{https://huggingface.co/datasets/multi_nli}{MultiNLI} (dev), \href{https://paperswithcode.com/dataset/qnli}{QNLI} (dev), \href{https://paperswithcode.com/dataset/rte}{RTE} (all), \href{https://leaderboard.allenai.org/anli/submissions/get-started}{$\alpha$nli} (dev)& \href{https://allenai.org/data/entailmentbank}{EntailmentBank} (test)&& \href{https://github.com/verypluming/MED}{MED} (test), \href{https://github.com/AI-secure/adversarial-glue/tree/main}{Adversarial GLUE (MNLI; QNLI; RTE)} (dev), \href{https://github.com/facebookresearch/anli?tab=readme-ov-file}{ANLI-R3} (test), \href{https://super.gluebenchmark.com/}{SuperGLUE (AX-g; cb)} (dev), \href{https://github.com/swarnaHub/ConjNLI}{ConjNLI} (test), \href{https://github.com/csitfun/ConTRoL-dataset}{ConTRoL (logical reasoning)} (test), \href{https://github.com/verypluming/HELP}{HELP} (test), \href{https://huggingface.co/datasets/multi_nli}{mnli} (test), \href{https://paperswithcode.com/dataset/rte}{RTE} (dev), \href{https://github.com/HKUST-KnowComp/NLI4CT}{NLI4CT (SemEval 2023 - Task 7)} (all), \href{https://github.com/microsoft/TaxiNLI}{TaxiNLI} (test), \href{https://huggingface.co/datasets/SetFit/wnli}{WNLI} (dev)
\\ \bottomrule                    
\end{tabular}
\caption{The names of datasets with low (Lo), moderate-low (M-Lo), moderate-high (M-Hi), and high (Hi) leakage, categorized according to the task. (1/3)}
\label{tab:data-leak_1}
\end{sidewaystable*}

\clearpage
\begin{sidewaystable*}[]
\centering
\small
\begin{tabular}{p{3cm}p{4.8cm}p{4.8cm}p{4.8cm}p{4.8cm}}
Task name&Lo&M-Lo&M-Hi&Hi\\\toprule
Natural Language Understanding&&&& \href{https://github.com/howl-anderson/ATIS_dataset/tree/master}{ATIS} (test), \href{https://github.com/sonos/nlu-benchmark}{SNIPS} (test)
\\ \hline \\
Paraphrasing&\href{https://www.microsoft.com/en-us/download/details.aspx?id=52398}{MRPC} (dev), \href{https://gluebenchmark.com/}{Glue (QQP)} (dev)&&&
\\ \hline \\
Politics&& \href{https://github.com/HLTCHKUST/Perplexity-FactChecking/tree/main}{Covid19 (Scientific; Social)} (test)&& \href{https://github.com/chuchun8/PStance}{P-Stance} (test), \href{https://afshinrahimi.github.io/semeval2016-task6/}{SemEval 2016 Task 6} (test), \href{https://github.com/cardiffnlp/tweeteval/tree/main/datasets/stance}{TweetEval (TweetStance)} (test)
\\ \hline \\
Programming&&&& \href{https://github.com/jkoppel/QuixBugs}{QuixBugs} (all)
\\ \hline \\
Psychology&&&& \href{https://github.com/Kali-Hac/ChatGPT-MBTI}{Myers–Briggs Type Indicator} (all)
\\ \hline \\
Question answering&\href{https://jmir.org/api/download?alt_name=mededu_v9i1e45312_app1.xlsx&filename=3c2adca5ee88328073c589af108a5697.xlsx}{Custom medical dataset from AMBOSS} (all), \href{https://github.com/facebookarchive/bAbI-tasks/tree/master}{bAbI (Task 16)} (test), \href{https://github.com/facebookresearch/clutrr}{CLUTTR} (test), \href{https://github.com/Waste-Wood/e-CARE/}{e-CARE} (dev), \href{https://github.com/SophonPlus/ChineseNlpCorpus}{FinanceZhidao} (all), \href{https://github.com/kelvin-jiang/FreebaseQA}{FreebaseQA} (all), \href{https://hotpotqa.github.io/}{HotpotQA} (dev), \href{http://lc-quad.sda.tech/}{LCQUAD 2.0} (all), \href{https://github.com/siatnlp/LegalQA}{LegalQA} (all), \href{https://github.com/lgw863/LogiQA-dataset}{LogiQA} (all), math (test), \href{https://github.com/CogComp/MCTACO}{MC-TACO} (dev), \href{https://github.com/UCSD-AI4H/Medical-Dialogue-System}{MedDialog} (all), \href{https://github.com/apple/ml-mkqa}{MKQA} (all), \href{https://github.com/ianporada/modeling_event_plausibility}{pep-3k} (all), \href{https://github.com/ybisk/ybisk.github.io/tree/master/piqa}{PIQA} (test), \href{https://whyu.me/reclor/}{ReClor} (all), \href{https://github.com/davidgolub/SimpleQA/tree/master/datasets/SimpleQuestions}{SimpleQuestions} (all), \href{https://github.com/HLR/SpartQA_generation}{SpartQA} (test), \href{https://github.com/ZhengxiangShi/StepGame}{StepGame} (test), \href{https://github.com/google-research-datasets/TimeDial}{TimeDial} (test), \href{https://www.microsoft.com/en-us/download/details.aspx?id=52763}{WebQuestions} (all), \href{https://github.com/brightmart/nlp_chinese_corpus}{WebTextQA \& BaikeQA} (all)&\href{https://github.com/facebookarchive/bAbI-tasks/tree/master}{bAbI (Task 15)} (test), Custom dataset from BCSC Self-Assessment Program (all), \href{https://aistudio.baidu.com/datasetdetail/38489}{Unnamed Chinese Psychological QA dataset} (all), \href{https://facebookresearch.github.io/ELI5/}{ELI5} (all), \href{http://tcci.ccf.org.cn/conference/2016/pages/page05_evadata.html}{NLPCC-DBQA} (all), \href{https://allenai.org/data/open-book-qa}{OpenBookQA} (dev), Custom dataset from OphthoQuestions (all), \href{https://github.com/ybisk/ybisk.github.io/tree/master/piqa}{PIQA} (dev), \href{https://allenai.org/data/qasc}{QASC} (dev), \href{https://www.cs.cmu.edu/~glai1/data/race/}{RACE} (test or dev), \href{https://allenai.org/data/socialiqa}{Social IQA} (dev), \href{https://huggingface.co/datasets/squad_v2}{SQuAD 2.0} (dev), \href{https://github.com/sylinrl/TruthfulQA}{TruthfulQA} (test), \href{https://www.microsoft.com/en-us/download/details.aspx?id=52419}{WikiQA} (all)&fiqa (all), \href{https://github.com/openai/grade-school-math}{GSM8K} (test), \href{https://thukeg.gitee.io/kqa-pro/}{KQA Pro} (all), \href{https://allenai.org/data/open-book-qa}{OpenBookQA} (test), \href{https://jmir.org/api/download?alt_name=mededu_v9i1e45312_app1.xlsx&filename=3c2adca5ee88328073c589af108a5697.xlsx}{Custom USMLE dataset} (all)&\href{https://github.com/AI-secure/adversarial-glue/tree/main}{Adversarial GLUE (qqp)} (dev), \href{https://github.com/zhongwanjun/AR-LSAT}{AR-LSAT}  (test), Custom QA dataset from BaiduBaike (all), \href{https://huggingface.co/datasets/google/boolq}{BoolQ} (test), \href{https://github.com/allenai/contrast-sets/tree/main/BoolQ}{BoolQ Contrast Set} (test), \href{https://github.com/ALFA-group/BRON}{Pre-processe version of BRON} (all), \href{https://cve.mitre.org/}{CVE (2021; ATT)} (all), \href{https://allenai.org/data/complexwebquestions}{ComplexWebQuestions} (all), \href{https://dblp.org/rdf/release/dblp-2022-06-01.nt.gz}{DBLP} (all), \href{https://efficientqa.github.io/}{EfficientQA} (dev), \href{https://dki-lab.github.io/GrailQA/}{GrailQA} (test), \href{https://github.com/ysu1989/GraphQuestions}{GraphQuestions} (all), \href{https://github.com/Hello-SimpleAI/chatgpt-comparison-detection}{HC3 (Chinese; English)} (all), \href{https://github.com/yongcaoplus/ProbingChatGPT}{Custom dataset based on the Hofstede Culture
Survey} (all), \href{https://github.com/AskNowQA/LC-QuAD2.0}{LC-QuAD 2.0} (all), \href{https://github.com/csitfun/LogiQA2.0}{LogiQA 2.0} (test), \href{https://zenodo.org/records/4617285#.YrNszNLMJhH}{MAG} (all), \href{https://jmir.org/api/download?alt_name=mededu_v9i1e45312_app1.xlsx&filename=3c2adca5ee88328073c589af108a5697.xlsx}{Custom medical dataset from NBME} (all), \href{https://ott-qa.github.io/}{OTT-QA} (all), \href{https://github.com/iesl/protoqa-data}{ProtoQA} (dev), \href{https://github.com/ag-sc/QALD/tree/master}{QALD-9} (all), \href{https://whyu.me/reclor/}{ReClor} (dev), \href{https://github.com/sylinrl/TruthfulQA/tree/main}{TruthfulQA (Generation subset) (test)}, Test of Understanding in College Economics (TUCE) (all), Wiki-csai (computer science-related concepts extracted from Wikipedia) (all), \href{https://github.com/tan92hl/Complex-Question-Answering-Evaluation-of-GPT-family/tree/main/datasets/WQSP}{WQSP} (all), \href{https://yago-knowledge.org/downloads/yago-4}{YAGO} (all)
\\ \hline \\
Reasoning \& common sense& \href{https://www.tau-nlp.sites.tau.ac.il/commonsenseqa}{CommonsenseQA}(test), \href{https://rowanzellers.com/hellaswag/}{HellaSwag} (dev), \href{https://github.com/taylorwwebb/emergent_analogies_LLM/tree/main/letter_string}{Letter String Analogies (Webb et al.)} (all)& \href{https://allenai.org/data/arc}{ARC 2018} (dev), \href{https://huggingface.co/datasets/skrishna/coin_flip}{Coin flip dataset} (all), \href{https://people.ict.usc.edu/~gordon/copa.html}{COPA} (dev), \href{https://cs.nyu.edu/~davise/papers/WinogradSchemas/WS.html}{WSC} (dev)&& \href{https://nyu-mll.github.io/CoLA/}{CoLA} (dev), \href{https://www.tau-nlp.sites.tau.ac.il/commonsenseqa}{CommonsenseQA} (dev), \href{https://github.com/google/BIG-bench/blob/main/bigbench/benchmark_tasks/date_understanding/README.md}{Date Understanding} (all), \href{https://github.com/RUCKBReasoning/CoT-KA}{Last letter dataset (all)}, \href{https://github.com/qiangning/MATRES}{MATRES} (test), \href{https://github.com/google/BIG-bench/blob/main/bigbench/benchmark_tasks/object_counting/README.md}{Object counting} (all), \href{https://allenai.org/data/strategyqa}{StrategyQA} (all), \href{https://github.com/aakanksha19/TDDiscourse}{TDDiscourse} (test), \href{https://www.usna.edu/Users/cs/nchamber/caevo/}{TimeBank-Dense} (test)
\\ \bottomrule \\                  
\end{tabular}
\caption{The names of datasets with low (Lo), moderate-low (M-Lo), moderate-high (M-Hi), and high (Hi) leakage, categorized according to the task. (2/3)}
\label{tab:data-leak_2}
\end{sidewaystable*}

\clearpage
\begin{sidewaystable*}[]
\centering
\small
\begin{tabular}{p{3cm}p{4.8cm}p{4.8cm}p{4.8cm}p{4.8cm}}
Task name&Lo&M-Lo&M-Hi&Hi\\\toprule
Semantic similarity& \href{https://adapterhub.ml/explore/sts/sts-b/}{STS-B (dev)}, \href{https://github.com/cardiffnlp/tweeteval/tree/main/datasets/emoji}{TweetEval (TweetEmoji)} (test or dev)&\href{https://github.com/MJ-Jang/BECEL/tree/main/data/mrpc}{BECEL (MRPC)} (test)&& \href{http://lcl.uniroma1.it/wsdeval/}{WSDEval (test or dev)}, \href{https://pilehvar.github.io/wic/}{WiC} (dev), \href{https://pilehvar.github.io/wic/}{WiC}(test or dev)
\\ \hline \\
Sentiment analysis& \href{https://github.com/Moradnejad/ColBERT-Using-BERT-Sentence-Embedding-for-Humor-Detection/tree/master/Data}{ColBERT} (test or dev), \href{https://www.kaggle.com/datasets/niraliivaghani/flipkart-product-customer-reviews-dataset}{Flipkart Product Reviews} (all), \href{https://www.cs.cornell.edu/people/pabo/movie-review-data/}{IMDb Movie Review Data} (test), \href{https://github.com/YJiangcm/SST-2-sentiment-analysis}{SST-2} (dev), \href{https://github.com/conversationai/unhealthy-conversations}{UCC} (test or dev), \href{https://github.com/ewulczyn/wiki-detox/}{UnhealthyPer} (test or dev) \href{https://github.com/ewulczyn/wiki-detox/}{WikiDetox (aggression task)} (test or dev), \href{https://github.com/CLARIN-PL/chatgpt-evaluation-01-2023/}{AggressionPer} &\href{https://github.com/google-research/google-research/tree/master/goemotions}{GoEmotions} (test or dev), \href{https://github.com/CLARIN-PL/chatgpt-evaluation-01-2023/}{GoEmoPer0-3} \href{https://github.com/SALT-NLP/implicit-hate}{Implicit Hate Corpus} (all), \href{https://www.kaggle.com/datasets/rmsharks4/sarcasmania-dataset}{Sarcasmania (sarcasm task)} (test or dev), \href{https://codalab.lisn.upsaclay.fr/competitions/7096#learn_the_details}{SemEval 2023 - Task 9 (test)}, \href{https://github.com/cardiffnlp/tweeteval/tree/main/datasets/sentiment}{TweetEval - Sentiment} (test or dev)& \href{https://github.com/allenai/real-toxicity-prompts}{Real Toxicity Prompts} (all)& \href{https://adversarialglue.github.io/instructions/}{AdvGLUE} (SST-2) (dev), CLARIN-Emo (test or dev), \href{https://chalearnlap.cvc.uab.cat/dataset/24/description/}{ChaLearn 2016 FI} (personality task) (all), \href{https://github.com/allenai/contrast-sets/tree/main/IMDb}{Contrast Sets (IMDb)} (all), \href{https://clarin-pl.eu/dspace/handle/11321/710}{PolEmo 2.0} (test or dev), \href{https://huggingface.co/datasets/sentiment140}{Sentiment140} (all), \href{https://www.kaggle.com/datasets/nikhileswarkomati/suicide-watch}{The Suicide and Depression Dataset} (all)
\\ \hline \\
Summarization&\href{https://huggingface.co/datasets/cnn_dailymail}{CNN DailyMail} (test), \href{https://github.com/csebuetnlp/CrossSum}{CrossSum (En - Zh)} (test), \href{https://github.com/ctr4si/MMN}{Reddit TIFU} (test), \href{https://github.com/esdurmus/Wikilingua}{WikiLingua (En - Zh/De)} (test), \href{https://github.com/krystalan/ClidSum/tree/main#2-clidsum-benchmark-dataset}{XSAMSum (En - Zh/De)} (test)& \href{https://github.com/honglizhan/CovidET}{CovidET} (test), \href{https://github.com/ali-bahrainian/NEWTS}{NEWTS} (test), \href{https://github.com/armancohan/long-summarization/tree/master}{PubMed dataset} (test), \href{https://github.com/Yale-LILY/QMSum}{QMSum} (test), \href{https://github.com/EdinburghNLP/XSum/tree/master/XSum-Dataset}{XSum} (test)& \href{https://github.com/nyu-mll/SQuALITY}{SQuALITY} (test)&\href{https://paperswithcode.com/dataset/samsum-corpus}{SAMSum} (test)
\\ \hline \\
Text classification& \href{https://github.com/inverse-scaling/prize}{Inverse Scaling Prize} (all datasets) (all) (all)&&&\href{https://huggingface.co/datasets/ml4pubmed/pubmed-classification-20k}{PubMed20K} (train), \href{https://www.kaggle.com/datasets/uciml/sms-spam-collection-dataset}{SMS Spam Collection V1} (test or dev), \href{https://www.kaggle.com/datasets/paultimothymooney/medical-speech-transcription-and-intent}{Symptoms dataset} (train)
\\ \hline \\
Text extraction& \href{https://mtsamples.com/}{MTSamples} (all)& \href{https://www.i2b2.org/NLP/Relations/}{I2B2 2010} (all)&& \href{https://paperswithcode.com/dataset/ace-2005}{ACE 2005} (all), \href{https://github.com/ZihanWangKi/CrossWeigh}{CoNLL++} (all), \href{https://huggingface.co/datasets/conll2003}{CoNLL 2003} (test), \href{https://github.com/zhoujx4/DuEE}{DuEE 1.0} (all), \href{https://github.com/zhoujx4/DuIE}{DuIEduie 2.0} (all), \href{https://github.com/OYE93/Chinese-NLP-Corpus/tree/master/NER/MSRA}{MSRA} (all), \href{https://github.com/truthless11/HRL-RE/tree/master/data/NYT11}{NYT11-HRL} (all)
\\ \hline \\
Text generation&& \href{https://www.comp.nus.edu.sg/~nlp/conll14st.html}{CoNLL 2014 Shared Task dataset} (test)&&\href{https://github.com/microsoft/ContextualSP}{ADVETA (ADD, RPL) }, \href{https://yale-lily.github.io/cosql}{COSQL} (dev), \href{https://taolusi.github.io/CSpider-explorer/}{CSpider} (dev), \href{https://github.com/luge-ai/luge-ai/tree/master/semantic-parsing}{DuSQL} (all), \href{https://github.com/salesforce/QGen/tree/main/Quiz_Design}{Quiz Design} (all), \href{https://github.com/taoyds/sparc}{SParC} (dev), \href{https://drive.usercontent.google.com/download?id=1TqleXec_OykOYFREKKtschzY29dUcVAQ&export=download&authuser=0}{Spider} (dev), \href{https://github.com/ygan/SpiderSS-SpiderCG}{Spider-CG (app, sub)} (all), \href{https://github.com/ygan/Spider-DK}{Spider-DK} (dev), \href{https://zenodo.org/record/5205322}{Spider-Realistic} (dev), \href{https://github.com/ygan/Spider-Syn}{Spider-Syn} (dev)
\\ \bottomrule \\                  
\end{tabular}
\caption{The names of datasets with low (Lo), moderate-low (M-Lo), moderate-high (M-Hi), and high (Hi) leakage, categorized according to the task. (3/3)}
\label{tab:data-leak_3}
\end{sidewaystable*}

\end{document}